\documentclass[preprint,12pt]{elsarticle}

\usepackage{lineno,hyperref}
\usepackage{amsmath}
\usepackage{mathtools}
\usepackage{amsfonts}
\usepackage{amssymb}
\usepackage{amsthm}
\usepackage{bm}
\usepackage{listings}
\usepackage[mathscr]{euscript}
\DeclareSymbolFont{rsfs}{U}{rsfs}{m}{n}
\DeclareSymbolFontAlphabet{\mathscrsfs}{rsfs}
\usepackage{subcaption}
\newtheorem{definition}{Definition}[section]

\newcommand{\tuple}[1] {\langle #1 \rangle}
\usepackage{cleveref}
\usepackage{algorithm}
\usepackage[noend]{algpseudocode}
\usepackage[hypcap=true]{subcaption}
\newtheorem{theorem}{Theorem}
\usepackage{multirow}
\let\oldReturn\Return
\renewcommand{\Return}{\State\oldReturn}
\usepackage{longtable}
\newcounter{example}[section]

\usepackage[dvipsnames]{xcolor}
\usepackage[table]{colortbl}
\usepackage[colorinlistoftodos,prependcaption,textsize=tiny]{todonotes}
\usepackage{tcolorbox}
\usepackage{pdfcomment}

\usepackage{ulem}
\usepackage{apxproof}

\modulolinenumbers[5]

\journal{a journal}

\begin{document}

\begin{frontmatter}

\title{Risk Awareness in HTN Planning} %

\author{Ebaa Alnazer\corref{cor1}}
\ead{ebaa.alnazer@iaas.uni-stuttgart.de}
\cortext[cor1]{Corresponding author.}

\author{Ilche Georgievski}
\ead{ilche.georgievski@iaas.uni-stuttgart.de}

\author{Marco Aiello}
\ead{marco.aiello@iaas.uni-stuttgart.de}

\address{Service Computing Department, Institute of Architecture of Application Systems, University of Stuttgart, Germany}

\begin{abstract}
Actual real-world domains are characterised by uncertain situations in which acting and using resources may entail the embracing of risks. Performing actions in such domains involves costs of consuming some resource, such as time, money, or energy, where the knowledge about these costs can range from known to totally unknown; it might even mean unknowable probabilities of costs. Think of the Autonomous Vehicles (AVs) domain, where actions (e.g., moving actions) and their costs (e.g., time, power, fuel, and money) are non-deterministic due to uncertainties like obstacles, traffic, and weather conditions. Deciding which action to take considering its cost on the available resources, such as choosing between a highway or shortcut, requires assessing the potential risks of delay, where each road has an uncertain level of traffic congestion. Thus, these domains call for not only planning under uncertainty but also planning while embracing risk. Resorting to Hierarchical Task Network (HTN) planning as a widely used planning technique in real-world applications, one can observe that existing approaches do not account for risk, as they compute the most probable or optimal plans based on single-valued action costs, reflecting only risk neutrality. In this work, we enhance HTN planning with risk awareness by considering expected utility theory, a representative concept of decision theory that enables action selection considering a probability distribution of their costs and a given risk attitude expressed using a utility function. Specifically, we introduce a general framework for HTN planning that allows modelling risk and uncertainty using a probability distribution of action costs upon which we define risk-aware HTN planning as being capable of accounting for the different risk attitudes and allowing the computation of plans that go beyond risk neutrality. We lay out that computing risk-aware plans requires finding plans with the highest expected utility. We argue that it is possible for HTN planning agents to solve specialised risk-aware HTN planning problems by adapting existing HTN planning approaches, and we develop an approach that surpasses the expressiveness of current approaches by allowing HTN planning agents to compute plans tailored to a particular risk attitude. An empirical evaluation of two case studies highlights the feasibility and expressiveness of this approach. Lastly, we discuss open issues introduced by the present proposal, including its generalisability to planning techniques other than HTN planning, addressing both modelling and plan generation aspects.
\end{abstract}


\end{frontmatter}

\tableofcontents

\normalem
\section{Introduction}
\label{intro}

The \textit{innovator's dilemma} is the hard decision organisations face when they have to choose between sustaining innovation or accepting disruptive innovation~\cite{christensen1997:innovator-dilemma}. Are they ready to give up the innovations they have made and invest in the unknown? Accepting disruptive innovation requires changing established attitudes that focus on security and aversion towards taking chance when it comes to decision making and resource allocation. We can, thus, say that the innovator's dilemma is all about embracing \textit{risk} in the presence of \textit{uncertainty}. 

The innovator's dilemma is also applicable to domains other than the business one. In general, real-world domains are typically non-deterministic and exhibit a wide-ranging spectrum of uncertainty, requiring planning and decision making processes to embrace risk in one form or another. In this context, decision theory offers mechanisms to rank options on how choice-worthy they are. The most representative mechanism is the expected utility theory, which sets a fundamental principle by which decisions are made in environments, where actions have a probability distribution of variable costs and outcomes, according to a given risk attitude~\cite{morgenstern1953theory}.

The requirement for planning and decision making processes to embrace risk in the presence of uncertainty also holds when automating the processes of decision making, which is of primary concern in Artificial Intelligence Planning. 
In its simplest form, AI planning is about the generation of a course of action whose execution in an initial state of the world satisfies some user objective. In AI planning literature, uncertainty is considered in the initial state and actions in the form of incomplete knowledge (e.g.,~\cite{hoffmann2005contingent,hoffmann2006conformant}) and multiple effects of actions (e.g.,~\cite{chen2021fully}), possibly with a probability of their execution (e.g.,~\cite{macedo2004emotional,bouguerra2004hierarchical}). 

The concept of risk, however, has not been treated much despite the fact that there are planning problems in which performing actions always incurs costs on a resource of interest, such as money, time, energy, and/or effort. For some problems, it might be preferred to accept plans with larger execution costs in order to avoid risk. For others, risky plans are preferred if they promise, even with a low probability, lower execution costs. The current practice that actions have a cost of typically one unit does not reflect reality. Action costs are not static, cannot be easily predefined, and sometimes cannot be even predicted reliably. Take, for example, the domain of autonomous vehicles (AVs) that employ advanced driving systems to automate route planning and vehicle's control, aiming at performing driving tasks given the constantly changing road environments. To achieve their objectives, the AVs need to plan their tasks by selecting driving actions under risk related to the costs of these actions, represented by travelling times, power, and even human lives. The existence of risk during planning stems from the inherent uncertainties in this domain caused by uncertain weather forecasts, traffic jams, the AV's technological capabilities, and emerging obstacles, among other factors. In general, the uncertainty spectrum starts at totally known probability distribution over action costs and ends at totally unknown and even unknowable costs and probabilities~\cite{lo2010warning}. Similarly to having actions with multiple effects, having this variability of action costs can lead to plans of a probabilistic nature. That is, the resource consumption of the same plan can differ from one plan execution to another. Thus, it is of utmost importance to compute plans that embrace risk by taking the variability of action costs into account. As a result, the quality of plans plays a role when planning under risk.

\subsection{HTN Planning and Its Risk Neutrality}

Among AI planning techniques, Hierarchical Task Network (HTN) planning is a widely used one in real-world domains, such as games, e.g.~\cite{kelly2008offline}, robotics, e.g.~\cite{weser2010:robot}, healthcare, e.g.,~\cite{fdez2011supporting}, Web service composition, e.g,~\cite{wu2003automating}, cloud computing, e.g.~\cite{georgievski2017:cloudapps}, and building automation, e.g.~\cite{georgievski2017:pmc}. This technique is essentially based on the idea of enriching planning domains with knowledge on how to accomplish tasks~\cite{georgievski2015htn,bercher2019survey}. This rich domain knowledge and its intrinsic hierarchical structure enable HTN planning to provide a natural approach to simulate the way in which one conceptualises decision making in domains that involve risk. In particular, HTN planning requires an initial state, an initial task network as an objective to be accomplished, and domain knowledge consisting of networks of primitive and compound tasks. A task network represents a hierarchy of tasks each of which can be directly executed, if the task is primitive, or decomposed via methods into subtasks, if the task is compound. One way to search for solutions is to start decomposing the initial task network and continue doing it until all compound tasks have been decomposed, i.e., a task network consisting of only primitive tasks is reached. The solution, if it exists, is a plan which equates to a linearisation of the set of primitive tasks applicable to the initial state.

In the realm of uncertainty, there are HTN planning approaches that deal with probabilistic plans and differ in what selection criteria each approach uses to rank plans. Some planning approaches do not consider action costs, but rather consider probabilistic effects of actions. These approaches aim at finding plans with the highest probability of success, e.g.,~\cite{biundo2004dealing}. Other approaches do take action costs into account, but aim at finding plans that have the highest probability of not exceeding a predefined cost limit, e.g.,~\cite{lin2020bounded}. There are even approaches that assume the world is deterministic and assign a single real-valued number to each action to quantify its resource consumption~\cite{bercher2017admissible,bechon2014hipop}. Such approaches usually aim at computing cost-optimal plans, where the plan's cost is the total sum of the costs of its constituent actions. 

In general, an agent acting on the basis of such HTN planning approaches is indifferent to the risk that may arise due to the made planning choices. Thus, this planning agent is \textit{risk neutral}. And the working and objectives of risk-neutral HTN planning agents do not entirely meet the expectations of decision makers in actual settings. Decision making and planning in real-world situations requires \textit{embracing} the risk.

\subsection{Proposal, Contributions, and Organization}

We show that \textit{risk awareness} of HTN planning can be achieved by considering the expected utility theory. Specifically, we draw on the view of expected utility theory towards rationality when making decisions under risk to suggest that different risk attitudes can be incorporated into HTN planning.
The risk attitudes are expressed using utility functions that map operator costs into real values, provided there is a variability of those costs. The calculation derived from expressing risk attitudes over action costs can be propagated to higher hierarchical levels, i.e., methods and compound tasks. This enables us to propose a definition of risk-aware HTN planning in which plans have the maximum expected utility and are computed by making informed planning choices at all levels of the hierarchy. As a result, the quality of plans or risk awareness of plans emerges.

This article makes the following contributions, in order of appearance:
\begin{itemize}
\item We provide a broader perspective of uncertainty and risk, where we define uncertainty and risk as two distinct concepts. We mainly draw knowledge from decision theory and its established mechanisms for embracing risk when making decisions. This opens the pathway for the presentation of our main technical contributions to AI Planning.
\item We identify the sources of uncertainty in relation to action costs and position them with respect to the available knowledge from decision theory.
\item We provide a general framework of HTN planning for uncertainty and risk based on the variability of action costs. While the framework builds upon an existing HTN planning formalism, to the best of our knowledge, it represents the first general approach that accounts for the variability of action costs. While this enables us to define our risk-aware HTN planning, it also represents a stepping stone for new approaches.
\item We propose the novel concept of risk-aware HTN planning, which is capable of embracing risk in real-world domains for which HTN planning is a fit. This concept is inspired by our preliminary idea~\cite{georgievski2014:ecai}, and, to the best of our knowledge, represents the first work that incorporates risk in HTN planning, and it does so upon established concepts from decision theory. Risk-aware HTN planning paves the way to having more specific risk-aware HTN planning problems and constructing algorithms that can solve them.
\item We develop an approach to solving risk-aware plan-based HTN planning problems under some simplifying assumptions by adapting existing HTN planning approaches.
\item We go one step further and suggest possible ways to solving state-based risk-aware HTN planning problems also by adapting existing HTN planning approaches.
\item We perform empirical evaluations using two case studies, namely, Autonomous Vehicles (AVs) and Satellite domains. The empirical evaluation demonstrates the feasibility and expressivity of our developed approach.
\item We provide a wider overview of works that deal not only with planning under uncertainty and risk but also with mechanisms that do not relate to risk but do support informed decision making in HTN planning.
\end{itemize}

\noindent
The remainder of the article is organised as follows. Section~\ref{sec:uncertainy-risk} presents the perspective of uncertainty and risk from a decision theory standpoint. Section~\ref{sec:uncertainty-risk-realworld} introduces sources of uncertainty and its effects on action costs. Section~\ref{sec:uncertainty-htnplanning} presents the formalism of the general framework we propose, while Section~\ref{riskAware-htnplanning} introduces risk-aware HTN planning. Section~\ref{sec:solving-plan-based} provides details about our approach to solving specific risk-aware plan-based HTN planning problems. Section~\ref{sec:onSolving-state-based} gives insights into some possible approaches for solving specific risk-aware state-based HTN planning problems. Section~\ref{sec:experiments} contains empirical evaluations on the Autonomous Vehicles (AVs) and Satellite case studies. Section~\ref{sec:uncertainty-aiplanning} provides an overview of the related work, while Section~\ref{sec:discussion} includes a discussion on selected questions related to our proposal. Section~\ref{sec:conclusion} finalises the article with concluding remarks.

\section{Uncertainty and Risk in a Broader Perspective}
\label{sec:uncertainy-risk}

Individuals are continuously confronted with situations that necessitate making decisions. If, for simple situations, an immediate and intuitive course of action suffices, in more intricate situations, more options will be available, and the effects of initial choices on subsequent states will be uncertain and largely hard to intuitively predict.

\subsection{General definitions}
\label{sec:general-definitions}

To systematise decision making, research in various fields has been carried out with the aim of providing decision makers with conceptual understanding and methodical ways to analyse and reason about different alternatives. In decision theory, two factors have been identified that increase the complexity of decision-making problems, namely \textit{uncertainty} and \textit{risk}~\cite{Park2017}.

The awareness of the distinction between uncertainty and risk has been present for decades in the field of economics. In 1921, Frank H. Knight made an explicit distinction between uncertainty and risk in his classic economic theory presented in~\cite{knight1921risk}. Knight defines uncertainty as a decision-making situation in which the likelihoods of alternative outcomes are unknown to the decision maker or are impossible to form, i.e., incalculable due to their uniqueness or due to their irregularity. Risk, on the other hand, is present when all outcomes and their probability of occurrence are known either a priori or from statistics gathered from past experience.

\begin{tcolorbox}
Despite such a clear distinction between uncertainty and risk, there have been also attempts to view these two terms as the same concept. For example, in 1966, the Committee on General Insurance Terminology defined risk as ``uncertainty as to the outcome of an event when two or more possibilities exist". This definition has been widely criticised due to its inability to distinguish between risk and uncertainty with respect to the probability distribution of outcomes~\cite{head1967alternative}.
Another example is the handbook for risk management, where risk is defined as ``uncertainty that, if it occurs, will have a positive or negative effect on the achievement of objectives"~\cite{hillson2016risk}. This means that risk is a subset of uncertainties that matter to the decision maker, i.e., affect the achievement of the decision maker's objectives. The authors also argue that Knight's distinction between risk and uncertainty is useful as a mathematical theory but that it may not yield useful solutions in practice.
\end{tcolorbox}

Similar knowledge can be found in game theory, where decision making is classified according to whether it is affected by certainty, uncertainty, and risk~\cite{luce1989games}. Certainty is defined as the situation in which the decision maker knows the exact outcome of alternatives, while uncertainty and risk are defined similarly to Knight's definitions.

\begin{figure}
	\centering
	\includegraphics[width=\columnwidth]{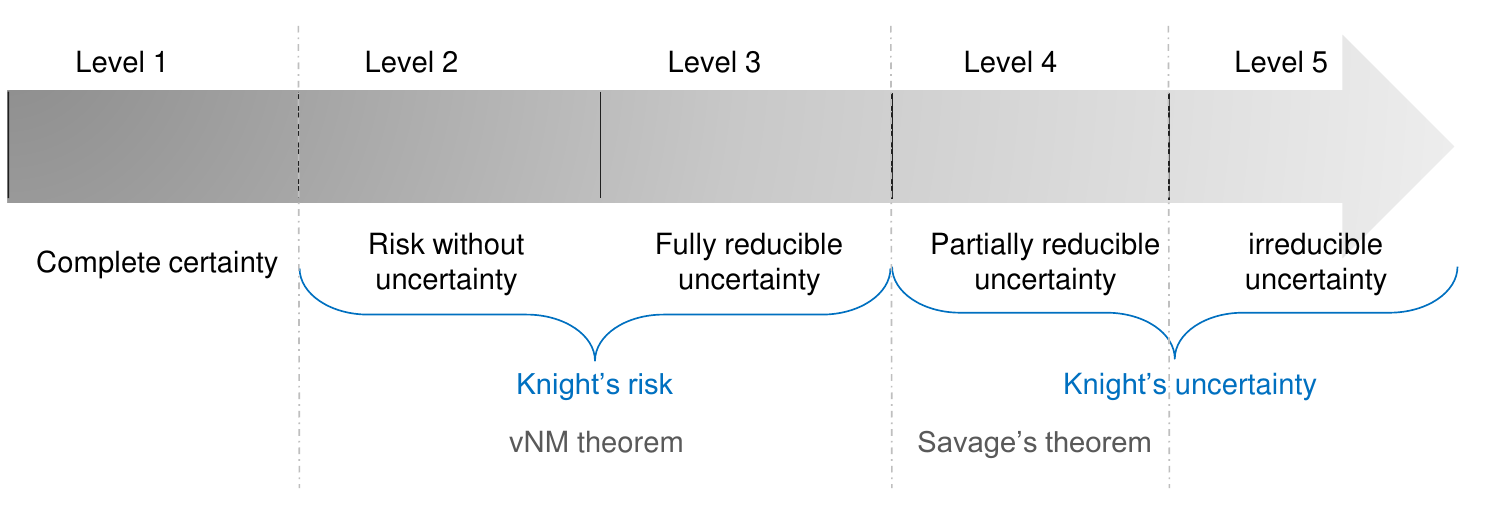}
	\caption{Uncertainty taxonomy from~\cite{lo2010warning} mapped to Knight's definitions of risk and uncertainty and vNM's and Savage's theorems.}
	\label{fig:uncertainty-taxonomy}
\end{figure}

Knight's distinction between uncertainty and risk has been further refined into a taxonomy that captures the relation between uncertainty and risk in physics~\cite{lo2010warning}. The uncertainty taxonomy provides a wide spectrum of uncertainty formulated in five levels, as illustrated in Figure~\ref{fig:uncertainty-taxonomy}. These levels range from complete certainty to irreducible uncertainty:
\begin{itemize}
    \item Level 1 defines \textit{complete certainty}, i.e., nothing is uncertain.
    \item Level 2 defines \textit{risk without uncertainty}, where outcomes and their probability distribution are known.
    \item Level 3 defines \textit{fully reducible uncertainty}, where the outcomes are fully known, but their probability distribution is unknown. The uncertainty in this level is fully reducible to risk by statistical inference of the probability distribution of outcomes.
    \item Level 4 is \textit{partially reducible uncertainty}, where there is a limit to what we can deduce about the outcomes and their probability even by significant statistical inference, and a significant amount of the outcomes and their probabilities are uncertain, which leads to model uncertainty. The probabilities in this level reflect beliefs rather than frequencies of repeated trials as defined in Levels~2 and~3.
    \item Level 5 defines \textit{irreducible uncertainty}, which is the state of total ignorance that can be solved by neither collecting more data nor using sophisticated statistical inference methods.
\end{itemize}

We map Knight's definition of risk to Levels 2 and 3 and his definition of uncertainty to Level 4, as shown in Figure~\ref{fig:uncertainty-taxonomy}. In this context, our work adopts the distinction of uncertainty and risk and, furthermore, defines risk as a combination of Knight's definition with Levels 2 and 3 of the uncertainty taxonomy and uncertainty as a combination of Knight's definition with Level 4 of the taxonomy. As a means to make rational decisions under risk and uncertainty, theories in decision-making can be categorised following the distinction between the two concepts. In particular, we map Level 2 and Level 3 to the \textit{vNM theorem} since it addresses decision-making under risk, and we map Level 4 to the \textit{Savage's theorem} as it deals with decision-making under uncertainty. A detailed explanation of these theorems is provided in Section~\ref{subsub:rationality-decision-making}. 

Thus, risk and uncertainty are defined based on the literature of decision theory as follows.

\begin{definition} [Risk]
   Risk is a decision-making situation in which either all outcomes and their probability of occurrence are known a priori or the probability distribution of outcomes is unknown but can be deduced using statistical inference.
\end{definition}

\begin{definition} [Uncertainty]
	 Uncertainty is a decision-making situation in which either there is a limit to what can be deduced about the probability distribution of alternative outcomes, where the probabilities represent degrees of the decision maker's beliefs, or the probability distribution is not only unknown but also unknowable.
\end{definition}

\subsection{Uncertainty and Risk Through the Lenses of Utility}
\label{uncertainty-risk-utilityTheory}

Risk and uncertainty play a central role in making rational decisions. Let us look at this through the lens of utility theory -- the most representative theory in decision-making that explains the acts of rational choices using the concept of utility. More specifically, utility theory provides a mathematical framework for modelling decision-making under risk and uncertainty and explains people's behaviour on the premise that people can rank choices based on their preferences in terms of the satisfaction of all decision outcomes~\cite{fishburn1968utility}.

We demonstrate that distinguishing between risk and uncertainty is useful for categorising theories within \textit{expected utility theory}--the axiomatic theory of choice--into those that address decisions under risk and those that handle decisions under uncertainty~\cite{karni1991utility}. We explain the existence of rationality in decision-making and how this can be expressed in different attitudes toward risk using utility functions.

\subsubsection{Rationality in Decision Making}
\label{subsub:rationality-decision-making}

In 1738, Daniel Bernoulli posited the \textit{expected utility theory} by making a clear distinction between the expected value and the \textit{expected utility}, as the latter uses weighted \textit{utility}, the value of the outcome to the decision maker, multiplied by probabilities instead of using weighted outcomes. The theory was first axiomatised and mathematically formulated by von Neumann and Morgenstern (vNM) in 1944~\cite{morgenstern1953theory}. They formulated what is known as the {\em vNM Theorem}, which suggests that maximising the expected utility is the objective of a rational agent, where the decision maker's preference structure over outcomes is assumed, \textit{utilities} of outcomes are known, and the decision maker knows the ``objective'' probability of outcomes. A utility measures the subjective worth of an outcome, whether it is a monetary value or any other type of value, by mapping outcomes to real-valued utilities using a \emph{utility function}. The utility function formalises the decision maker's preference structure. The expected utility is the sum of the utilities of outcomes weighted by the corresponding probabilities. Since the theorem assumes ``objective'' utilities are known, it incorporates the notion of risk through the expected utility theory.

Savage's theorem is considered the generalisation of the vNM theorem, and it approaches decision-making under uncertainty~\cite{savage1972foundations}. The theorem suggests that a rational decision maker makes choices as if she/he is maximising the expected utility using a ``subjective'' probability distribution, which is a translation of the decision maker's beliefs about the outcomes and differs from one decision maker to another.

Considering the taxonomy of risk and uncertainty from Section~\ref{sec:general-definitions}, we can map the different taxonomy levels to the two theorems of expected utility, as illustrated in Figure~\ref{fig:uncertainty-taxonomy}. The vNM theorem focuses on decision-making situations that involve either risk or uncertainty that is fully reducible to risk, i.e., Levels 2 and 3 of the taxonomy. Savage's theorem, on the other hand, addresses situations that involve partially reducible uncertainty, where the probability of the outcomes represents the decision maker's beliefs, i.e., Level 4 of the taxonomy.

\subsubsection{Risk Attitudes}
\label{subsec:riskAttitudes-utilityFunctions}

In the world of uncertainty and risk, agents make decisions that reflect a specific \emph{risk attitude}, which defines people's mindset towards taking risk. Some decision makers have a simple objective of minimising expected loss. These decision makers are indifferent to the risk involved in the various choices and they focus solely on the expected loss each alternative entails. In other words, they are \textit{risk neutral}. However, in domains that are characterised by huge wins or losses, that is, \emph{high-stake domains}, decision makers have objectives that go beyond minimising expected costs to account for the degree of risk associated with each choice. In reality, decision makers are seldom risk neutral. In fact, decision makers might be \textit{risk averse}, i.e., they avoid risky choices that can expose them to a high degree of loss. For example, risk-averse decision makers will always choose to insure valuable assets, such as homes and cars, to avoid the potential loss of these assets. Although the probability of a loss may be small, the potential loss of the asset itself would be very large. Thus, these individuals are willing to rather pay a monthly fee to insurance companies rather than face the risk of potential losses.

The opposite of the risk-averse attitude is the \textit{risk seeking} one. Decision makers with this attitude tolerate losses more than risk-averse individuals and prefer risky alternatives that have the potential to result in high returns. Thus, when they are offered two choices with the same expected utility, they prefer the risky choice if it has the potential to result in higher returns. For example, if a risk-seeking individual is given the choice between a gamble and a sure outcome, s/he prefers the gamble if there is a probability of higher returns.

\subsubsection{Dynamics of Risk Attitudes}
Risk attitudes of decision-makers usually follow one of two patterns: \textit{static} or \textit{dynamic}. Decision-makers have a static risk attitude when their attitudes do not change over time and are not affected by any factor, such as their wealth level. On the other hand, some decision-makers have a dynamic risk attitude that changes with some factors, such as the wealth level or the decision-making history of the decision-maker. For example, some studies statistically prove that the income of a decision-maker is positively related to her/his risk attitude. This means that an increase in the decision-maker's income increases the odds of him/her being risk-seeking, and a decrease in income increases the odds of the decision-maker being risk-averse~\cite{wright2017extent}. Arguments exist that most people are risk-averse when they have a small amount of money and become more and more risk-neutral when they get richer and richer~\cite{liu2008exact,liu2005risk}. This property of switching attitude was proposed in~\cite{bell1988one} and studied further in~\cite{bell2001strong}. An example of such behaviour where a contestant in the TV show “Who Wants to be a Millionaire” has reached the one million dollar question, for which s/he does know the answer, and s/he has two alternatives to choose from is illustrated in~\cite{liu2005risk}. S/He can either leave with \$500,000 for sure or guess the answer and then win \$1,000,000 with 50\% probability (if the answer is correct) and \$32,000 with 50\% probability (if the answer is wrong). For ordinary people, the rewards are high compared to their wealth. Thus, they are expected to be risk-averse and leave. However, if the contestant is a billionaire, the wealth levels are low compared to her/his wealth. Thus, it is expected that s/he chooses to answer the question.

The history of the decision maker is another factor that can influence the dynamics of his/her risk attitude. A history-dependent risk-aware model for decision making defines a behavioural model in which the decision maker's risk attitude changes based on his/her history of disappointments and elation~\cite{dillenberger2015history}. To determine whether an outcome of a certain choice is disappointing or elating, the decision maker assigns a threshold above which the outcome is considered elating or disappointing otherwise. This history of elation and disappointments reinforces the risk attitude of the decision maker, i.e., the decision maker's risk aversion decreases after an elating experience and increases after a disappointing one. In addition, the decision makers are proved to show a \emph{primacy effect}. The primacy effect indicates that the sequence of outcomes matters, i.e., the earlier the decision maker is disappointed, the more risk averse she/he becomes. This behavioural model is evident in a variety of real-world domains. For example, after the 2008 financial crisis in Italy, Italian investors showed a substantial increase in their risk aversion during risky investments compared to their risk attitudes before the crisis~\cite{guiso2018time}. Examples from other studies show how the primacy effect plays a big role in shaping the risk attitude of decision makers~\cite{barron2008effect,kaustia2008investors}.

\subsubsection{Utility Functions}

The risk attitudes of decision-makers are determined by utility functions. In particular, each decision-maker has a strictly monotonically non-decreasing utility function that transforms the real-valued outcomes into real-valued utilities. The decision-maker always tries to maximise his/her expected utility under a set of axioms. 

For decision-makers who are not risk-sensitive, i.e., they are risk neutral, the utility function is linear, and, thus, their behaviour is reward maximisation (in gain domains) or cost minimisation (in loss, i.e., cost-based domains). On the other hand, if the decision-maker is risk-sensitive, their utility function is non-linear; for instance, it could be exponential. If the utility function is concave, the decision-maker is risk-averse, while if the utility function is convex, the decision-maker is risk-seeking. Such utility functions express the static risk attitude of decision-makers. 

A different type of utility function should be used to model dynamic risk attitudes. For example, Bell~\cite{bell1988one} and Bell and Fishburn~\cite{bell2001strong} discuss a class of utility functions that can be used to express the switch in the decision-maker's attitude with the change of his/her wealth level. The utility functions belonging to this class are called $m$-switch utility functions, where $m$ is the number of switches in the attitude. Then, the zero-switch utility functions are the utility functions that express static risk attitudes. The most commonly used utility function from this class is the one-switch utility function, which expresses that for every pair of alternatives whose ranking is not independent of the wealth level, a wealth level exists above which one alternative is preferred, below which the other is preferred. This utility function is a linear combination of linear and exponential utility functions~\cite{liu2008exact,zeng2014risk}. In addition to the studies that focus on wealth-dependent risk attitudes, there are studies that focus on studying utility functions that are dependent on other factors. For example, one can find utility functions for history-dependent risk awareness in~\cite{dillenberger2015history}.
\section{Sources and Effects of Uncertainty}
\label{sec:uncertainty-risk-realworld}

To fulfil our objective of creating a general framework for HTN planning under uncertain situations, we start by studying the sources of uncertainty in real-world domains and their effects on the execution costs of actions.

\subsection{Sources of Uncertainty}

One important factor that \textit{planning agents}, i.e., the decision makers in an autonomous system, need to account for when planning in real-world domains is the source of uncertainty. A decision maker can be either a human being or an AI planning component of an autonomous system. However, henceforth, we use planning agents to refer only to the latter type because our focus here is on automated planning. To understand the sources of uncertainty, it is important to know what type of agents operate in a given domain, what we refer to as \textit{executing agents}. More specifically, we distinguish two types of executing agents. The first type is human beings, where a human is either instructed by the planning agent to execute actions or acts based on his/her free will, while the second type is a system, which represents any type of executors that are not human beings, e.g., robots and Internet of Things (IoT) actuators.

\begin{figure}[t!]
	\centering
	\includegraphics[width=\textwidth]{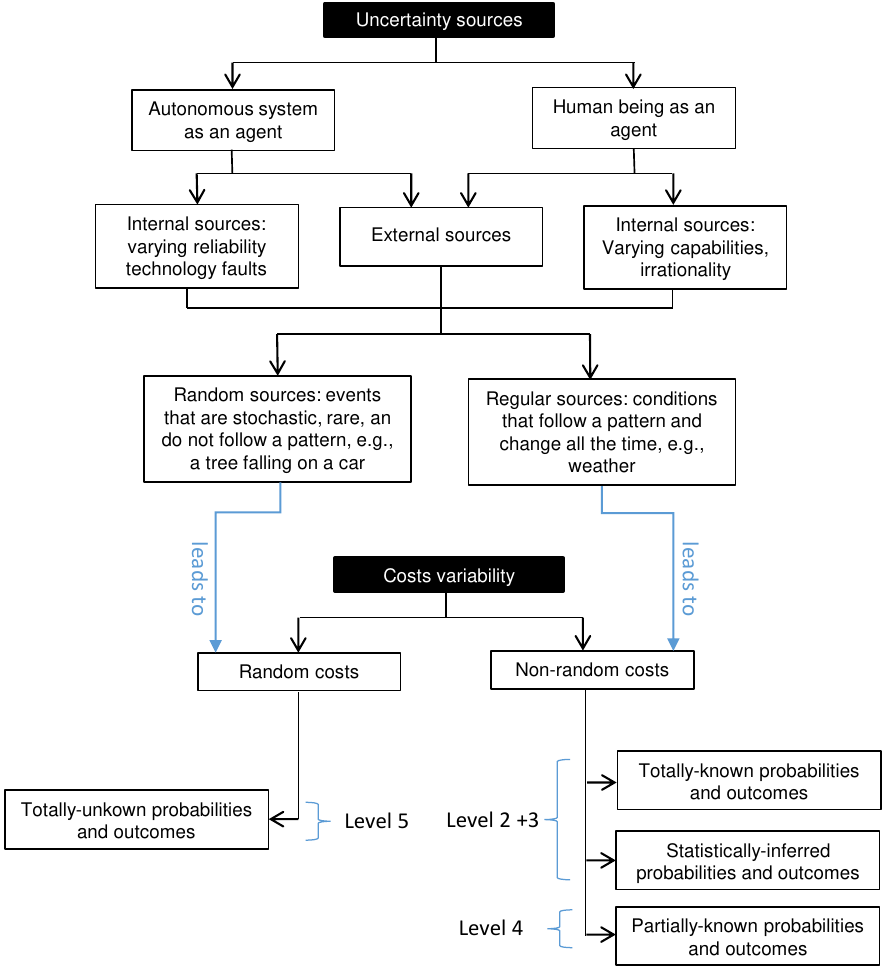}
	\caption{Categorisation of uncertainty sources and their effects on action costs.}
	\label{fig:uncertainty-sources}
\end{figure}

Given these types of executing agents, we can refine and categorise the sources of uncertainty that planning agents should consider in three groups. The first and second categories include sources of uncertainty when the executing agent is a system and when it is a human, respectively. The third category considers sources of uncertainty in domains where both systems and humans can execute actions. These categories of sources of uncertainty are depicted in Figure~\ref{fig:uncertainty-sources}.

In each category, we further distinguish two types of sources of uncertainty, namely \textit{internal sources} and \textit{external sources}. This categorisation is similar to that used in the context of AVs~\cite{azevedo2021internal}. While these sources are referred to as sources of risk in that context, we generalise the term and refer to them as sources of uncertainty. Internal sources are those caused by the agent itself. If the agent is an autonomous system, internal sources of uncertainty appear due to the varying reliability of the system or due to malfunctioning of the whole agent or parts of it. If the agent is a human being, these sources include the variable capabilities or irrationality of humans. The external sources, on the other hand, are sources of uncertainty caused by the environment surrounding the agent and independent of the agent's actions. Each type of source is further categorised into \textit{random sources} and \textit{regular sources}. Random sources of uncertainty are events that are stochastic, rare, and do not follow a pattern. Regular sources are conditions that follow some pattern and change all the time within the agent itself, in the case of internal sources, or within the surrounding environment, in the case of external sources. For illustrations of these sources, see~\ref{apx:case-studies}, where we present two examples of real-world domains that exhibit elements of uncertainty, namely smart homes and marine environments.

The sources of uncertainty reported here are related to the uncertainty dimensions presented in~\cite{georgievski2016automated}. In particular, there are three dimensions of uncertainty: (1) unexpected events, which are the events that happen in exceptional and unpredictable situations, (2) actions contingencies, which can be either failures or timeouts, and (3) partial observability, which refers to the imperfectness and incompleteness of information about environmental states. The first dimension is related to the random sources of uncertainty. The second and third dimensions, however, are relevant to all uncertainty sources in our categorisation. In particular, the sources of uncertainty, internal or external, can result in incomplete or imperfect information. Similarly, the failure of actions can be relevant to internal sources caused by the unreliability of the agent, or external sources, random or regular. 

\subsection{Effects of Uncertainty on Action Costs}
\label{subsec:effects-uncertainty-action-costs}

Performing actions changes the state of the environment and can entail a cost. Costs are typically incurred in terms of money, time, fuel, or effort. In the world of certainty, these costs are deterministic. However, in the presence of uncertainty, actions have uncertain costs. In particular, the exact costs of performing actions are not necessarily known with certainty at the time of planning because, most often, actions are not guaranteed to have the same execution cost every time they are executed. In other words, actions have variable costs. To capture this fact, we call actions in uncertain domains \emph{cost-variable actions}.

The effects of uncertainty, i.e., the variability of action costs, can also be categorised based on the sources of uncertainty, as illustrated in Figure~\ref{fig:uncertainty-sources}. When uncertainty exists because of random sources, the variability of costs is unpredictable, which means it cannot be modelled in advance or dealt with during offline planning. This kind of cost variability corresponds to Level~5 of the uncertainty taxonomy shown in Figure~\ref{fig:uncertainty-taxonomy}. On the other hand, when uncertainty is caused by regular sources, the variability of costs can be further described by two categories. The first category corresponds to Level~2 and Level~3 of the uncertainty taxonomy, where the costs and their probability distribution are either known or can be statistically inferred. We call actions in this category \emph{risk-inducing actions}. We call the domains that contain this kind of action \textit{risk-involving domains}. The second category corresponds to Level~4 of the uncertainty taxonomy and includes actions that have a probability distribution representing the decision-maker's beliefs. We call actions from this category \emph{uncertainty-inducing actions}. Similarly, we call the domains that contain this kind of action \textit{uncertainty-involving domains}. This categorisation of cost variability is applied to actions that have a single effect on the environment or multiple uncertain effects. In other words, an action can have either a deterministic effect that always changes the state in one way but with variable costs or multiple uncertain effects with variable costs.
\section{A Framework of HTN Planning with Uncertainty and Risk}
\label{sec:uncertainty-htnplanning}

The foundations of HTN planning with uncertainty and risk require a formal framework of HTN planning upon which one can build and incorporate the concepts of uncertainty and risk. Two models of HTN planning exist that differ in the search space in which the planning process operates and in the representation of some \textit{hierarchical constructs}, like \textit{methods}. These models are plan-based HTN planning and state-based HTN planning~\cite{georgievski2015htn,georgievski2015coordinating} (or, similarly, decomposition-based HTN planning and progression-based HTN planning in~\cite{bercher2019survey}). In plan-based HTN planning, the search space consists of \textit{task networks}. The search starts at an initial task network, and each decomposition of a \textit{compound task} creates a new task network. Decomposing compound tasks is repeated until all compound tasks are decomposed, i.e., a primitive task network that consists of only \textit{primitive tasks} is reached. Each task network in the search space is considered a partial plan. The solution is then a linearisation of the primitive task network. 

Having partial plans as search nodes potentially leads to a more compact---but more complex---representation of the search space compared to state-based planning. However, since primitive tasks are not executed during planning, and thus the \textit{state} is not tracked in this model, the planner does not have information about the current state of the world during planning. Conversely, in state-based HTN planning, the search space consists of a subset of the state space restricted by task decompositions. The search starts at an initial state with an empty plan. The goal is to compute a plan by searching for a state that accomplishes the initial task network. During the search, when encountering a compound task, the task is decomposed if there is an \textit{applicable} method, i.e., a method that can decompose this compound task and has its \textit{preconditions} satisfied in the current state. These preconditions define the state of the world that must exist before the method can be used to decompose a task into subtasks. Then, the task decomposition continues on the next decomposition level but in the same state. If the task is primitive, it is executed, removed from the task network, and added to the plan. The search then continues into a successor state.

As for the representation of hierarchical constructs, state-based HTN planning uses method preconditions that should be satisfied in the current state during search in order to decompose compound tasks. On the contrary, in plan-based HTN planning, methods usually have constraints instead of preconditions. The constraints restrict the order of the tasks resulting from the decomposition, the bindings of task network variables, and the order of tasks and state \textit{predicates}.

As providing a formalism for both models of HTN planning would unnecessarily complicate our presentation, without loss of generalization, we choose to present a formal framework for plan-based HTN planning as a widely-used model in the research of HTN planning. For a formalism of state-based HTN planning, we refer to our previous work~\cite{georgievski2015htn}.

\subsection{Classical HTN Planning}
\label{subsec:classical-htn-planning}
An \textit{HTN planning problem} is a 3-tuple $P=\langle s_0, tn_0,D \rangle$, where $s_0$ is the \textit{initial state}, $tn_0$ is a task network, called \textit{initial task network}, and $D$ is a planning domain consisting of a set of \textit{operators} and \textit{methods}. A \textit{state} $s \in 2^Q$ is a set of \textit{ground predicates} and follows the closed-world assumption---the state consists of all and only true predicates. A predicate, which evaluates to true or false, consists of a predicate symbol $p \in P$, where $P$ is a finite set of predicate symbols, and a list of terms $\tau_1, \dots, \tau_k$. A {\em term} is either a constant symbol $c \in C$, where $C$ is a finite set of constant symbols, or a variable symbol $v \in V$, where $V$ is an infinite set of variable symbols. We can now define a predicate as \textit{ground} if its terms contain no variable symbols.

A {\em task network} is a triple $\langle T_n, \varphi , \psi\rangle$, where $T_n$ is a set of tasks, $\varphi: T_n \rightarrow T_n'$ is a function that labels tasks with task names to enable identifying uniquely many occurrences of a task name in the same task network, and $\psi$ is a set of constraints over $T_n$ that restricts variable bindings, denoted as $\mapsto \subseteq V \times V \cup V \times C$, task orderings, denoted as $\prec \  \subseteq T \times T$, and the ordering between tasks and state predicates, denoted as $\vdash_{\prec} \   \subseteq T \times Q \cup Q \times T \cup T \times Q \times T$~\cite{georgievski2015htn}. 

A task can be either primitive $t_p(\tau_p) \in T_p$ or compound $t_c(\tau_c) \in T_c$, where $\tau_p$ and $\tau_c$ are lists of terms, and $t_p$ and $t_c$ are the names of primitive task and compound task, respectively. A task is {\em primitive} if it can be accomplished directly by an operator $o=\ \tuple{pt(o),pre(o),\mathit{eff(o)}}$, where $pt(o), pre(o),$ and $\mathit{eff(o)}$ are the operator's parameterised name, precondition, and effects, respectively. The operator's parameterised name $pt(o) = t_p(\tau_p)$ is identical to the primitive task that can be executed by this operator. There is a one-to-one mapping between operators and primitive tasks. A task is said to be {\em compound} if it must be decomposed into further primitive and compound sub-tasks using a method $m=\tuple{ct(m),tn(m)}$, where $ct(m)$ and $tn(m)$ are the method's parameterised name and the method's task network, respectively. The method's parameterised name is equal to the compound task that can be decomposed by this method, i.e.,  $ct(m)= t_c(\tau_c)$. One compound task can be decomposed by multiple methods.

When the terms of a task are all constant symbols, we call it a \textit{ground task}. We use $\overline{t_p}(\tau_p)$ and $\overline{t_c}(\tau_c)$ to denote ground primitive and compound tasks, respectively. Similarly, we use $\overline{o}$ to denote a ground operator, in which $\overline{pt}(o)$ is identical to the corresponding ground primitive task, and $\overline{pre}(o) \in 2^Q$ and $\overline{eff}(o) \in 2^Q$ are sets of ground predicates. We denote the set of bindings of $\mathit{m}$ according to constraints $\psi$ by $\mathit{\overline{m}|_{\psi}}$. We call this set \textit{ground method}, where $\overline{ct}(m)= \overline{t_c}(\tau_c)$ is a ground compound task and $\forall t_i(\tau_i) \in tn(m)$, $\tau_i \in C$ with respect to $\psi$. Lastly, we use $\overline{D}|_P=(\overline{O},\overline{M}$) to denote the ground version of domain $\mathit{D}$ according to the initial state $s_0$ and initial task network $tn_0$ in $\mathit{P}$, where $\overline{O}$ and $\overline{M}$ are sets of ground operators and ground methods, respectively, and all the terms in the domain are constant symbols from the $s_0$ and $tn_0$.

In plan-based HTN planning, the planning process starts by decomposing the initial task network into a refined task network and continues by repeatedly decomposing tasks from a newly refined task network until (1) the task network is reduced to a primitive task network that constitutes a solution to the planning problem, i.e., until all compound tasks are decomposed, (2) there is a linearisation of the tasks in the primitive task network such that this linearisation is compatible with $\psi$, and (3) the corresponding sequence of operators is executable in the initial state.

Adopting the formalism of plan-based planning from~\cite{georgievski2015htn}, we formally define the decomposition, executable task network, and solution as follows.

\begin{definition}[Decomposition]
\label{def:decomposition-plan_based}
    Let $m = \langle ct(m), pre(m), tn(m) \rangle$ be a method and $tn_c = \langle T_c, \varphi_c, \psi_c \rangle$ be a task network. Method $m$ decomposes $tn_c$ into a new task network $tn_n$ by replacing task $t$, denoted as $tn_c {\xrightarrow[\text{t,m}]{}}_D  tn_n$, if and only if $t \in T_c, \varphi_c(t) = ct(m)$, and there exists a task network $tn' = \langle T', \varphi', \psi' \rangle$ such that $tn' \equiv tn(m)$ and $T' \cap T \neq \phi$, and  
    \begin{align*}
        tn_n &:= ((T_c \setminus \{t\}) \cup T', \varphi_c \cup \varphi', \psi_c \ \psi' \cup \psi_D), where \\
        \psi_D &:= \{(t_1,t_2) \in T_c \times T' | (t_1,t) \in \prec_c\} \cup \{(t_1,t_2) \in T' \times T_c | (t,t_2) \in \prec_c\} \cup \\
       &\{(q,t_1) \in Q \times T' \ | \  (q,t) \in \vdash_{\prec c}\} \  \cup  \{(t_1,q) \in T' \times Q \ | \  (t,q) \in \vdash_{\prec c}\} \  \cup \\
       &\{ (t_1,q,t_2) \in T' \times Q \times T' | (t,q,t_2) \in \vdash_{\prec c}\}
    \end{align*}     
\end{definition}

A task network $tn= \langle T, \varphi, \psi \rangle$ is isomorphic to $tn' = \langle T', \varphi', \psi' \rangle$, denoted as $tn \equiv tn'$, if and only if there exists a bijection $\beta: T \rightarrow T'$, such that
\begin{itemize}
    \item $\forall t,t' \in T$ it holds $(t,t') \in $ if and only if $(\beta(t),\beta(t')) \in \prec'$,
    \item $\forall v_1,v_2 \in V$ and $c \in C$ it holds $(v_1,v_2) \in \mapsto$ or $(v_1, c) \in or (v1, c) \in \mapsto$ if and only if there exist $v'_1,v'_2 \in V$ and $c' \in C$ such that $v_1 = v_1', v_2 = v_2'$ and $(v_1',v_2') \in \mapsto'$ or $v_1 = v_1',c = c'$ and $(v_1',c) \in \mapsto'$,
    \item $\forall t,t' \in T$ and $q \in Q$ it holds $(t,q) \in \vdash_{\prec}$ or $(q,t) \in \vdash_{\prec}$ or $(t,q,t') \in \vdash_{\prec}$ if and only if $(\beta(t),q) \in \vdash'_{\prec}$ or $(q,\beta(t)) \in \vdash'_{\prec}$ or $(\beta(t),q,\beta(t')) \in \vdash'_{\prec}$,
\end{itemize}

and $\varphi(t) = \varphi'(\beta(t))$.

Given an HTN planning problem $P$, $tn_c {\rightarrow}_D^*  tn_n$ indicates that $tn_n$ results from $tn_c$ by an arbitrary number of decompositions using methods from $M$.

To formally define the solutions in plan-based HTN planning, we need to define the notion of \textit{executable task networks}.

\begin{definition}[Executable task network]
\label{def:exe-taskNetwork}
   Given an HTN planning problem $P$, $tn = \langle T, \varphi, \psi \rangle$ is executable in state $s$ if and
only if it is primitive and there exists a linearisation of its tasks $t_1, \ldots,t_n$ that is compatible with $\psi$, i.e., it is consistent with the set of constraints $\psi$, and the corresponding sequence of operators $\varphi(t_1), \ldots, \varphi(t_n)$ is executable in $s$.
\end{definition}

A solution is given by the following definition.

\begin{definition}[Solution]
\label{def:solution-plan_based}
   Let $P$ be an HTN planning problem. A task network $tn_s$ is a solution to $P$ if and only if $tn_s$ is executable in $s_0$ and $tn_0 {\rightarrow}_D^*  tn_s$.
\end{definition}

\subsection{HTN Planning with Cost-Variable Operators}
\label{subsec:htn-planning-cost-variable}
Postulating that the variability of action costs in real-world domains constitutes a source of uncertainty and risk as intrinsic properties of these domains entails the need to transfer such knowledge to planning. This requires the concepts of risk and uncertainty to be explicitly modelled and taken into account when generating plans.

Variability of costs can be modelled as a probability distribution over the possible costs of operators. To be more specific, variability of costs can appear in two types of actions, risk-inducing and uncertainty-inducing actions (see Section~\ref{subsec:effects-uncertainty-action-costs}). Then, to model risk-inducing actions, one needs to encode the probability distribution obtained from past experience or statistical inference over an operator's costs. We call such operators \emph{risk-inducing operators}, which correspond to risk-inducing actions. These operators can be mapped to Level~2 and Level~3 of the uncertainty taxonomy in Figure~\ref{fig:uncertainty-taxonomy}. Whereas to model uncertainty-inducing actions, the probability distribution over an operator's costs is modelled as a representation of the agent's beliefs. We call such operators \emph{uncertainty-inducing operators}, which correspond to uncertainty-inducing actions. Similarly, these operators are mapped to Level~4 of the uncertainty taxonomy.

We now extend the classical HTN planning framework to account for variable-cost actions. To achieve this, we define operators with probabilistic effects and costs.

\begin{definition} [Cost-Variable Operator]
  A cost-variable operator $o$ is defined as a tuple $o=\ \tuple{pt(o),pre(o),\mathit{eff(o)}, c(o)}$, where $pt(o)$, and $pre(o)$ are defined as in the operators definition in the classical HTN planning framework (Section~\ref{subsec:classical-htn-planning}), and $\mathit{eff(o)}$, and  $c(o)$ are tuples that represent the effects and the costs of the operator, respectively and are defined as follows. $\mathit{eff(o)} = \tuple{(p_{1}(o),\mathit{eff_{1}(o)}), (p_{2}(o),\mathit{eff_{2}(o)}), \cdots, (p_{n}(o),\mathit{eff_{n}(o)})}$ and $c(o) = \tuple{(p_{1}(o),c_{1}(o)),(p_{2}(o),c_{2}(o)), \cdots, (p_{n}(o),c_{n}(o))}$, such that
  $\mathit{eff(o)}$ are the variable effects of the operator and $c(o)$ is the variable costs of the operator, such that

  \begin{itemize}
  	\item $\mathit{eff_i(o)}$ and $c_i(o)$ are the $i$-th effect with its corresponding cost, respectively
  	\item $p_i(o) \in \mathbb{R}^{+}$ is a probability of the $i$-th effect and its corresponding cost, such that $\forall n>0 \ \forall i \in [1,n],\ 0 \leq p_i(o) \leq 1$
  	\item $\sum\limits_{i=1}^n p_i(o) = 1$
  	\item $c_i(o) < 0$ is an unbounded negative cost function.
  \end{itemize}
\end{definition}

Since we are dealing with a cost-based domain, we assume costs have negative values in our definition. However, the definition can be easily generalised to model costs as positive values, to have rewards instead of costs, or to model any function of both rewards and costs. Moreover, the definition allows defining the degenerate cases of having single-effect operators with a probability distribution of costs and single-effect operators with a single cost. We define a \textit{ground cost-variable operator} as a ground operator with variable effects and costs.

We differentiate four types of cost functions depending on sources/factors influencing action costs. The cost function can be (1) \emph{external}, denoted as $c^{ie}(o)$; (2) \emph{state-dependent}, denoted as $c^{is}(o)$; (3) \emph{constant}, denoted as $c^{ic}(o)$; or (4) external and state-dependent, i.e., a hybrid function denoted as $c^{ies}(o)$. Consider, for example, an electric vehicle domain, where an electric vehicle can navigate roads and charge its battery at charging stations. Consider the action of charging the vehicle, where the action's cost represents the price paid for charging the vehicle. This cost can be calculated using an external function $c^{ie}$ that depends on several external factors not necessarily modelled directly in the domain. These factors may include the charging station's price, the market electricity prices, and the applicable taxes. We can, however, model the domain such that whenever the vehicle charges, it charges the same amount at the same price, where the charging price can be calculated using a constant function $c^{ic}(o)$ that incurs the same price every time the charging action is executed. For different scenarios, we can calculate the cost of charging based on the current charging state of the vehicle's battery, the current vehicle's position, and how far it is from the destination. Thus, we use a state-dependent cost function $c^{is}(o)$. Lastly, the charging cost can be calculated based on state-dependent information in addition to the external factors. In that case, we use a hybrid function $c^{ies}(o)$. In a ground domain $\overline{D}|_P=(\overline{O},\overline{M})$, some variables in the state-dependent or hybrid cost functions can get assigned constant values from the initial state and the initial task network. Take, for example, an operator that moves the electric vehicle between locations. Assuming that the cost of this operator is the time needed by the vehicle to move. This time can depend on state-dependent knowledge, such as the vehicle's position and the distance between the two locations, and on external factors, such as the weather conditions and the traffic between the two locations. The operator's cost is calculated using a hybrid cost function $c^{ies}(o)$. When grounding the domain, the state-dependent variables in the hybrid cost function get their constant values depending on the specific length of the road between the two locations, which should be given in the initial state.

\subsection{HTN Planning Domain as a Graph}

Analysing and characterising an HTN planning domain model can be cumbersome due to the rich domain knowledge included in tasks and their relationships. To facilitate the analysis and characterisation of an HTN planning domain, we resort to using graphs as a modular, flexible, and effective tool for representing versatile data structures, including hierarchical ones, and analysing and manipulating data constructs and their relationships. Therefore, we define an HTN planning domain with cost-variable operators as a graph, and we build our definition over the task decomposition graph presented in~\cite{elkawkagy2012improving}. In a task decomposition graph, vertices represent both tasks and methods, and the directed edges go from a task vertex to all method vertices that can decompose it and from a method vertex to all task vertices in its task network. In our definition, the task decomposition graph follows a similar definition but represents the parameterised domain instead of the parameter-free (i.e., ground) domain, and it also incorporates cost-variable operators. We refer to the resulting task decomposition graph as \textit{Cost-Variable Task Decomposition Graph (CV-TDG)}.

\begin{definition} [Cost-Variable Task Decomposition Graph]\label{def:CVTDG}
	Let $D=\ \tuple{O,\ M}$ be an HTN planning domain, where $O$ is a set of cost-variable operators. The directed graph $G=\ \tuple{V_{TC},V_{TP},V_M,E_{TC\to M},E_{M\to TC},E_{M\to TP}}$, where $V_{TC}$ is a set of compound task vertices, $V_{TP}$ is a set of primitive task vertices, $V_M$ is a set of method vertices, and $E_{TC\to M}, E_{M\to TP}$, and $E_{M\to TP}$ are sets of edges, is a Cost-Variable Task Decomposition Graph (CV-TDG) of D if and only if:
	\begin{enumerate}
		\item $\forall m=\tuple{ct(m),tn(m)} \in M$:
		\begin{itemize}
			\item $v_{tc} \in V_{TC}$ such that $v_{tc} = ct(m)$
			\item  $v_m \in V_M$ such that $v_m = m$
			\item $(v_{tc},v_m) \in E_{TC\to M}$
                \item $\forall t_p(\tau_p) \in tn(m) | ~\exists o=\tuple{pt(o),pre(o),eff(o),c(o)}\in O \land pt(o) = t_p(\tau_p)$: $(v_m,v_{tp}) \in E_{M\to TP}$ such that
                \begin{itemize}
                    \item $v_{tp}=o ~\land $
                    \item $\exists c: c: v_{tp} \rightarrow c(o)$
                \end{itemize}  
                \item $\forall t_c(\tau_c) \in tn(m)$:  $ (v_m,v_{tc}) \in E_{M\to TC}$ such that $v_{tc}=t_c(\tau_c)$
		\end{itemize}
            \item $G$ is minimal, such that (1) holds.
	\end{enumerate}
\end{definition}

\begin{figure}[t!]
	\centering
	\includegraphics[width=0.75\textwidth]{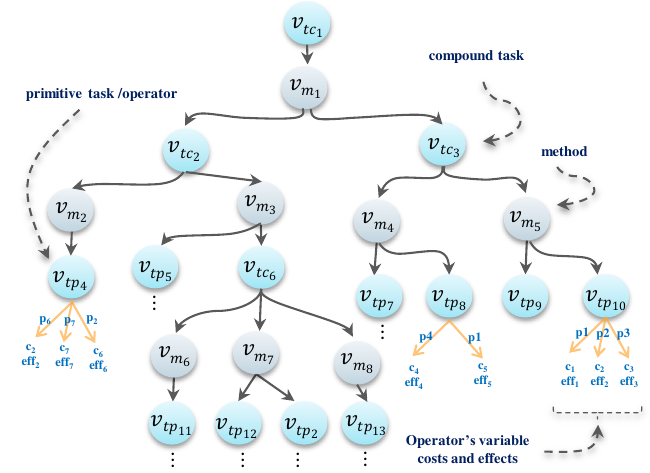}
	\caption{Cost-Variable Task Decomposition Graph (CV-TDG) representation of an abstract domain description. 
 }
	\label{fig:abstract-domain}
\end{figure}

An example of CV-TGD for an abstract HTN domain with four compound tasks and ten primitive tasks is shown in Figure~\ref{fig:abstract-domain}. The graph has three types of vertices labelled with $v_{tc_i}$, $v_{tp_j}$, and $v_{m_k}$ to represent compound tasks, primitive tasks, and methods vertices, respectively. Costs and effects variability is illustrated under the corresponding primitive tasks by probabilities $p_l$, effects $\mathit{eff_r}$, and costs $c_r$.

A special case of the CV-TDG is the Ground CV-TDG (GCV-TDG), which is constructed by binding the variables in the HTN planning domain according to the initial state and initial task network part of the planning problem. The GCV-TDG is defined as follows.

\begin{definition} [Ground Cost-Variable Task Decomposition Graph]\label{def:GCVTDG}
	Let $P=\langle s_0, tn_0,\overline{D}\rangle$ be an HTN planning problem, where $\overline{D}|_P=(\overline{O},\overline{M}$) is a ground HTN planning domain, and $\overline{O}$ and $\overline{M}$ are sets of ground cost-variable operators and ground methods, respectively. The directed graph $G=\ \tuple{V_{\overline{TC}},V_{\overline{TP}},V_{\overline{M}},E_{\overline{TC} \to \overline{M}},E_{\overline{M} \to \overline{TC}},E_{\overline{M} \to \overline{TP}}}$, where $V_{\overline{TC}}$ is a set of ground compound task vertices, $V_{\overline{TP}}$ is a set of ground primitive task vertices, $V_{\overline{M}}$ is a set of ground method vertices, and $E_{\overline{TC} \to \overline{M}}, E_{\overline{M} \to \overline{TP}}$, and $E_{\overline{M} \to \overline{TP}}$ are sets of edges, is a Ground Cost-Variable Task Decomposition Graph (GCV-TDG) of $\overline{D}|_P$ if and only if:
	\begin{enumerate}
		\item $\forall \overline{m}|_{\psi}$, such that $\overline{m}=\tuple{ct(m),tn(m)} \in \overline{M}$:
		\begin{itemize}
			\item $v_{\overline{tc}} \in V_{\overline{TC}}$ such that $v_{\overline{tc}} = \overline{ct}(m)$
			\item $v_{\overline{m}} \in V_{\overline{M}}$ such that $v_{\overline{m}} = \overline{m}|_{\psi}$
			\item $(v_{\overline{tc}},v_{\overline{m}}) \in E_{\overline{TC}\to \overline{M}}$
			\item $\forall \overline{t_p}(\tau_p) \in \overline{tn}(m) | \exists \overline{o}=\tuple{\overline{pt}(o),\overline{pre}(o),\overline{eff}(o),c(o)}\in \overline{O} \land \overline{pt}(o) = \overline{t_p}(\tau_p)$:  $ (v_{\overline{m}},v_{\overline{tp}}) \in E_{\overline{M} \to \overline{TP}}$ such that $v_{\overline{tp}}=\overline{o} \land \exists c: v_{\overline{tp}} \rightarrow c(o)$ 
			\item $\forall \overline{t_c} \in \overline{tn}(m)$:  $ (v_{\overline{m}},v_{\overline{tc}}) \in E_{\overline{M} \to \overline{TC}}$ such that $v_{\overline{tc}}=\overline{t_c}$		
		\end{itemize}	
		\item $G$ is minimal, such that (1) holds.
	\end{enumerate}
\end{definition}

\section{Risk-Aware HTN Planning}
\label{riskAware-htnplanning}

To enable HTN planning to address risk, foundations must be established that define the types of risks and incorporate a decision theory to support action choices. Initially, the types of planning decisions that may occur in HTN planning and their susceptibility to risk influences are explored. Subsequently, concepts from decision theory are examined, providing a framework for decision-making when multiple choices exist in the presence of risk. Particular attention is given to the concept of risk-aware decision-makers and the risk attitudes of HTN planning agents.

The proposal and discussion focus on planning in domains where risk-inducing actions arise from internal and external sources of uncertainty. This implies that the vNM theorem is applicable to these planning problems. To streamline the discussion, attention is concentrated on risk-inducing actions that have a single effect.

\subsection{Risk in Planning Decisions}
\label{subsec:risk-planning-decisions}

An agent can make three types of planning decisions when solving HTN planning problems. The first type represents the choice of a method to use when decomposing a compound task. The second is about the choice of values that can be assigned to variables in HTN domain constructs, or \textit{bindings}. These values are restricted by the predicates in the initial state and tasks in the initial task network. The third is about deciding the order in which compound tasks in task networks are chosen for decomposition and the order of operators in the plan. The last type of choice is applicable in plan-based planning since there is a partial order between tasks. In particular, in the partially ordered HTN planning, the new tasks resulting from the decomposition can be ordered in parallel whenever possible, with respect to the constraints. These planning choices influence the outcome of the planning process~\cite{georgievski2015htn}.

\begin{figure}[!t]
	\centering
	\includegraphics[width=\textwidth]{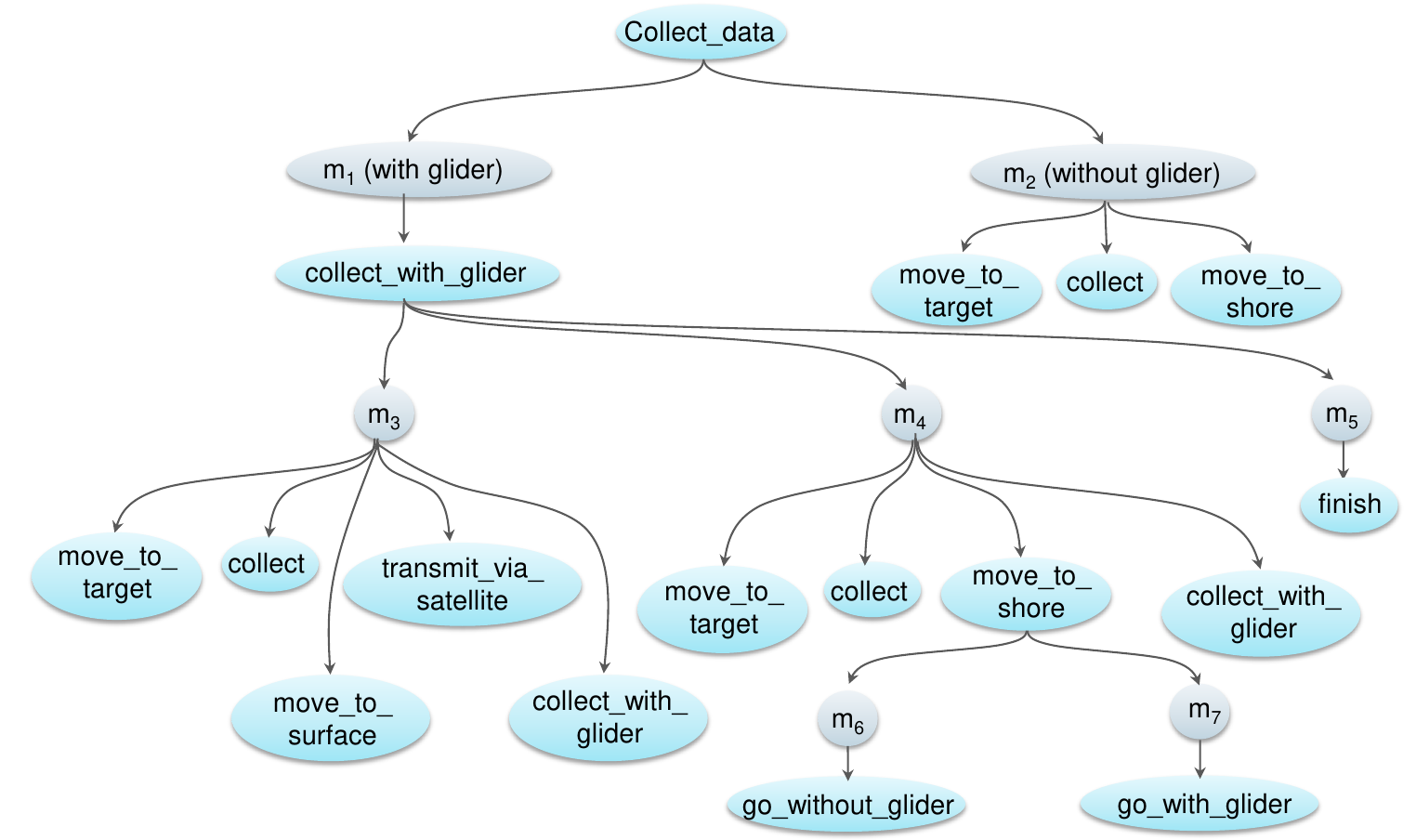}
	\caption{HTN model for the marine domain. 
 }
	\label{fig:glider-domain}
\end{figure}

Let us briefly illustrate a model of a marine domain to be able to exemplify these planning decisions. \ref{apx:marine} provides a detailed description of the marine domain. The model can be populated in two ways: a diver can dive alone, or a glider can accompany the diver. In the first case, the diver moves to the target, collects data, and moves to the shore. In the second case, two further options to collect data are available. The glider moves with the diver to the target, collects part of the data, moves to the surface, and transmits the data. The task of data collection should be executed again if there is still data that has not been collected. Of course, this is only possible if the glider has enough remaining power available. In the second option, the diver should move to the target, collect data, move to the shore, and repeat these steps until data is available for collection. When the diver dives back to the shore, he can go alone or go with the glider to guarantee higher safety. Once all the data is collected, the task is considered complete. Figure~\ref{fig:glider-domain} shows an HTN representation of this domain model, where grey nodes $m_1, \ldots, m_7$ represent methods, and blue nodes represent primitive and compound tasks.

Going back to the planning decisions, if the planning agent chooses the option that the diver should do a solo dive without the glider, the solution would be different than if the choice is to go with the glider. That is, if the planning agent chooses the other way, it faces another planning decision, i.e., the choice between methods $m_3$ and $m_4$, where each decision would eventually lead to a different plan. Now, assume that we want to solve the problem where we have different gliders that can accompany the diver, and these gliders have different amounts of available power. The choice of which glider to accompany the diver -- binding the glider variable to a specific glider -- may also affect the computed plan. Imagine now that the tasks in $m_4$'s task network are all compound and partially ordered. If the planning agent chooses the task \textit{move\_to\_shore} to decompose first, this might eventually result in a different plan if the first chosen task to decompose was the \textit{move\_to\_target} task.

Bringing risk into perspective in HTN planning by means of risk-inducing operators requires evaluating risk in each planning choice of every type when solving the HTN planning problems. To illustrate what we mean, say now the driver needs 10 minutes for sure to return to the shore with the glider. This means that going to the shore with the glider results in guaranteed time, i.e. cost and does not involve taking risks. However, when the diver dives back by himself, it will take him 2 minutes to reach the shore with a probability of 80\%, and 20 minutes with a probability of 20\%. Thus, going alone involves more risk of being delayed than going with the glider. Let us assume that the actions \textit{collect}, \textit{move\_to\_surface}, \textit{transmit\_via\_satellite}, and \textit{move\_to\_target} when the glider is with the diver have certain times. However, similar to the action \textit{go\_without\_glider}, the action \textit{move\_to\_target} when the diver is alone involves taking risks. The existence of the \textit{go\_without\_glider} risk-inducing action characterises the choice between methods $m_6$ and $m_7$ as risk-involving. In particular, it is always safer to decompose the \textit{collect\_with\_glider} compound task with $m_3$ since choosing $m_4$ involves taking risks if $m_6$ is chosen to decompose the \textit{move\_to\_shore} compound task. Similarly, the existence of the \textit{move\_to\_target} risk-inducing action also makes the choice between $m_1$ and $m_2$ risk-involving. This shows an example of the influence of risk on the first type of planning decisions, i.e., method choices. To illustrate how the existence of risk can influence the decision of variable binding, i.e., the second type of planning decisions, assume that the driver can go to several locations on the shore to transmit the collected data. Going to each location involves a different amount of risk based on the state of the path (e.g., the flow and obstacles on the way to the shore) that the driver will take to reach the location and based on the distance between the current location of the diver and all transmission locations. This makes the costs of \textit{go\_without\_glider} and \textit{go\_with\_glider}, for example, hybrid, where the different bindings of the variable representing the transmission location can lead to different amounts of time needed to reach the location, i.e., some bindings can involve more risks than others. Similar reasoning can be applied to choices of task orderings.

\subsection{Risk Awareness}
The choices made during planning influence the outcome of the planning process---the plans. If the quality of plans is not important, these choices can be made non-deterministically. Otherwise, we need to make informed planning decisions, even in the presence of risk. If we consider that rational decision-makers aim at maximising their expected utility as a quality criterion (see Section~\ref{sec:uncertainy-risk}) and apply this approach to HTN planning with risk, it turns out that the planning agent should make planning choices that maximise the expected utility of the resulting plan.

The agent can maximise the plan's expected utility by evaluating the expected utility obtained from each planning decision. To enable this, we follow a bottom-up approach by expressing the planning agent's risk attitude with respect to the different outcomes of operators and using this to make choices for all planning decisions. To express the risk attitude adopted by the planning agent when solving planning problems, we employ utility functions. In particular, we assign a utility function to the planning agent to express its preferences over the different outcomes of operators. This enables the planning agent to compute the expected utilities of the operators. Then, the main process is evident: these expected utilities are used to make all planning decisions. In particular, the expected utilities are propagated to methods to allow making informed choices that maximise the plan's expected utility. That is, the agent chooses the method, among all applicable ones used to decompose a compound task, that eventually leads to operators with the highest expected utility.\footnote{Our approach is similar to how risk and utilities are modelled and reasoned about in utility theory using decision trees~\cite{crundwell2008decision}. In decision trees, the leaves can be risk-inducing nodes that hold the possible outcomes, and the decisions are made on higher levels in the tree. In order to choose options that maximise the expected utility of the decision maker, expected utilities are computed for leaves and then propagated to decision nodes on higher tree levels~\cite{cappello2016expected}.} Computing the expected utilities of operators enables making informed decisions for variables binding, i.e., the second type of planning decisions. In particular, if the operator cost functions are state-dependent or hybrid, when binding variables, the planning process should consider the binding that maximises the plan's expected utility. Going back to the marine domain explained in Section~\ref{subsec:risk-planning-decisions}, when having multiple transmission locations and/or multiple gliders, the planning process should consider using the variable bindings of method and task terms that lead to operators with the maximum expected utility since the different bindings of operator terms can lead to different operator costs. 

We formally define HTN planning problems cognizant of risk, based on the necessary ingredients presented thus far, as follows.

\begin{definition}[Risk-aware HTN Planning]
\label{def:risk-aware-htn-planning}
A risk-aware HTN planning problem is a 4-tuple $P_r=\tuple{s_0,tn_0,D,U}$, where 
\begin{itemize}
    \item $s_0$ is the initial state,
    \item $tn_0$ is the initial task network,
    \item $D=\tuple{O,M}$ is a risk-involving planning domain consisting of cost-variable operators $O$ and a set of methods $M$, and
    \item $\forall o \in O, U: c(o) \rightarrow \mathbb{R}$ is a utility function that expresses the risk attitude in terms of operator costs.
\end{itemize}
A plan $\pi$ is a solution to $P_r$ if and only if $\pi$ complies with Definition~\ref{def:solution-plan_based} and has a maximum expected utility $EU(\pi)$ computed using $U$ to reflect the risk attitude of the risk-aware HTN planning agent, i.e., $EU(\pi) = max \{EU(\pi_i): i = 0 \ldots n$\}, where $n$ is the number of all primitive task networks that are applicable in $s_0$ and produced by repeatedly decomposing $tn_0$.
\end{definition}

Note that for risk-aware state-based HTN planning, the solution is defined similarly to the previous definition, but it complies with the solution of the classical state-based HTN planning (see Definition 13 in~\cite{georgievski2015htn}) instead of Definition~\ref{def:solution-plan_based}, which is for plan-based HTN planning.

\begin{definition} [Risk-aware HTN planning agent]
A risk-aware HTN planning agent is an HTN planning agent that solves risk-aware HTN planning problems $P_r$ and computes plans with the maximum expected utility by making informed planning choices that reflect its risk attitude expressed by a utility function.
\end{definition}

\subsection{Risk Attitudes}

As with other decision-makers, risk-aware HTN planning agents can show various attitudes toward risk in different decision-making situations. Depending on situation dynamics, we categorise risk-aware HTN planning agents following the two patterns presented in Section~\ref{sec:uncertainy-risk}. The first category includes planning agents with a static risk attitude, and the second one consists of agents with a dynamic risk attitude. These two categories differ in the type of utility functions that planning agents use and how they evaluate the expected utility of individual planning choices. While a planning agent with a static risk attitude accounts only for the increase and/or decrease in the resource consumption (not in the amount of resource itself) when making planning decisions, a planning agent with a dynamic risk attitude makes planning decisions while considering the amount of the resource itself.

\subsubsection{Static Risk Attitudes}

An agent's static attitude does not change during planning. It is given in advance (e.g., encoded by a domain expert) and can be selected based on the degree of risk tolerance required for each planning problem. We assume agents have unlimited resources, i.e., no limit on how much operators can consume. Clearly, agents should act rationally by making choices that contribute to the maximisation of the plan's expected utility.

To enable expressing a static risk attitude of risk-aware HTN planning agents, we define a family of utility functions denoted as $U_c$. This family includes linear and exponential functions commonly used to express risk-sensitivity~\cite{wood2015exponential,de2020risk,koenig1994risk}. \\

\begin{figure} 
	\centering
    \begin{subfigure}[b]{0.49\textwidth}
       \centering
	\includegraphics[width=\textwidth]{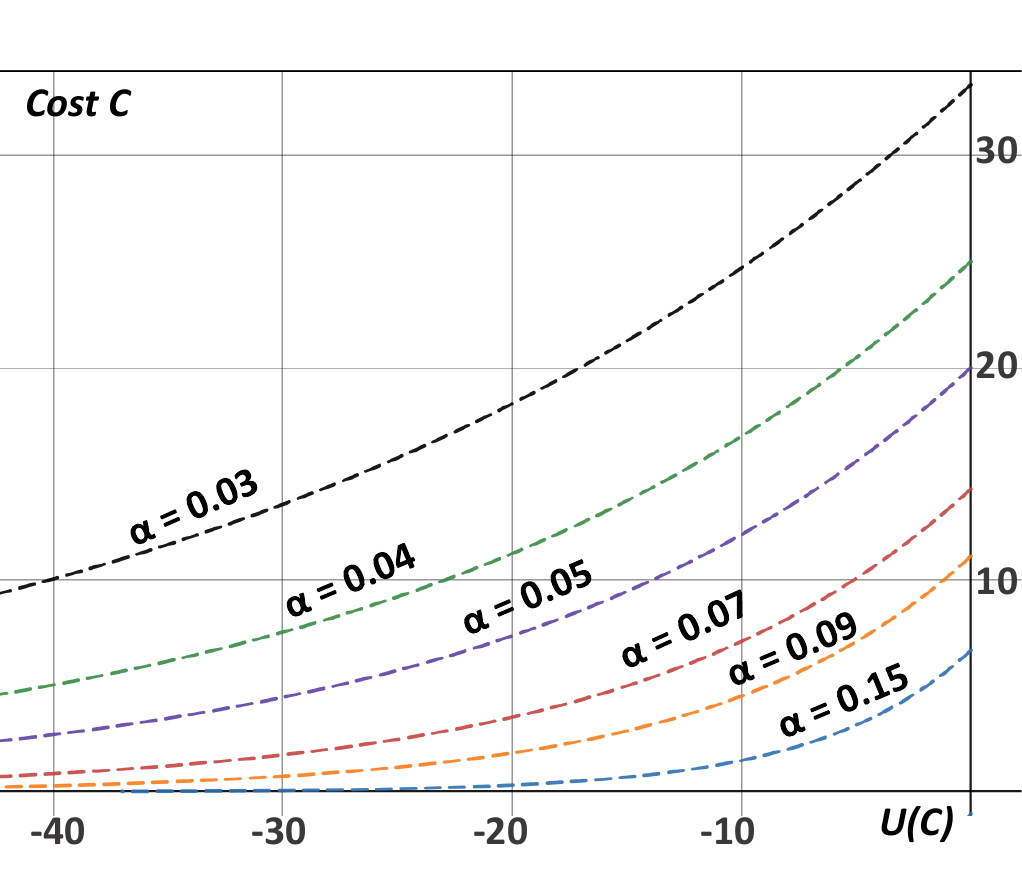}
	\caption{Utility functions of risk-seeking attitudes.}
	\label{fig:riskSeeking}
	 \end{subfigure}
	 \hfill
	 \begin{subfigure}[b]{0.49\textwidth}
    \centering
	     \includegraphics[width=\textwidth]{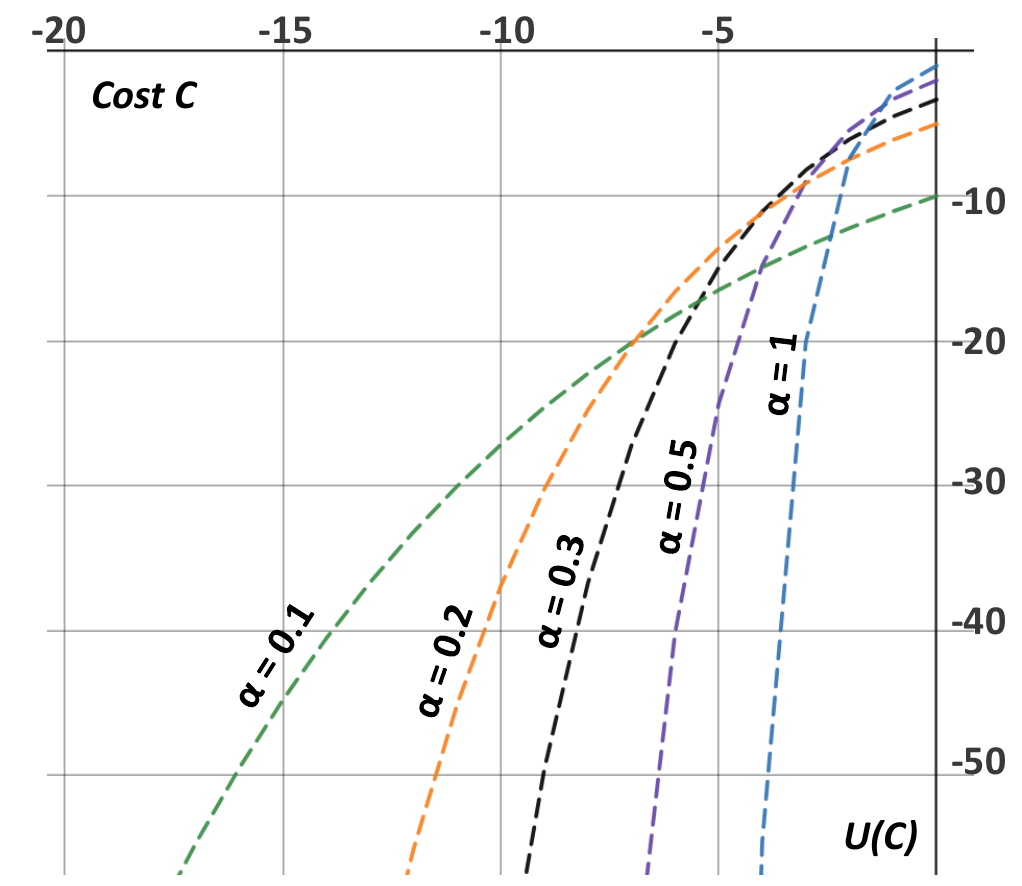}
	\caption{Utility functions of risk-averse attitudes.}
	\label{fig:riskAverse}
	 \end{subfigure}
	 \caption{An illustration of the exponential utility function for (a) risk-seeking and (b) risk-averse attitudes with varying values of the parameter $\alpha$.}
	 \label{fig:utilityFunctions}
\end{figure}

\noindent $\forall v_{tp} \in V_{TP}$, such that $v_{tp} = o = \tuple{pt(o),pre(o),\mathit{eff(o)}, c(o)}$ and $c(o) = \tuple{(p_{1}(o),c_{1}(o)), (p_{2}(o),c_{2}(o)), \cdots, (p_{n}(o),c_{n}(o))}$ and $i \in [1,n]$:

 \begin{equation} \label{equ:exponentialUtilityFunction}
     U_c(c_{i}(o)) =
        \begin{cases}
            c_{i}(o),       &  \text{if neutral}; \\ \\
            \frac{a e^{a\alpha c_{i}(o)}}{\alpha},  &  \text{otherwise}, 
        \end{cases}
\end{equation}

\noindent where:
\begin{itemize}
    \item $a$ is an attitude-determinant coefficient, and
    \item $\alpha$ is a curving coefficient driving the shape of the utility function. 
\end{itemize}
When the attitude-determinant coefficient $a$ is positive ($a > 0$), the utility function is used to express a risk-seeking attitude, while when negative ($a < 0$), the utility function expresses a risk-averse attitude. Using the curving coefficient $\alpha$, we can express a whole spectrum of risk-sensitive attitudes such that the bigger $\alpha$ is, the more risk-sensitive the agent is. For $a > 0$, the $\alpha$ parameter allows expressing a range of risk-seeking attitudes from being extremely risk-seeking such that the agent assumes that nature makes the outcomes as much suited for the agent as possible to the least degree of risk-seeking attitude. Similarly, when $a < 0$, $\alpha$ allows expressing a spectrum of risk-averse attitudes from being extremely risk-averse such that the agent assumes that nature plays against it and it hurts as much as it can to being at the least degree of risk-averse. Figures~\ref{fig:riskSeeking} and~\ref{fig:riskAverse} show examples of the exponential utility function with varying values of $\alpha$ for both the risk-seeking and the risk-averse attitudes, respectively. We see that in the utility function of the risk-seeking attitude, the utility decreases for higher operator costs but at a slower rate, i.e., the slope of the function decreases, which makes high operator costs look smaller and makes the planning agents that adopt this attitude willing to choose methods that have a high risk if it has a possibility of upside potential -- the possibility of leading to small operator costs. On the other hand, we see that the utility function for the risk-averse attitude has a downward concave curve, where the concavity increases dramatically for large operator costs (the slope increases). This gives an exaggerated negative weight to the possible large operator costs. This kind of utility function allows a risk-averse planning agent to follow an avoidance strategy by shying away from method choices that would expose it to possible large operator costs, even if such methods have the possibility of upside potential -- the possibility of leading to operators with possibly small costs.

\subsubsection{Dynamic Risk Attitudes}
\label{subsubsec:dynamic-risk-attitudes-HTN}

An agent's risk attitude is dynamic if it changes during planning. How it changes can, for example, depend on the amount of resources the agent has. A \textit{resource} is an object with a limited capacity $\bar{r}$ available for use by operators. Thus, we define a resource $R$ as a positive real value constrained by the capacity $\bar{r}$, $0<R\leq\bar{r}$.~\footnote{We focus on numeric value resources rather than binary value resources, which indicate whether a resource is free or in use~\cite{georgievski2015htn}.} For example, in the marine domain, the resource can be the onboard energy that the glider's battery has, the amount of air the glider has, the time of the mission, or any combination of those. Furthermore, we consider resources that are \textit{disposable} or \textit{consumable}---a type of resource that can be used a limited number of times until they are fully exhausted. Each time an operator is executed, the resource is decreased by an amount equivalent to the operator's cost. For example, the energy held in the glider's batteries decreases each time an operator executes an action. Notice that we focus on consumables because our work treats cost-based domains. Nevertheless, if executing some operators can result in gaining rewards instead of resource consumption, we can consider the resources that can be replenished, or \textit{renewable resources}~\cite{georgievski2015htn}.

Having a utility function that expresses a dynamic risk attitude allows a risk-aware HTN planning agent to switch its risk attitude depending on the available amount of resources. To that end, we use a one-switch utility function, which supports a single switch of the risk attitude, to define a family of utility functions that model a dynamic risk attitude, $U_d$.

\noindent $\forall v_{tp} \in V_{TP}$, such that $v_{tp} = o = \tuple{pt(o),pre(o),\mathit{eff(o)}, c(o)}$ and $c(o) = \tuple{(p_{1}(o),c_{1}(o)), (p_{2}(o),c_{2}(o)), \cdots, (p_{n}(o),c_{n}(o))}$ and $i \in [1,n]$:

\begin{equation} \label{equ:one-switch}
    U_d(R + c_j(o_i)) = R + \mathscrsfs{D} \ (\frac{- e^{- \alpha (R+ c_j(o_i))}}{\alpha})
\end{equation}

\noindent where the parameter $\mathscrsfs{D} > 0$ determines the trade-off between the risk-aversion and the risk-neutrality, $\alpha$ determines the degree of risk-aversion, and $R$ is the remaining amount of the resource.

\begin{figure}
 	\centering
 	\includegraphics[width=0.75\columnwidth]{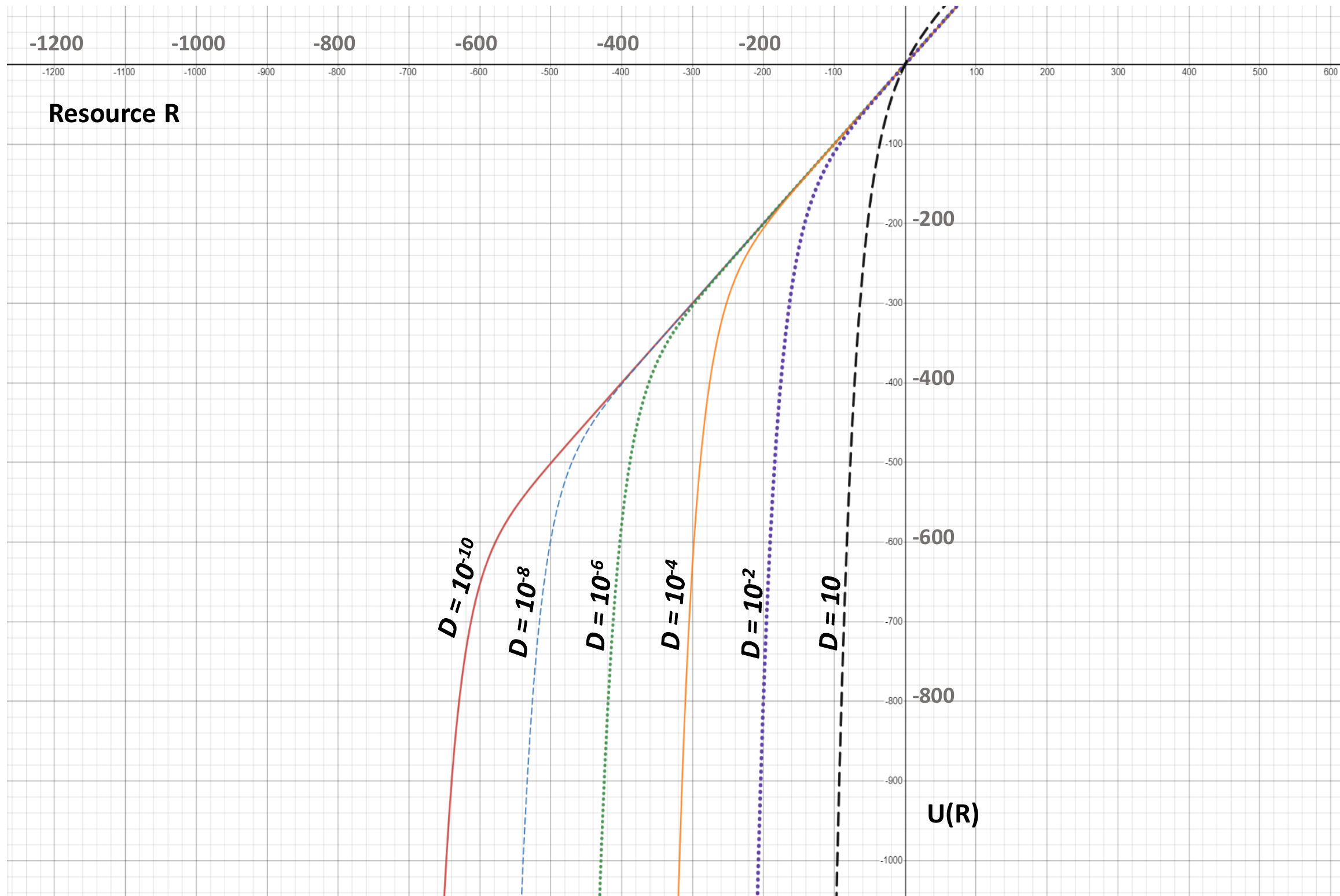}
 	\caption{One-switch utility function with varying values for $\mathscrsfs{D}$, which controls the trade-off between risk neutrality and risk aversion. The utility function is risk-averse for very low resource and switches to a risk-neutral attitude after a threshold. All illustrated cases have $\alpha = 0.04$.}
 	\label{fig:oneswitch}
\end{figure}

Figure~\ref{fig:oneswitch} shows examples of the one-switch utility function with varying values for the parameter $\mathscrsfs{D}$ and a constant $\alpha$ value of 0.04. Since we are dealing with solving HTN planning problems in cost-based domains, the domain resource will decrease after each operator's execution. In this case, the attitude of an HTN planning agent can change from being risk-neutral to being risk-averse after a certain threshold.

\subsection{Plan's Expected Utility}

The ultimate objective when solving risk-aware HTN planning problems is to find optimal plans that follow a specific risk attitude. Plan optimality here entails plans with the highest expected utilities. The definition of a plan's expected utility dictates the approach that can be taken to solve risk-aware HTN planning problems. Therefore, we provide several definitions for the plan's expected utility by exploiting existing knowledge.

\begin{figure}[!t]
	\centering
	\includegraphics[width=0.75\columnwidth]{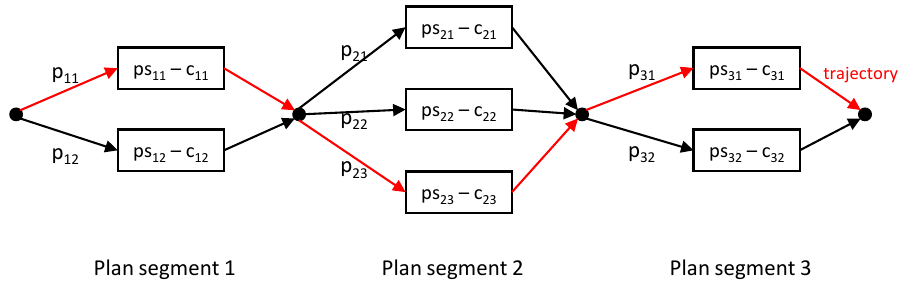}
	\caption{A plan consisting of three segments. Each segment $i$ contains the probability distribution over possible states and the corresponding action cost, i.e., resource consumption in a particular state. $p_{ij}$ denotes the probability of having state number $ps_{ij}$ in plan segment $i$ with the corresponding resource consumption of $c_{ij}$. Adapted from~\cite{koenig1996modeling}.}
	\label{fig:plan-seg}
\end{figure}

One way to define the expected utility of plans is by combining the expected utilities of smaller parts of the planning problem. Consider that planning problems can be divided into subproblems, each of which is solved separately, and the results are combined into plans.
A plan is then composed of multiple segments such that each segment has an action, or course of actions in the general case, with possible states (as a consequence of action's possible effects) and their corresponding costs~\cite{koenig1996modeling}. For example, Figure~\ref{fig:plan-seg} shows a plan with three segments. Since the actions have a probability distribution over possible action effects with their corresponding costs, each segment also contains a probability of distribution over possible states and corresponding action costs. We highlight one possible plan's \textit{trajectory} in Figure~\ref{fig:plan-seg}, where a plan's trajectory is the plan with one possible cost and one possible effect among all possible costs and effects of each of its operators. These considerations lay the basis for formulating the equation for computing the expected utility of such plans. Without loss of generality, we will consider each segment to consist of one operator. The same equations and discussion are generalisable to segments consisting of multiple operators. Moreover, the same equations are customisable for operators that have single effects but variable costs.

\begin{equation} \label{equ:3}
   EU(\pi) = \sum_{i=1}^{2} \sum_{j=1}^{3} \sum_{k=1}^{2} [p_{1i} p_{2j} p_{3k}  u(c_{1i} + c_{2j}+ c_{3k})]
\end{equation}

\noindent where $c_{1i}$, $c_{2j}$, and $c_{3k}$ are the costs of actions of the first, second, and third segments of the plan, respectively, with their corresponding probabilities $p_{1i}$, $p_{2j}$, and $p_{3k}$.

In the general case and for risk-aware HTN planning problems, we define the expected utility of a plan $\pi = \tuple{o_1, o_2, \ldots, o_s}$ as follows.

\begin{equation} \label{equ:planExpectedUtility-generalCase}
 EU(\pi) = \sum_{i=1}^{x_i} \sum_{j=1}^{x_j} \ldots \sum_{m=1}^{x_m} [p_{1i} p_{2j} \ldots p_{sm}  u(c_{1i} + c_{2j} + \ldots + c_{sm})]
\end{equation}
where:

\begin{itemize}
\item $x_k$ is the number of possible effects/costs for the $k$-th plan segment; 
\item $c_{sy}$ is the $y$-th possible cost of the $s$-th plan segment;
\item $p_{sy}$ is the $y$-th possible probability of the $s$-th plan segment;
\item $u$ is the utility of all plan's trajectories with respect to their costs. 
\end{itemize}

We can also maximise the expected utility of operators within each segment provided that the utility functions support segmentation, such as the family of concave and convex exponential utility functions. The plan's expected utility would then be a multiplication of each segment's result. For example, for exponential utility functions expressing risk-seeking and risk-averse attitudes, the calculation of the expected utility can be segmented as follows.

\begin{equation} \label{equ:planExpectedUtility-segmentation}
    \begin{split}
 EU(\pi) & = \sum_{i=1}^{x_i} \sum_{j=1}^{x_j} \ldots \sum_{m=1}^{x_m} [p_{1i} p_{2j} \ldots p_{sm} \dfrac{ae^{\alpha (c_{1i}(o_1) + c_{2j}(o_2) + \ldots + c_{sm}(o_s))}}{\alpha}]\\
 & = \dfrac{\alpha^{s-1}}{a^{s-1}} EU(o_1) \times EU(o_2) \times \ldots EU(o_s)
\end{split}
\end{equation}
where $EU(o_1)$,  $EU(o_2)$, $\ldots$, $EU(o_s)$ are the expected utilities of the plan's operators. The expected utility of each cost variable operator $o$, where $c(o) = \tuple{(p_{1}(o),c_{1}(o)), (p_{2}(o),c_{2}(o)), \cdots,
(p_{n}(o), c_{n}(o))}$, is defined as follows.

\begin{equation} \label{equ:action-eu}
  EU(o) =  \sum_{i=1}^{n} U_c(c_i(o)) \cdot p_i(o)
\end{equation} 

Note that in Equation~\ref{equ:planExpectedUtility-segmentation}, since the expected utilities of operators are multiplied with one another and since the utility function can take both positive and negative values, i.e., the expected utility can also be positive or negative, the multiplication with $\frac{1}{a^{s-1}}$ is necessary to preserve the sign of the expected utilities, regardless of the plan's length.

\begin{theorem}
\label{theo:planExpectedUtility-segmentation}
Equation~\ref{equ:planExpectedUtility-segmentation} holds true, and plan segmentation is possible when using exponential utility functions (Equation~\ref{equ:exponentialUtilityFunction}).
\end{theorem}

The proof of Theorem~\ref{theo:planExpectedUtility-segmentation} is presented in~\ref{appendix:proof-eu-segmentation}.

For the risk-neutral attitude, Equation~\ref{equ:planExpectedUtility-segmentation} fails to hold. This arises due to the inherent inability of the linear utility function to support segmentation. It is worth mentioning that Koening et al. assert a linear utility function that allows segmentation~\cite{koenig1996modeling}. However, that assertion is not accompanied by proof, and we could not prove it either. Nevertheless, in accordance with utility theory, the risk-neutral attitude evaluates its choices based on the average value of each choice. Consequently, enumerating all trajectories of a plan can be equivalent to summing the average cost of the plan's operators. In other words, it is reasonable to postulate that the risk-neutral attitude evaluates the plan's expected utility by summing the expected utility of its constituent operators as follows.

\begin{equation}
´\label{equ:linear-utility-eu}
\begin{split}
EU(\pi) & = EU(o_1) + EU(o_2) + \ldots + EU(o_s)
  \end{split}
\end{equation}

Another way to define the expected utility of plans draws inspiration from~\cite{magnenat2012integration}, where probabilities of success and utilities of actions play a leading role. The assumption here is that each action has a probability of success and utility. Then, the expected utility of a plan is expressed as a multiplication of two products. The first one is the success probability of all successful actions, such that the success probability of each action is either independent or depends on some of the successful actions that were previously executed in the plan. The second product, on the other hand, is the product of the utilities of all successfully executed actions.

This discussion brings us to the definition of the expected utility of a plan $\pi=\tuple{o_1, o_2, \ldots, o_k}$ for a risk-aware HTN planning problem.

\begin{equation} \label{equ:5}
  EU(\pi) =  \prod_{i=1}^{k} p(r_i| r_{1} ^{i-1} =1, \pi) \prod_{i=1}^{k} u(o_i; r_i =1)
\end{equation}
\noindent where $p(r_i| r_{1} ^{i-1} =1, \pi)$ is the probability of a successful execution of operator $o_i$ ($r_i =1$) in the plan $\pi$ (this probability depends on some of the previously executed operators $r_{1} ^{i-1} =1$), and $u(a_i; r_i =1)$ denotes the utility of an operator $o_i$ if it executes successfully $r_i =1$. The utility of a failure is 0.

Equation~\ref{equ:5} restricts the effects of operators to binary values, i.e., failure or success. Thus, in order to use this equation to compute the expected utility of plans for risk-aware HTN planning problems, we are pulled to make restrictive assumptions of the possible operator effects and utility computation. In particular, operators can have either successful or unsuccessful effects. Moreover, since the utility of unsuccessful operators is 0, operator utilities are only computed for the successful outcomes (effects), where each operator is assigned a single utility to model its usefulness. Having a single utility assigned to each operator and binary effects of operators, where unsuccessful operators have zero utility, implicitly restricts the computation to operators with a single cost. Our framework is more general and allows operators to have multiple possible costs and effects.

The last possibility defines the expected utility of a plan trajectory. This can be done by using operators with a probability distribution over effects and utilities of each state resulting from an operator's execution, i.e., possible effects, inspired by~\cite{meneguzzi2018goco}.

\begin{equation} \label{equ:6}
    EU(\pi_k) = \sum_{i=1}^{n} p(o_i) \times u(c_{ij} (o_i))
\end{equation}
where $\pi_k$ is the $k$-th trajectory of a plan and $c_{ij}(o_i)$ is the cost of one of the possible outcomes of the operator $o_i$ in the trajectory.
\section{Solving Risk-Aware Plan-Based HTN Planning Problems}
\label{sec:solving-plan-based}
 
A general solution to the Risk-aware HTN planning problem is provided by parametrizing it and introducing six Risk-Aware parameters (RA-parameters). Defining these parameters helps scoping, specialising, and categorising risk-aware HTN planning problems and hence, the approaches that can solve them. The RA-parameters $i, j, k, l, m,$ and $n$ are defined as follows.

\begin{itemize}
    \item $i$ is the HTN planning type, which can be state-based ($S$) or plan-based ($P$), i.e., $i \in \{S,P\}$;
    \item $j$ is the number of action's effects, where $j \in \{1,n\}$;
    \item $k$ is the risk attitude type, which can be static (0) or dynamic (1), i.e., $k \in \{0,1\}$;
    \item $l$ is the utility function type, which can be linear (0), exponential (1), or other (2), where the value of 2 contains all types of utility functions except for the linear and exponential utility functions family, i.e., $l \in \{0,1,2\}$; 
    \item $m$ is the utility function nature, which allows (1) or doesn't allow (0) segmentation, $m \in \{0,1\}$; and 
    \item $n$ is the plan's expected utility definition, which can be according to Equation~\ref{equ:planExpectedUtility-segmentation} (0), Equation~\ref{equ:linear-utility-eu} (1), Equation~\ref{equ:5} (2), or Equation~\ref{equ:6} (3), i.e., $n \in \{0,1,2,3\}$.
\end{itemize}
We use the notation $Pijklmn$ to define risk-aware HTN planning problems as special cases of the general framework of risk-aware HTN planning problems using the RA-parameters. For example, a $PP10110$ refers to plan-based risk-aware HTN planning problems ($i = P $), where actions have one effect ($j = 1$) and variable cost, and are solved by agents with a static risk attitude ($k = 0$) using an exponential utility function ($l = 1$) that allows segmentation ($m = 0$), with the expected utility defined according to Equation~\ref{equ:planExpectedUtility-segmentation} ($n = 0$).

In total, there are at least 64 combinations of these parameters that define the possible special cases of risk-aware HTN planning problems. The 64 combinations are computed considering only the utility function type $l \in \{0,1\}$ since there can be an infinite number of utility functions when $l = 2$, which makes the number of possible combinations infinite. Moreover, since we consider only the linear and exponential utility functions in the computation of the least number of combinations, the parameter $m = 0$ since both families of utility functions allow segmentation. 
 
Using the RA-parameters, we can express a full spectrum of risk-aware planning problems in terms of their generality and complexity, where the most general and complex planning problems are $PPn1\ast \ast \ast$ and $PSn1\ast \ast \ast$, where operators have a probability distribution of multiple possible effects and their corresponding costs. Fixing $j$ to 1 and $k$ to 0 means that no further specialization is possible regarding the number of action's effects and the dynamics of the agent's risk attitude. 

In this work, we start discussing possible solutions to these cases. Particularly, we deal with  $PP10011$ and $PP10110$ risk-aware planning problems. The main reason is that the computation of the expected utility can be divided into segments. We refer to these models for plan-based HTN planning as \textit{risk-aware plan-based HTN planning}. Furthermore, we discuss the $PS10011$ and $PS10110$ cases, and we refer to cases that use state-based HTN planning model as \textit{risk-aware state-based HTN planning}. We also provide a discussion of the $PP1\ast20\ast$ and $PS1\ast20\ast$ cases, which include an infinite number of possible risk-aware HTN planning problems since there can be an infinite number of utility functions that do not allow segmentation. In these cases, we discuss, for the risk-seeking and risk-averse attitudes, the usage of utility functions that do not allow segmentation, i.e., utility functions other than the family of utility functions from Equation~\ref{equ:exponentialUtilityFunction}. Then, finding the plan with the highest expected utility, as defined in Equation~\ref{equ:planExpectedUtility-generalCase}, is more complex as it can require the enumeration of all plan trajectories in the worst case.

Searching for a plan with maximum expected utility implies searching for a specific, desired outcome, not any outcome. In AI planning, heuristics are a popular mechanism for guiding the search towards a desired outcome, meaning HTN planning augmented with heuristic search is one relevant mechanism for constructing approaches for solving risk-aware HTN planning problems. In this context, the question that arises is whether we can exploit existing heuristics from classical HTN planning, such as the admissible heuristic defined for a hybrid framework that extends the standard HTN planning model with partial order causal link (POCL)
planning~\cite{bercher2017admissible}. As this heuristic allows for finding optimal plans, it could provide the foundation for defining a heuristic for finding plans with maximum expected utility. Inspired by this cost-aware heuristic based on a ground TDG, we define our risk-aware heuristic over a ground CV-TDG, i.e., GCV-TDG, where the graph nodes represent instantiations of the parametrised tasks and methods according to the problem instance (see Definition~\ref{def:GCVTDG}). In the GCV-TDG, method nodes are nodes that are connected by multiple edges to the tasks in their task networks.\footnote{In~\cite{bercher2017admissible}, method nodes represent partial plans, i.e., the whole task network resulting from decomposing a specific task using the method.} The basic premise is that primitive tasks are assigned a probability distribution of negative costs, and the heuristic will estimate the expected utilities of methods and tasks by preprocessing the GCV-TDG in a bottom-up manner. The GCV-TDG preprocessing is followed by a plan generation, during which the expected utility estimations are used to compute the plan with the highest expected utility. In order to compute such a plan, the risk-aware heuristic must be admissible.

As typical in heuristic-based planning, the approach consists of two key phases: \textit{preprocessing}, where we compute the risk-aware heuristic, and \textit{plan generation}, where we compute the plan with the highest expected utility. In the plan generation phase, we adopt the A* algorithm due to its completeness and optimality. In the preprocessing phase, we compute the heuristics using the logarithm in order for the A* algorithm to work correctly. The intuition behind the heuristic computation using the logarithm becomes clearer when explaining the plan generation phase in Section~\ref{subsub:plan_based-stepTwo}.

\subsection{Phase One: Preprocessing}

Preprocessing is based on the concepts of \textit{decomposition path}, \textit{cycle}, and \textit{cycle membership}. The decomposition path defines a possible sequence of nodes in the GCV-TDG from a root node, i.e., a task in the initial task network, to a leaf node, i.e., a primitive task. Since the GCV-TDG can be a cyclic graph, we define the concepts of cycle and cycle membership, which we use in our approach to handle cycles in the preprocessing phase.

\begin{definition}[Decomposition path]
\label{def:decomposition_path-preprocessing}
A {decomposition path} $\pi_d \in \Pi$ is a sequence of tasks $\langle v_{\overline{t_1}}, v_{\overline{t_2}}, \ldots, v_{\overline{t_n}} \rangle$, such that
\begin{itemize}
    \item $\Pi$ is the set of all decomposition paths,
    \item $\forall v_{\overline{t}_i}, v_{\overline{t}_{i+1}} \in \pi_d, \exists v_{\overline{m}} \in V_{\overline{M}}$, such that
    \begin{itemize}
        \item $i \in \{1,2, \ldots, n-1 \}$,
        \item $v_{\overline{t_i}} \in V_{\overline{T_C}}$ and $ct(v_{\overline{m}})=v_{\overline{t}_i}$, and
        \item $\overline{t}_{i+1} \in tn(v_{\overline{m}})$.
    \end{itemize}
\end{itemize}
\end{definition}

\begin{definition}[Cycle]
\label{def:cycle-preprocessing}
A cycle $c$ in GCV-TDG is a decomposition path $\pi_d = \langle v_{\overline{t}_1}, v_{\overline{t}_2}, \ldots, v_{\overline{t}_n} \rangle$ such that $v_{\overline{t}_1}$ = $v_{\overline{t}_n}$, i.e., $v_{\overline{t}_1}$ and $v_{\overline{t}_n}$ have the same task name and parameters. $c \in C$, where C is a finite set of cycles in the GCV-TDG.
\end{definition}

\begin{definition}[Cycle membership]
\label{def:cycle-membership}
A method $v_{\overline{m}}$ is said to be part of a cycle $c$ if and only if $\exists c = \langle \ldots, v_{\overline{t}_i}, v_{\overline{t}_{i+1}}, \ldots \rangle \in C$ such that $ct(\overline{m})$ = $v_{\overline{t}_i}$ and $v_{\overline{t}_{i+1}} \in tn(\overline{m})$.
\end{definition}

These definitions apply to the GCV-TDG, which is relevant for the preprocessing phase only as nodes represent individual tasks and not task networks (i.e., partial plans).

The preprocessing phase consists of three steps. The first step is concerned with computing the expected utilities of primitive task nodes, i.e., operators. The second step is concerned with initialising the expected utility of the method and compound task nodes. The third step is concerned with the computation of the expected utilities of the method and compound task nodes while handling cycles. To realise the first and second steps, we propose Algorithm~\ref{alg:computeUtilities-initialisation}, which takes as an input the ground cost-variable TDG (GCV-TDG) and the utility function that expresses the risk attitude. The algorithm starts by computing the expected utilities of primitive tasks (lines 1-5). The expected utility of a primitive task equals the expected utility of the corresponding operator, which is the weighted average of the utilities of each of the possible costs of the operator. The weights represent the probability distribution of these costs. Thus, $\forall v_{\overline{tp}} \in V_{\overline{TP}}$, such that $v_{\overline{tp}} = \overline{o} = \tuple{\overline{pt}(o),\overline{pre}(o),\mathit{\overline{eff}(o)}, c(o)}$ and $c(o) = \tuple{(p_{1}(o),c_{1}(o)), (p_{2}(o),c_{2}(o)), \cdots, \\
(p_{n}(o), c_{n}(o))}$, we define the expected utility of $\overline{o}$ according to Equation~\ref{equ:action-eu}.

The utility of each outcome, $U_c(c_i(o))$, is computed using the utility function $U$, which can be one of the functions defined in Equation~\ref{equ:exponentialUtilityFunction} and allows choosing a risk attitude and determining its degree/intensity. For the risk-neutral attitude, the expected utility of primitive tasks is computed according to Equation~\ref{equ:action-eu} (line 3). For the risk-seeking and risk-averse attitudes, we compute the expected utility of primitive tasks by using the logarithm of the product of the expected utility with the $\alpha$ value, which is the value that determines the degree/intensity of the risk attitude in the exponential utility function. To preserve the sign of the expected utility, we multiply the product with $a$, which is the value that determines the sign of the exponential utility function and the expected utility as a consequence. As mentioned before, we defer the reasoning for taking the logarithm of the expected utility and the multiplication with $\alpha$ and $a$ until we get to phase two, when this becomes apparent. The corresponding equation is defined below and positioned in line 5 in Algorithm~\ref{alg:computeUtilities-initialisation}.

\begin{equation} \label{equ:action-eu-log}
  EU(\overline{o}) =  a \cdot \log_{10}(\alpha \cdot \sum_{i=1}^{n} U_c(c_i(o)) \cdot p_i(o))
\end{equation} 

\begin{algorithm}[h] 
	\caption{Initialisation and computation of primitive tasks' expected utility}
	\label{alg:computeUtilities-initialisation}
        \hspace*{\algorithmicindent} \textbf{Input:} \textit{GCV-TDG} $= \tuple{V_{\overline{TC}},V_{\overline{TP}},V_{\overline{M}},E_{\overline{TC} \to \overline{M}},E_{\overline{M} \to \overline{TC}},E_{\overline{M} \to \overline{TP}}}$ \\ 
        \hspace*{\algorithmicindent} \textbf{Input:} $U$ (utility function corresponding to the chosen risk attitude)
        \begin{algorithmic}[1]
            \ForAll{$v_{\overline{t_p}} \in V_{\overline{TP}} $}
                \If{$U$ is linear}
                    \State $EU(v_{\overline{t_p}})$ $\leftarrow$ $EU(v_{\overline{o}})_{\overline{pt}(o) = \overline{t_p}(\tau_p)} = \sum_{i=1}^{n} U_c(c_i(\overline{o})) * p_i(\overline{o})$
                \Else
                    \State $EU(v_{\overline{t_p}})$ $\leftarrow$ $EU(v_{\overline{o}})_{\overline{pt}(o) = \overline{t_p}(\tau_p)} = a \cdot \log_{10}(\alpha \cdot \sum_{i=1}^{n} U_c(c_i(\overline{o})) * p_i(\overline{o}))$
                \EndIf
            \EndFor
            \ForAll{$v_{\overline{m}} \in V_{\overline{M}} $}
                \State $EU(v_{\overline{m}}).initialise(+ \infty)$
                \State $EU(v_{\overline{tc}}).initialise(+ \infty): \overline{ct}(m) = v_{\overline{tc}}$
            \EndFor
            \State $gid=0$
            \ForAll{$v_{\overline{t_c}} \in V_{\overline{TC}} $}
                    \State $gid \leftarrow gid + 1$
                    \State path $\leftarrow$ null
                    \State $EU(v_{\overline{tc}}).add$ (\Call{ComputeTaskEU}{$v_{\overline{tc}}$, $path$, $gid$, $V_{\overline{M}}$, $U$}, $gid$)
            \EndFor	
	\end{algorithmic}
\end{algorithm}

The algorithm initialises an \emph{expected utility vector} of all methods and compound tasks by assigning the biggest representable number, denoted by the infinity symbol,
to indicate that these values are not yet computed (lines 6-8). Having an expected utility vector for each method and each compound task instead of a single-value expected utility allows us to keep the expected utilities of compound tasks and methods as being part of all possible decomposition paths, including those that form a cycle. This helps us later in the process of handling cycles. The reasoning behind the expected utility vector will become apparent when we explain the process of handling cycles.
Lastly, the algorithm iterates over all compound tasks and invokes the \verb|ComputeTaskEU| function that returns the heuristic value that determines the tasks' expected utilities (lines 10-13). However, as we are iterating over all compound tasks in order to compute their expected utilities by using this function, it may happen that one of the tasks gets its expected utility computed while computing another task’s expected utility due to its existence in the same decomposition path. Even if this is the case, we still need to compute the expected utility of the task when it is part of other decomposition paths and store them in the expected utility vector of the task. In particular, it is important to consider the occurrences of the task in all decomposition paths since its expected utility might get another value when it is part of a cycle. We assign a unique identity to each part of the GCV-TDG whose root task is the current compound task for which we are computing the expected utility, denoted by $gid$. This concept allows us to track the occurrences of methods and compound tasks in all possible decomposition paths.

As heuristics that determine the expected utility of compound tasks and method nodes, we use Equations~\ref{equ:plan_based_heuristic-compound_task} and~\ref{equ:plan_based_heuristic-method}, respectively. For a compound task, the expected utility $EU_T(v)$ is the maximum of the expected utilities of all methods that can decompose it. The expected utility of a method is the product of the expected utilities of the tasks in its task network.

\begin{equation} \label{equ:plan_based_heuristic-compound_task}
     EU_T(v) =
\begin{cases}
EU(\overline{o}),   &  \text{if $v=\overline{o} $ and $v = v_{\overline{tp}}$ and $v_{\overline{tp}} \in V_{\overline{TP}}$}; \\ 
max_{(v,v_{\overline{m}})\in E_{\overline{TC} \to \overline{M}}} EU_{M} (v_{\overline{m}}),  &  \text{if $v = v_{\overline{tc}}$ and $v_{\overline{tc}} \in V_{\overline{TC}}$}. 
\end{cases}
 \end{equation} 
 
\begin{equation} \label{equ:plan_based_heuristic-method}
  EU_{M} (v_{\overline{m}}) = a  \prod_{(v_{\overline{m}},v_{\overline{t_i}})\in E_{\overline{M} \to \overline{T_n}} } |EU_T(v)|
\end{equation}

Since we are using the logarithmic value of the expected utility for primitive tasks, as in Equation~\ref{equ:action-eu-log}, we need to modify Equation~\ref{equ:plan_based_heuristic-method} such that the expected utility of the method is computed by summing up the expected utilities of the tasks in its task network, as shown in Equation~\ref{equ:plan_based_heuristic-method-log}.

\begin{equation} \label{equ:plan_based_heuristic-method-log}
  EU_{M} (v_{\overline{m}}) =  \sum_{(v_{\overline{m}},v_{\overline{t_i}})\in E_{\overline{M} \to \overline{T_n}} } EU_T(v)
\end{equation}

The recursive \verb|ComputeTaskEU| function in Algorithm~\ref{alg:computeUtilities} is used to compute these heuristics. The function takes as input: (i) the ground task node \emph{groundTsk} for which the heuristic will be computed using Equation~\ref{equ:plan_based_heuristic-compound_task}; (ii) a \emph{decompPath}, which is a decomposition path that includes all the previously visited task nodes until reaching the current task node; (iii) the \emph{pGCVTDGid}, which is the decomposition path identifier; (iv) the set of all ground method nodes $V_{\overline{M}}$ from the GCV-TDG; and (v) the utility function $UtilityF$ that expresses the chosen risk attitude. The algorithm starts from a compound task node and iterates over all the methods that can decompose it in order to compute their expected utilities (lines 6-16). We use the expected utility vector of the method to store the expected utilities of the method computed for all decomposition paths that the method is part of. For each decomposition path that the method is part of, we store the expected utility of the method as being part of this decomposition path at an index equal to the path identifier \emph{pGCVTDGid} in the expected utility vector. In order to avoid computing the expected utility of the method multiple times for the same decomposition path, we check if this expected utility is already computed at this index (line 7). If the expected utility of the method was not computed for the current decomposition path, the algorithm then iterates over the sub-task nodes in each method’s task network and invokes the function recursively to compute the expected utilities of these sub-task nodes (lines 8-10). The expected utility of each sub-task is stored in its expected utility vector at the index that is equal to the path identifier \emph{pGCVTDGid} (line 10). The expected utility of each method is calculated based on the task expected utilities of the method’s task network considering Equation~\ref{equ:plan_based_heuristic-method-log} (lines 12-15). The expected utility of the current task is computed, according to Equation~\ref{equ:plan_based_heuristic-compound_task}, after iterating over all the methods that can decompose it. Since our goal is to maximise the expected utility, the algorithm returns the maximum of the expected utilities of all methods that can decompose the task (lines 15-16).

\begin{algorithm}[h!t] 
	\caption{Compute expected utilities for all task and method nodes}
	\label{alg:computeUtilities}
        \hspace*{\algorithmicindent} \textbf{Input:} $groundTsk$ \\ 
        \hspace*{\algorithmicindent} \textbf{Input:} $decompPath$ \\ 
        \hspace*{\algorithmicindent} \textbf{Input:} $pGCVTDGid $\\
        \hspace*{\algorithmicindent} \textbf{Input:} $V_{\overline{M}}$ \\ 
        \hspace*{\algorithmicindent} \textbf{Input:} $UtilityF$ \\ 
        \hspace*{\algorithmicindent} \textbf{Output:} $taskEU$ \\ 
        \begin{algorithmic}[1]
        \Procedure{ComputeTaskEU}{$groundTsk, decompPath, pGCVTDGid, V_{\overline{M}}$, $UtilityF$}
        \If{task is primitive}
            \Return $EU(groundTsk)$
        \Else
            \If{$v_{\overline{t}_i} \in decompPath = \langle \ldots, v_{\overline{t}_i}, \ldots, v_{\overline{t}_n} \rangle : groundTsk = v_{\overline{t}_i}$}
                    \State $(EU({v_{\overline{m}'}})).add(-\infty, gid) :\exists v_{\overline{t}_{i-1}} \in decompPath \land v_{\overline{t}_{i-1}} = \overline{ct}(m')$
            \EndIf
            
            \ForAll{$v_{\overline{m}} \in V_{\overline{M}} : \overline{ct}(m)= groundTsk$}
                \If{$EU(v_{\overline{m}})$.at(gid) = $+ \infty$}
                    \ForAll{$SubT \in tn(\overline{m})$}
                        \State pathCopy $\leftarrow$ copy(decompPath).append(groundTsk)
                        \State \parbox[t]{.5\linewidth}{$EU(v_{\overline{t}}).add(\Call{ComputeTaskEU}{v_{\overline{t}}, pathCopy, gid, \\
                        V_{\overline{M}}, UtilityF}, gid) : {v_{\overline{t}} =SubT}$}
                    \EndFor
                    \If{$EU(v_{\overline{m}})$.at(gid) = $+ \infty$}
                        \ForAll{$SubT \in tn(\overline{m})$}
                            \State $EUTemp += EU(subT).$at(gid)
                        \EndFor
                        \State $EU(v_{\overline{m}}).add(EUTemp, gid)$
                
                    \EndIf
                \EndIf
        \EndFor
        \State maxEU $\leftarrow$ $max(EU(v_{\overline{m}}).at(gid)) : v_{\overline{m}} \in V_{\overline{m}} \land \overline{ct}(\overline{m}) = groundTsk$
        \Return maxEU
        \EndIf
        \EndProcedure
	\end{algorithmic}
\end{algorithm}

\subsubsection*{Handling Cycles}

The presence of cycles in the preprocessing phase requires additional treatment, which consists of two main steps. The first is about detecting cycles by tracking the reoccurrence of compound tasks in the same decomposition path. The second is about correctly assigning expected utilities for methods that are members of a cycle (Definition~\ref{def:cycle-membership}) and compound tasks that form the cycle. This step ensures that during the bottom-up preprocessing phase, the expected utilities of non-cycle decomposition paths can be computed while accounting for methods that are not members of a cycle as alternative options. These two steps are necessary to enable the preprocessing phase to terminate in the presence of cycles. This is especially important when having positive expected utility values, such that going infinitely through a cycle increases the expected utility infinitely. Detecting cycles is done by using the decomposition path (Definition~\ref{def:decomposition_path-preprocessing}). In particular, starting from a compound task node $v_{tc}$, a decomposition path is formed by adding this task to the path, and at each decomposition level, one of its sub-tasks is also appended to this path, i.e., it includes all the previously visited task nodes until reaching the current task node. The same process is repeated until reaching either a primitive task or a task that is already in the path, i.e., forming a cycle. A copy of this sequence is passed from a task to the tasks in its method task networks (lines 9-10), such that a cycle can be found if the same task existed formerly in the path. When the algorithm detects a recursive task, i.e., a task that is part of a cycle, it assigns the value of $-\infty$ to the expected utility at \emph{pGCVTDGid} index of the last method, i.e., the method that has the recursive task in its task network, in this decomposition path (lines 4-5). Since we are maximising, assigning the $-\infty$ value to the last method in the cycle allows us to consider the expected utility of the other methods that might not lead to a cycle. In particular, this allows the algorithm to terminate while also computing the correct maximum heuristic.

\begin{figure}
     \centering
     \begin{subfigure}[b]{\textwidth}
         \centering
         \includegraphics[width=0.611\textwidth]{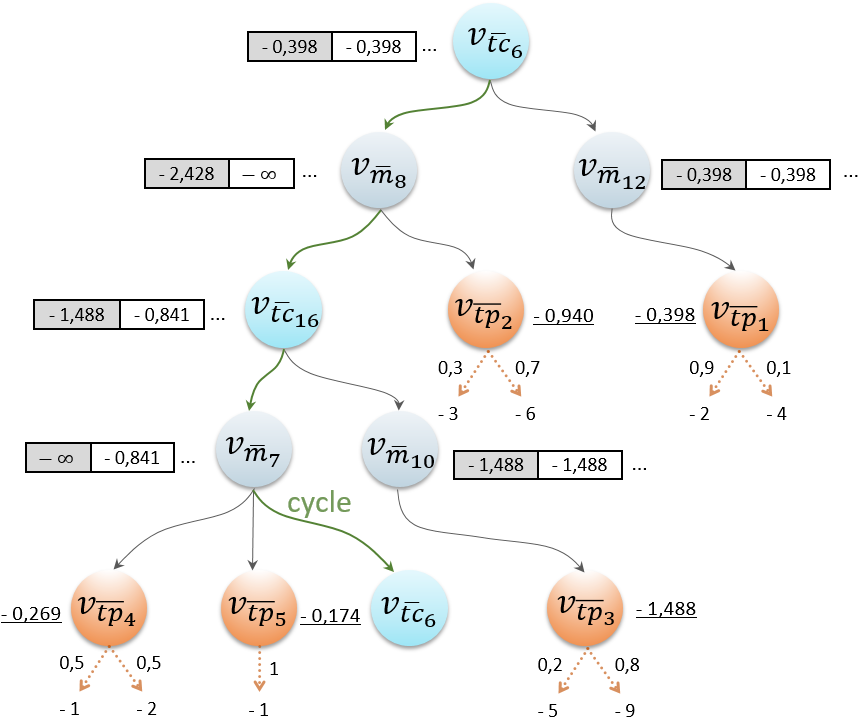}
         \caption{Partial graph of GCV-TDG with $v_{\overline{t}_6}$ as a root node.}
         \label{subfig:plan-based-ex1}
     \end{subfigure}
     
     \begin{subfigure}[b]{\textwidth}
         \centering
         \includegraphics[width=0.611\textwidth]{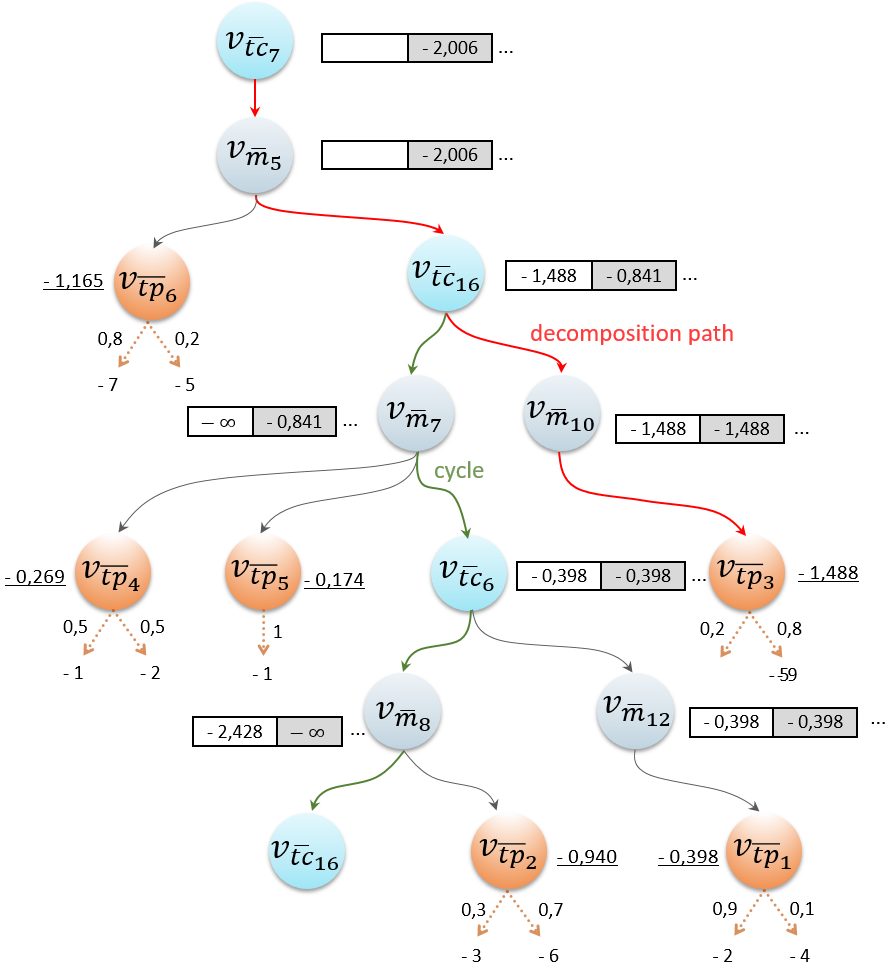}
         \caption{Partial graph of GCV-TDG with $v_{\overline{t}_7}$ as a root node.}
         \label{subfig:plan-based-ex2}
     \end{subfigure}
        \caption{Abstract example of two partial graphs of GCV-TDG in the preprocessing phase of risk-aware plan-based HTN planning. Blue, orange, and grey nodes represent abstract tasks, cost variable operators, and methods, respectively. Orange arrows represent the probability distribution of the costs of each operator. Underlined values besides the operators are the expected utilities computed using Equation~\ref{equ:action-eu-log}. Rectangles represent the expected utility vector, where the value corresponding to the current subgraph is highlighted in grey.}
        \label{fig:abstract_example_plan-based}
\end{figure}

To illustrate the idea of heuristic calculation in the presence of cycles and when the same tasks and methods are present in multiple decomposition paths, consider two abstract graphs that are part of a GCV-TDG, illustrated in Figure~\ref{fig:abstract_example_plan-based}. The first partial graph (Figure~\ref{subfig:plan-based-ex1}) has the task $v_{\overline{tc}_6}$ as its root and the second graph (Figure~\ref{subfig:plan-based-ex2}) has the task $v_{\overline{tc}_7}$ as its root node. Figure~\ref{fig:abstract_example_plan-based} shows the expected utility function of all primitive tasks as computed by Algorithm~\ref{alg:computeUtilities-initialisation} using the logarithmic value (Equation~\ref{equ:action-eu-log}) and the exponential utility function for risk-averse attitude with $\alpha = 0.4$ (Equation~\ref{equ:exponentialUtilityFunction}). Algorithm~\ref{alg:computeUtilities} computes the expected utility of task $v_{\overline{tc}_6}$ recursively in a bottom-up manner. When decomposing $v_{\overline{tc}_6}$ using $v_{\overline{m}_8}$ and decomposing $v_{\overline{tc}_{16}}$ using $v_{\overline{m}_7}$, the algorithm detects the cycle $c_1 = \langle v_{\overline{tc}_6}, v_{\overline{tc}_{16}}, v_{\overline{tc}_6}\rangle$ and assigns the value of $- \infty$ to the last method in the cycle, i.e., $v_{\overline{m}_7}$. Assigning this value allows maximising the expected utility in the presence of cycle since the expected utility of $v_{\overline{tc}_{16}}$ takes the maximum of the expected utilities of $v_{\overline{m}_7}$ and $v_{\overline{m}_{10}}$. Since $v_{\overline{m}_7}$ leads to a cycle, assigning the expected utility of $v_{\overline{m}_{10}}$ allows accounting for better options than a cycle when such options exist, as shown in Figure~\ref{subfig:plan-based-ex1}. In the second partial graph (Figure~\ref{subfig:plan-based-ex2}), we see that $v_{\overline{t}_{16}}$ is part of decomposition paths different from the ones in Figure~\ref{subfig:plan-based-ex1}. Thus, it might have a different expected utility, which is the case in this example. In this partial graph, there is also a cycle $c_2 = \langle v_{\overline{tc}_{16}}, v_{\overline{tc}_6}, v_{\overline{tc}_{16}}\rangle$, which makes the expected utility of $v_{\overline{m}_8}$ equals $-\infty$. However, unlike the first partial graph, the expected utility of $v_{\overline{m}_7}$ in this partial graph does not equal $-\infty$. The different expected utilities of all methods and compound tasks are stored in their expected utility vectors. During the search, the maximum expected utility in each vector is considered. For example, the expected utility vector of $v_{\overline{tc}_{16}}$ includes the values of $-1,487$ and $-0,841$. During plan generation, the algorithm considers the value of $-0,841$ as a heuristic for $v_{\overline{tc}_{16}}$. This means that the heuristic value is equal to or bigger than the actual expected utility of the task, i.e., the heuristic overestimates the actual expected utility. As we will see, this makes the heuristic admissible, which is important for the optimality of the algorithm.

Before assigning the value of the method’s expected utility, the algorithm examines again whether this value is already computed (line 12). The reason behind this repeated examination is that the current task might be part of a direct cycle, and the algorithm may realise that during the recursive calls. Thus, one of its methods, which might be the current one, could have been assigned an expected utility value that equals $-\infty$. Thus, we do the second examination to avoid erasing the already computed expected utility of the method.

For each method or compound task, we have an expected utility vector instead of a single expected utility value because it might happen that the task appears in several decomposition paths, where it can be part of a cycle in one decomposition path but not part of a cycle in another. Thus, as for the preprocessing phase, in the presence of a cycle, the expected utilities are computed differently.  For a compound task within a cycle, this approach enables the computation of the task’s expected utility while considering alternative decomposition paths where the task does not participate in a cycle. We store these multiple values of the task's expected utility in the task's expected utility vector. During the second phase, the maximum of these values is considered. If a compound task that is part of a cycle also occurs in at least one decomposition path that does not form a cycle, the maximum value in the task's expected utility vector will correspond to the alternative decomposition paths. The task's expected utility is stored in the expected utility vector at the index corresponding to the $pGCVTDGid$, which is an identifier we use to enumerate all possible decomposition paths. The reason behind storing it at the $pGCVTDGid$ index is to avoid computing the task's expected utility multiple times for the same decomposition path.

\subsection{Phase Two: Plan Generation}
\label{subsub:plan_based-stepTwo}

For the plan generation, we propose an algorithm, inspired by~\cite{alford2012htn}, that is based on A* search and employs
 an A* adaptation of the heuristic given by Equation~\ref{equ:utility_partial-plan} to compute the plan with the highest expected utility. Equation~\ref{equ:planExpectedUtility-segmentation} is used to model risk-seeking and risk-averse attitudes and Equation~\ref{equ:linear-utility-eu} for the risk-neutral attitude. We explain this in detail next.

Given that we deal with partial plans at each step of the computation, and since partial plans are also plans, their expected utilities are computed in the same way as solution plans. However, as we do not know the exact expected utility of compound tasks during planning, we use the heuristic computed in the preprocessing phase. In particular, for each partial plan, we retrieve the maximum expected utility of a compatible grounding of each task $comp (t(\bar{\tau}))$. Then, the heuristic that computes the expected utility of this partial plan is the product of the expected utilities of all its compound tasks as shown in Equation~\ref{equ:utility_partial-plan}. Partial plans are sorted by A* based on the product of the expected utilities of their compound tasks multiplied by the product of the expected utilities of their primitive tasks. The first product represents an overestimation of the expected utility gained from decomposing the set of compound tasks into primitive tasks, whereas the second product represents the expected utilities of the already refined tasks, that is, the expected utility of the primitive tasks resulting from all previous refinements. Sorting partial plans in A* according to their expected utility provides the approach with the ability to implicitly make informed choices of methods when decomposing a compound task. These informed choices are made based on the expected utility of the partial plan resulting from this decomposition.

\begin{equation} \label{equ:utility_partial-plan}
  EU_{TDG} (P) = a  \prod_{l:t(\bar{\tau})\in P \land t(\bar{\tau}) compound} |(max_{v_{\overline{tc}} \in comp (t(\bar{\tau}))} EU_T (v_{\overline{tc}}))|
\end{equation}

However, we still need to adapt the computation of partial plans' expected utility to work with A*. In particular, the A* algorithm sorts the nodes in the fringe, i.e., the open list, based on the function $f(n)=g(n)+h(n)$, where $g(n)$ is the cost from the start node to the current node, and $h(n)$ is the heuristic that estimates the cost needed to reach the goal. In our case, the $f(n)$ function, i.e., the expected utility of partial plans, is computed as a product between $g(n)$, which represents the expected utility of primitive tasks (already refined tasks that are in the partial plan), and $h(n)$, which represent the heuristic that estimates the expected utility of the compound tasks that are still to be refined. To adapt the $f(n)$ function to work with A*, we have to adapt the plan's expected utility by taking the logarithm of the expected utility for the risk-averse and risk-seeking attitudes, as shown in Equation~\ref{equ:plan_expected_utility-log}. This is done by taking the logarithm of the plan's expected utility since, if $y = x \cdot z$, $log_{10}(y) = log_{10}(x) + log_{10}(z)$ and if $y < a \rightarrow log_{10}(y) < log_{10}(a)$. Thus, for two plans $\pi_1$ and $\pi_2$ if $EU(\pi_1) < EU(\pi_2) \rightarrow log_{10}(EU(\pi_1))  < log_{10}(EU(\pi_2)) $. This means that taking the logarithm of the plan's expected utility, as in Equation~\ref{equ:plan_expected_utility-log}, preserves the ordering among plans based on their expected utility. 

\begin{equation} \label{equ:plan_expected_utility-log}
 \begin{split}
 log_{10}(EU(\pi)) & = log_{10}(\dfrac{\alpha^{s-1}}{a^{s-1}} EU(o_1) \times EU(o_2) \times \ldots EU(o_s))\\
 & = log_{10}(\dfrac{a}{\alpha} \cdot \dfrac{\alpha^s}{a^s} EU(o_1) \times EU(o_2) \times \ldots EU(o_s))\\
 & = log_{10}(\dfrac{a}{\alpha} \cdot \dfrac{\alpha}{a} \cdot EU(o_1) \times \dfrac{\alpha}{a} \cdot EU(o_2) \times \ldots \dfrac{\alpha}{a} \cdot EU(o_s))\\
 & = a \cdot log_{10}(\dfrac{1}{\alpha}) + a \cdot log_{10}(\alpha \cdot EU(o_1)) +  a \cdot log_{10}(\alpha \cdot EU(o_2)) + \\
 & \ \ \ \ldots + a \cdot log_{10}(\alpha \cdot EU(o_s))
\end{split}
\end{equation}

Note that in Equation~\ref{equ:plan_expected_utility-log}, we distribute the term $\dfrac{\alpha^{s-1}}{a^{s-1}}$ to the expected utility of the individual operators. The reason for this is to keep the computation of the plan's expected utility modularisable, i.e., able to segment it, since $s$ is the number of the plan's operators, which is not known until the full plan is computed. In other terms, it cannot be computed for partial plans since the number of operators of the partial plan is not known at plan generation time. The heuristic computed in the preprocessing phase complies with Equation~\ref{equ:plan_expected_utility-log}. That is, for each operator $o_k$, we compute the heuristic as the $a \cdot log_{10}(\alpha \cdot EU(o_k))$, and the heuristic for methods is the sum of the heuristic of all tasks in its task network.\footnote{When comparing the logarithm of the expected utility of different plans, calculated using Equation~\ref{equ:plan_expected_utility-log}, the term $a \cdot log_{10}(\dfrac{1}{\alpha})$ can be disregarded, as it remains constant across all plans.}

Based on the adaption of the plan's expected utility, the computation of the $f(n)$ value for each partial plan becomes as follows. For each partial plan, we retrieve the maximum expected utility of a compatible grounding of each task $comp (t(\bar{\tau}))$. Then, the heuristic that computes the expected utility of this partial plan is the sum of the expected utilities of all its compound tasks as shown in Equation~\ref{equ:utility_partial-plan-sum}. Partial plans are sorted by A* based on the sum of the expected utilities of their compound tasks added to the sum of the expected utilities of their primitive tasks computed in the preprocessing phase.

\begin{equation} \label{equ:utility_partial-plan-sum}
  EU_{TDG} (P) =   \sum_{l:t(\bar{\tau})\in P \land t(\bar{\tau}) compound} |(max_{v_{\overline{tc}} \in comp (t(\bar{\tau}))} EU_T (v_{\overline{tc}}))|
\end{equation}

To realise the second phase, i.e., the planning phase, we propose Algorithm~\ref{alg:computeHighestExpectedUtilityPlan}. The three concepts of decomposition, executable task networks, and solution introduced in Definitions~\ref{def:decomposition-plan_based},~\ref{def:exe-taskNetwork}, and \ref{def:solution-plan_based} are relevant to plan-based planning and, thus, are important to understand how Algorithm~\ref{alg:computeHighestExpectedUtilityPlan} works. The definition of executable task networks is important to define what constitutes a solution for plan-based planning.

Definition~\ref{def:solution-plan_based} is a general one for HTN plan-based planning. The definition of the risk-aware plan-based planning solution follows the same definition in addition to having the highest expected utility, i.e., reflecting the agent's risk attitude (see Definition~\ref{def:risk-aware-htn-planning}).

Going back to Algorithm~\ref{alg:computeHighestExpectedUtilityPlan}, the algorithm takes as an input the risk-aware HTN planning problem to be solved and the fringe, which represents the open list in A*. The output of the algorithm is the plan with the highest expected utility. The algorithm starts with the initial task network and computes the $f(tn_0)$ value for it (lines 1-6)  by summing up the expected utility of primitive tasks to compute the value of $g(tn_0)$ (lines 2-3) and summing up the compatible expected utilities of the compound tasks to compute the value of $h(tn_0)$ (lines 4-5). Then, the algorithm starts by initialising the \emph{fringe} and the $V$ lists with the initial task network (line 7). These lists represent the open and closed lists of the A* algorithm, where the fringe contains unexpanded task networks while the closed list contains the already decomposed task networks. The algorithm then loops until the \emph{fringe} is empty. When the fringe is empty, but no solution is returned (line 11), it means that there is no solution to the planning problem. At each iteration, the first node, i.e., the first partial plan, is popped from the fringe. This node has the highest plan's expected utility since nodes in the fringe are ordered based on their expected utilities. If this partial plan, i.e., task network $tn_{current}$, is \emph{primitive} and \emph{executable} (see Definition~\ref{def:exe-taskNetwork}), it constitutes a solution to the planning problem (see Definition~\ref{def:solution-plan_based}). Thus, it is returned, and the algorithm terminates (line 11). The solution task network is the result of an arbitrary number of decompositions to the initial task network $tn_0$. This is denoted as $tn_0 \rightarrow^{*}_{D} tn_{current}$.

If the popped task network $tn_{current}$ is not a solution yet, we add all its immediate decompositions (see Definition~\ref{def:decomposition-plan_based}) to the \emph{decompositions} set (line 12). For each possible decomposition, we loop over all its tasks and compute the $f(tn)$ value in the same way as we did for the initial task network (lines 13-19). After that, we add all the decompositions to the fringe except for $V$, which are the nodes that are already expanded, i.e., decomposed (line 20). The reason for excluding $V$ is to avoid cycles, i.e., avoid decomposing a task network that we already decomposed. Examining whether a node, i.e., a task network was already decomposed requires examining whether two task networks are isomorphic. Since examining the isomorphism of two task networks is out of the scope of this paper, we refer for a discussion about cycles detection and task network isomorphism to~\cite{georgievski2015htn,holler2021loop}. The fringe is then sorted descendingly based on the expected utility of the task networks (line 21). Lastly, the algorithm adds the decompositions to the closed list $V$ to void expanding them more than once. If the algorithm reaches a point where the fringe is empty, it means that no solution is found, and failure is returned.

\begin{algorithm}[!t] 
	\caption{Compute the highest expected utility plan(s)}
	\label{alg:computeHighestExpectedUtilityPlan}
        \hspace*{\algorithmicindent} \textbf{Input:} $P_r= \langle s_0, tn_0, \overline{D}, U \rangle$ - risk-aware HTN planning problem, where $tn_0=\langle T_{n_0}, \psi \rangle$ \\
        \hspace*{\algorithmicindent} \textbf{Input:} $fringe$ - decomposition path \\
        \hspace*{\algorithmicindent} \textbf{Output:} $\pi $- Solution plan 
	\begin{algorithmic}[1]
            \ForAll{$v_{\overline{t}} \in T_{n_0}$}
                \If{$v_{\overline{t}}$ is primitive}
                    \State $g(tn_0) += EU(v_{\overline{t}})$
                \ElsIf{$v_{\overline{t}}$ is compound}
                    \State $h(tn_0) += max_{comp} EU(v_{\overline{t}})$
                \EndIf
            \EndFor
            \State $f(tn_0) \leftarrow g(tn_0) + h(tn_0)$
            \State V $\leftarrow$ fringe $\leftarrow$ \{$tn_0$\}
            \While{fringe $\neq \phi$}
                \State $tn_{current} \leftarrow$ \Call{popFirst}{fringe}
                \If{$tn_{current}$ is primitive $\land$ $tn_{current}$ is executable}
                    \Return $tn_{current}$
                \EndIf
                \State decompositions $\leftarrow$ \{$tn': tn_{current} \xrightarrow[\text{t,m}]{} {}_D tn'$\}
                \ForAll{dec in decompositions}
                    \ForAll{$v_{\overline{t}} \in dec$}
                        \If{$v_{\overline{t}}$ is primitive}
                            \State $g(dec) += EU(v_{\overline{t}})$
                            \ElsIf{$v_{\overline{t}}$ is compound}
                            \State $h(dec) += max_{comp} EU(v_{\overline{t}})$
                        \EndIf
                    \EndFor
                \State $f(dec) \leftarrow g(dec) + h(dec)$
                \EndFor
                \State fringe $\leftarrow$ fringe $\cup$ (decompositions\textbackslash V)
                \State \Call{sortDescendingly}{fringe}
                \State $V \leftarrow V \cup$ decompositions
                
            \EndWhile
        \Return fail
	\end{algorithmic}
\end{algorithm}

In~\ref{app:formal-properties}, we present proofs of the formal properties of Algorithms~\ref{alg:computeUtilities} and~\ref{alg:computeHighestExpectedUtilityPlan}. Specifically, we demonstrate that Algorithm~\ref{alg:computeUtilities} terminates and that the proposed heuristic is admissible. Furthermore, we establish that Algorithm~\ref{alg:computeHighestExpectedUtilityPlan} is both sound and complete, and computes optimal solutions. Additionally, we provide a complexity analysis for this algorithm.

\section{On Solving Risk-Aware State-Based HTN Planning}
\label{sec:onSolving-state-based}

In state-based HTN planning, the current state of the world is tracked at each planning step. Having the state at hand is useful to solve risk-aware state-based HTN planning for two reasons. First, we can adapt existing approaches that use state-based heuristics to guide HTN planning towards cost-optimal plans to solve risk-aware state-based HTN planning problems. Second, tracking the state during planning allows extending our work to planning agents that can express their preferences with respect to the current state of the world. For example, being in a particular state can make a risk-averse agent have a different utility  from that of a risk-seeking agent. 

Also, in the case of risk-aware state-based HTN planning, we resort to a heuristic approach and turn to A*~\cite{holler2018generic,holler2020htn}. In the scope of state-based HTN planning, one can adopt a generic method for guiding the search process by using an arbitrary classical heuristic~\cite{holler2018generic,holler2020htn}. The method is based on relaxing the HTN planning problem into a classical planning problem, which is used to calculate the heuristics. The relaxed model contains two types of actions: actions that are converted from HTN methods $A_M$ and the original actions that exist in the HTN problem $A$. The heuristics calculated in the relaxed model estimate the number of steps needed to reach the goal, i.e., the sum of decompositions and actions needed to reach the goal for each node.\footnote{A search node consists of three elements: the current state, a network of tasks that still need to be processed, and the sequence of actions included so far in the plan.} 

The relaxed model can be used to create admissible heuristics for state-based HTN planning, which can then be used to find optimal plans. This is exactly the feature that makes the present approach suitable also for solving state-based risk-aware HTN planning problems. In particular, it has been suggested that an admissible heuristic could be computed by introducing action costs in the relaxed model, where all converted actions $A_M$ could be given a zero cost, while the original actions $A$ could be given an arbitrary positive costs~\cite{holler2018generic}. Then, we could find optimal plans by employing the A* algorithm in combination with an admissible classical heuristic (e.g., LM-cut~\cite{helmert2009landmarks}) for the relaxed model~\cite{behnke2019finding}.

We propose to solve the risk-aware state-based HTN planning problem as a maximisation problem using Algorithm~\ref{alg:findPlans}. The algorithm uses A* to search for plans and takes as an input the fringe, which represents all search nodes $n$ explored together with the value that A* uses to order these nodes in the fringe, and the domain description $D$, and the problem description $P$. The value that A* uses represents an estimation of the expected utility of the plan that can result after decomposing the task network of $n$. It is computed for each search node $n$ as the sum of two values: the first one is the sum of the expected utilities of all operators that are added to the plan so far $sum(n'. \pi)$; the second one is an admissible heuristic that estimates the expected utility of plan segments computed by the relaxed model to guide the search in state-based HTN planning by \textsc{computeRCAdHeur} in Algorithm~\ref{alg:computeRCAdHeur}). The heuristic is computed on the relaxed model after setting the cost of method actions $A_M$ to zero, and assigning the original actions $A$ a cost equal to the expected utilities of the corresponding operators in the domain description (lines 5 and 6 in Algorithm~\ref{alg:computeRCAdHeur}). The expected utility of an operator is computed as the weighted average of the utilities of each possible cost of the operator. The weights represent the probability distribution of these costs.

Upon initialization of Algorithm~\ref{alg:findPlans}, the fringe contains an initial node that consists of the initial state, initial task network, and an empty plan. An admissible heuristic is then computed (line 2) and the node is added to the fringe. The algorithm keeps looping to process all the nodes in the fringe until the fringe is empty. At each iteration, a test is made on whether a plan is created, i.e., all tasks in the node's task network are primitive tasks and the node's plan is applicable at the initial state and can accomplish the initial task network (lines 7 and 8). If this is the case, the plan is returned. Otherwise, the set of all tasks that do not have predecessors in the node's task network are returned (line 9). For each of these returned tasks, if the task is primitive, it is applied and added to the node's plan (line 12). Then, the heuristic of the node is computed again and the node is added to the fringe according to its annotated value (lines 13 and 14). If the task is compound, for all methods that can decompose it and are applicable, a new search node is generated, which equals to the current search node, but with replacing the compound task in the search node's task network by the tasks in the method's task network. After that, the node is added to the fringe in its right order after computing its value (lines 18 and 19).

\begin{algorithm}[t]
	\caption{FindPlans}
	\label{alg:findPlans}
	\begin{algorithmic}[1]
	\State $n_{init}$ $\leftarrow $ ($s_0$, $tn_1$, $\phi$)
	\State $n_{init}$.heuristic $\leftarrow$ \Call{computeRCAdHeur}{$n_{init}$ ,D, P}
	\State fringe $\leftarrow$ \{ $n_{init}$ \}
		\Function{FindPlans}{fringe, D, P}
		\While {fringe $\neq \phi$ }
		\State $n \leftarrow$ fringe.first()
		\If{$\forall$ t $\in$ n.tn : isPrimitive(t) $\land$ applicable(n.$\pi$, n.$s_0$) }
		\State return n.$\pi$
		\EndIf
		\State U $\leftarrow$ \Call{findUnconstrainedTasks}{n.tn}
		\ForAll{t $\in$ U}
		\If{isPrimitive(t) $\land$ $\exists$ o $\in$ D.O: pt(o) = t $\land$ pre(o) $\subseteq$ n.s }
		\State n' $\leftarrow$ n.apply(t)
		\State n'.heuristic $\leftarrow$ sum(n'.$\pi$)+\Call{computeRCAdHeur}{n',D,P}
		\State fringe.addAndOrder(n')
		
		\Else
		\ForAll{m $\in$ D.M :ct(m) =t $\land$ pre(m) $\subseteq$ n.s}
		\State n' $\leftarrow$ n.decompose(t,m)
		\State n'.heuristic $\leftarrow$ sum(n'.$\pi$)+\Call{computeRCAdHeur}{n',D,P}
	    \State fringe.addAndOrder(n')
		\EndFor
		\EndIf
		\EndFor
		\EndWhile
		\State return Failure
		\EndFunction
	\end{algorithmic}
\end{algorithm}

\begin{algorithm}[t]
	\caption{computeRCAdHeur}
	\label{alg:computeRCAdHeur}
	\begin{algorithmic}[1]
		\Function{computeRCAdHeur}{n, D, P}
		
		\State RC $\leftarrow$ \Call{buildRCModel}{n, P}
		\ForAll{$A_m$ $\in$ RC}
		\State $A_m$.setCost(0)
		\EndFor
		
		\ForAll{$A$ $\in$ RC : $\exists$ o $\in$ D.O $\land$ pt(o) = A}
		\State $A$.setCost(\Call{ComputeExpectedUtility}{P.utilityFunction, o})
		\EndFor
		\State AdmissableH $\leftarrow$ \Call{computeH}{RC}
		\State return AdmissableH
		\EndFunction
	\end{algorithmic}
\end{algorithm}

\section{Experimental Evaluation}
\label{sec:experiments}
To empirically evaluate the feasibility of the proposed risk-aware plan-based HTN planning approach and demonstrate how several solutions are generated for the various risk attitudes to better represent the actual goals of the users, we conduct an evaluation process structured into three key aspects: (1) Implementing the risk-aware plan-based HTN planning algorithm (see Section~\ref{sec:solving-plan-based}), (2) modelling HTN planning domains characterised by risk and uncertainty compatible with the risk-aware HTN planning framework, and (3) designing a suitable experimental setup. Regarding (1), we implemented the risk-aware plan-based HTN planning algorithm and integrated it into the PANDA planner~\cite{holler2021panda} to expand its functionality to support solving risk-aware HTN planning problems. Next, we provide details about aspects (2) and (3).

\subsection{Overview of HTN Planning Domain Models}
To address aspect (2), we model two HTN planning domains: an Autonomous Vehicles (AV) domain and a Satellite domain. Both these domains represent realistic environments characterised by risk and uncertainty. The domains are modelled using our extension of the Hierarchical Domain Definition Language (HDDL)~\cite{holler2020hddl}, which allows modelling a probability distribution of operator costs. A detailed analysis of the uncertainty sources within these domains, their impact on the variability of operator costs, and the rationale behind the probability distributions assigned to operators, as well as details on the domain models and their hierarchical structures, are provided in~\cite{alnazer2025domains}.

The Autonomous Vehicles (AVs) domain includes various driving tasks an autonomous vehicle must handle to navigate routes and reach required destinations successfully. This includes handling incidents, moving (e.g., pedestrians) and still (e.g., construction works), handling road conditions (e.g., slippery roads), turning on lights, and choosing specific routes (e.g., shortcuts and highways). The domain offers a variety of options to agents with different risk attitudes. For example, for driving on roads in bad conditions, the vehicle can accelerate without caring about the road condition, only decelerate, activate the Electronic Stability Program (ESP) and decelerate, or activate the ESP but accelerate. Choosing the first option is the riskiest due to the high probability of slipping on the road. In the present treatment, we consider the effects of three sources of uncertainty, namely (1) the speed at which the pedestrians walk the pedestrian crossing, which is considered a regular external source of uncertainty, (2) the traffic on roads, which is also considered an external regular source of uncertainty, and (3) the ability of the autonomous vehicle to stabilise on slippery roads, which is considered an internal regular source of uncertainty.

The Satellite domain model was first featured in the International Planning Competition of 2020 (IPC-2020) as one of the partial-order set of benchmark domains for HTN planning~\cite{schattenberg2020hierarchical}. It is about satellite coordination for the goal of remote imaging, and we extend it to bring the domain closer to reality and to demonstrate the differences in the behaviours of planning agents with varying risk attitudes~\cite{alnazer2022bringing}. The idea behind extending this domain is to enable satellites to power two or three instruments that are hosted onboard simultaneously~\cite{powell2024sampled}. This can be done under the risk of a power system failure or power loss if the satellite's power source (e.g., solar panels, Radioisotope Thermoelectric Generators, or he1 cells) is unable to provide the power required~\cite{morgan2005fault}. As a result, time might be lost as the satellite has to recover from the power failure.

\subsection{Experiments Setup}
Regarding aspect (3), we design our experiments as follows. For the Autonomous Vehicles (AVs) domain, we design a set of experiments involving five problem instances with increasing complexity in terms of the number of roads and road obstacles and conditions. For each problem instance, we use the utility functions from Equation~\ref{equ:exponentialUtilityFunction}, where, for each of the risk-seeking and risk-averse attitudes, we set $\alpha = 0.1$ and $\alpha= 0.9$ to assess the behaviour of the low and high-intensity risk attitudes, respectively. We also set $a=1$ and $a=-1$, for the risk-seeking and risk-averse, respectively. For the risk-neutral attitude, we use a linear utility function. This results in running each problem instance with five settings related to the risk attitudes and their intensities, $RS(0.1), RS(0.9), RA(0.1), RA(0.9)$, and $RN$, totalling 25 experiments. 

The five problem instances have the following increasing complexities with the number of roads (R), locations (L), obstacles (I), moving obstacles (M), still obstacles (S), and roads with bad conditions (B):
\begin{itemize}
    \item \textbf{P1}: R=8, L=7, I=2 (M=1, S=1 ), B=1.
    \item \textbf{P2}: R=16, L=13, I=3 (M=2, S=1), B=2. 
    \item \textbf{P3}: R=41, L=33, I=8 (M=4, S=4), B=4. 
    \item \textbf{P4}: R=64, L=52, I=14 (M=8, S=6), B=5. 
    \item \textbf{P5}: R=76, L=63, I=16 (M=8, S=8), B=8. 
\end{itemize}

In the Satellite domain, we design a set of experiments involving sixteen problem instances with varying complexity in terms of the number of celestial observations to be made, modes (hence instruments), and satellites. For example, increasing the number of observations and/or modes makes the problem more complex. However, when maintaining a consistent number of observations and modes, introducing more satellites actually simplifies solving the planning problem. This is because when instruments are spread across a larger number of satellites, the frequency of decisions regarding the parallel operation of instruments is reduced. Problem instances are denoted as \textit{RA-xobs-ysat-zmod}, where \textit{x}, \textit{y}, and \textit{z} denote the number of observations, satellite, and modes, respectively. For each problem instance, we use the utility functions from Equation~\ref{equ:planExpectedUtility-segmentation}, where, for both risk-seeking and risk-averse attitudes, we set $\alpha = 0.5$, while we set $a=1$ for risk-seeking attitude and $a=-1$, for risk-averse attitude. For the risk-neutral attitude, we use a linear utility function. This results in running each problem instance with three settings related to the three risk attitudes, totalling 48 experiments.

For each experiment in both domains, we collect the plan's expected utility (\textbf{EU}), the logarithmic value of the plan's expected utility ($\mathbf{log_{10}(EU)}$) computed using Equation~\ref{equ:plan_expected_utility-log}, plan's expected cost (\textbf{EC}), plan length (\bm{$|\pi|$}), and planning time (\textbf{PT}) in seconds, averaged over 10 runs. The experiments were performed on a computer running Windows 11 64-bit with a (Intel(R) Core(TM) i7-8565U CPU @ 1.80GHz) processor and 16,0GB of RAM.

\subsection{Results}
The results of solving five problem instances for the Autonomous Vehicles domain and sixteen problem instances for the Satellite domain are shown in Tables~\ref{tab:results} and~\ref{tab:results-sat}. Since the expected utility values can be very long, we also show their logarithmic values. In~\ref{app:planning-problems}, we present a comprehensive discussion and analysis of the results obtained from solving the first problem instance in both domains.

For the domain of the Autonomous vehicle, let us first focus on the results for \textbf{P1}, which is illustrated in Figure~\ref{fig:problem-instance} and explained in detail in~\ref{app:avs-problem-instance}. The extreme risk-averse agent with $RA(0.9)$ chooses to take the $S, l_1, l_2, l_4, E$ route, which is the only plan with guaranteed outcomes for all its constituent actions, i.e., it is the plan with the least risk compared to all plans with the same expected value. On the other hand, the risk-seeking agents with $RS(0.9)$ and $RS(0.1)$ choose to take the $S, l_3, l_5, l_4, E$, which has a risky non-guaranteed outcome. For these agents, choosing the short road ($S$, $l_1$) is preferable over the long road ($S$, $l_3$). The reason is that, although the short road is riskier than the long road, it has a chance of resulting in a better outcome, i.e., saving two hours compared to the long road. The risk-averse agent with $RA(0.1)$ is less sensitive to risky choices and chooses the same route as the risk-seeking attitude. Additionally, the extreme risk-seeking agent chooses to accelerate on the icy road without activating the ESP. This action is riskier than all other options, but with a 20\% probability, it might result in a better outcome, i.e., spending six hours on the icy road instead of a guaranteed ten hours. On the other hand, the risk-averse agents with $RA(0.9)$ and $RA(0.1)$ choose to decelerate on the icy road after activating the ESP, which is the safest option. The risk-seeking agent with $RS(0.1)$ takes an in-between choice, i.e., activating the ESP but accelerating after that. We see that no agent chooses to decelerate without activating the ESP since, with a probability of 20\%, this choice saves four hours compared to accelerating after activating the ESP, while with a probability of 80\%, this choice takes only one hour more than accelerating. Finally,  the risk-neutral agent chooses to take the $S, l_3, l_5, l_4, E$ as it has a lower expected cost. However, it is indifferent to the options on the icy road, except for the option of accelerating without activating the ESP, which has an average cost that is higher than other options. Thus, this risk attitude leads to multiple optimal plans.

For the Satellite domain, we focus first on the results for \textbf{RA-3obs-1sat-3mod}, which is explained in detail in~\ref{app:satellite-problem-instance}. In this problem instance, the risk-seeking attitude chooses to activate all three instruments at once, which is considered the riskiest choice. On the opposite, the risk-averse attitude avoids simultaneous operations of instruments and prioritises a plan that has a guaranteed cost
even if, on average, this plan entails higher expected costs and is longer compared to the alternatives. The risk-neutral attitude chooses a balanced solution. That is, it activates two instruments in parallel but then switches them off before activating the third one to complete the final observation.

\begin{table}[H]
\centering
\footnotesize
\caption{Results of 25 experiments in the AVs domain. For each plan, we show the plan's expected utility (\textbf{EU}), the logarithmic value of the plan's expected utility ($\mathbf{log_{10}(EU)}$), plan's expected cost (\textbf{EC}), plan length (\bm{$|\pi|$}), and planning time (\textbf{PT}) in seconds, averaged over 10 runs.} 
\label{tab:results}
\begin{tabular}[h]{|l|l|l|l|l|l|l|}
\hline
\textbf{\#P}                 & \textbf{Attitude}    & \textbf{EU }          & $\mathbf{log_{10}(EU)}$ &     \textbf{EC}     &  $\bm{|\pi|}$ & \textbf{PT (s)} \\ \hline
\multirow{5}{*}{P1}          & RS (0.1)             & 1,793       & 0,254          &   -18,35   &  9      & 3,59     \\ \cline{2-7} 
                             & RS (0.9)             & 0,117905E-3  & -3,928         &   -22,8    &  8      & 6,67     \\ \cline{2-7} 
                             & RA (0.1)             & -64,159    & -1,807         &   -18,35   &  9      & 4,78     \\ \cline{2-7} 
                             & RA (0.9)             & -42514702,52 & -7,629        &   -44,08   &  9      & 4,55     \\ \cline{2-7} 
                             & RN                   & -18,35       & -              &   -18,35   &  8      & 3,64     \\ \hline
\multirow{5}{*}{P2}          & RS (0.1)             & 0,795       & -0,1          &   -26,7    &  13     & 3,66     \\ \cline{2-7} 
                             & RS (0.9)             & 2,68168E-07   & -6,572        &   -30,66   &  12     & 5,18     \\ \cline{2-7} 
                             & RA (0.1)             & -137,873    & -2,139        &   -25,9    &  13     & 3,23     \\ \cline{2-7} 
                             & RA (0.9)             & -5,6946E+10   & -10,755        &   -27,4    &  11     & 4,9      \\ \cline{2-7} 
                             & RN                   & -25,41       & -              &   -25,41   &  13     & 7,42     \\ \hline
\multirow{5}{*}{P3}          & RS (0.1)             & 0,375       & -0,426         &   -33,7     &  18     & 6,84     \\ \cline{2-7} 
                             & RS (0.9)             & 5,71832E-13   & -12,243        &   -36,95 &  14     & 7,06     \\ \cline{2-7} 
                             & RA (0.1)             & -334,285   & -2,524         & -34,1  &  18     & 6,96     \\ \cline{2-7} 
                             & RA (0.9)             & -4,42359E+17 & -17,646        & -37,95  &  20     & 6,84     \\ \cline{2-7} 
                             & RN                   & -33,7        & -              &  -33,7 &  18     & 6,65     \\ \hline
\multirow{5}{*}{P4}          & RS (0.1)             & 0,0266       & -1,575          &  -61,7 &  26     & 13,07    \\ \cline{2-7} 
                             & RS (0.9)             & 3,52078E-18   & -17,453         &  -73,45 &  21     & 11,86    \\ \cline{2-7} 
                             & RA (0.1)             & -5718,583     & -3,757         &   -61,7 &  26     & 13,67    \\ \cline{2-7} 
                             & RA (0.9)             & -7,34207E+31 & -31,866        &   -69,75 &  30     & 15,71    \\ \cline{2-7} 
                             & RN                   & -61,7        & -              &   -61,7 &  25     & 13,29    \\ \hline
\multirow{5}{*}{P5}          & RS (0.1)             & 0,001 & -3,199         &  -100,1 &  32     & 242,85   \\ \cline{2-7} 
                             & RS (0.9)             & 2,15427E-24   & -23,667        &  -125,2 &  27     & 53,35    \\ \cline{2-7} 
                             & RA (0.1)             & -287089,873  & -5,458        &   -100,5 &  36     & 241,87   \\ \cline{2-7} 
                             & RA (0.9)             & -4,18505E+47   & -47,622        &   -108,55 &  29     & 306,37   \\ \cline{2-7} 
                             & RN                   & -97,3        & -              &  -97,3 &  30     & 242,14   \\ \hline
\end{tabular}
\end{table}

{\footnotesize
\begin{longtable}[H]{|l|l|l|l|l|l|l|}
\caption{Results of the experiments in the Satellite domain. For each planning problem, we show the risk attitude, the plan's expected utility (\textbf{EU}), the logarithmic value of the EU (\bm{$log_{10}(EU)$}), the plan's expected cost (\textbf{EC}), the plan expected cost (\textbf{EC}), the plan length (\bm{$|\pi|$}), and planning time (\textbf{PT}) in seconds.} 
\label{tab:results-sat} \\
\hline 
\textbf{problem}                            & \textbf{attitude} & \textbf{EU}              &    \bm{$log_{10}(EU)$}     &      \textbf{EC}      & \bm{$|\pi|$}   &    \textbf{PT (s)} \\ \hline
\multirow{3}{*}{RA-3obs-1sat-3mod} & RS       &  2,60771E-49    &       -48,884         &    -245,1    &    15     &    4,098    \\ \cline{2-7} 
                                   & RA       &  -4,71603E+55   &       -55,674         &    -255      &    17     &    5,457    \\ \cline{2-7} 
                                   & RN       &  -240,85        &          -            &    -240,85   &    16     &    4,187     \\ \hline
\multirow{3}{*}{RA-4obs-1sat-3mod} & RS       &  7,97705E-56    &        -55,098       &    -275,1    &    17     &    25,575      \\ \cline{2-7} 
                                   & RA       &  -1,64751E+75   &        -75,217        &    -345      &    23     &    26,866      \\ \cline{2-7} 
                                   & RN       &  -275,1         &            -          &    -275,1    &    17     &    36,772      \\ \hline
\multirow{3}{*}{RA-4obs-1sat-4mod} & RS       &  7,77892E-69    &        -68,109         &    -316,7    &    21     &    8,149      \\ \cline{2-7} 
                                   & RA       &  -1,64751E+75   &        -75,217        &    -345      &    23     &    5,2      \\ \cline{2-7} 
                                   & RN       &   -316,7        &           -           &    -316,7    &    21     &    8,366     \\ \hline
\multirow{3}{*}{RA-5obs-1sat-3mod} & RS       &   2,4402E-62    &        -61,613       &    -305,1    &    19     &    10,359      \\ \cline{2-7} 
                                   & RA       &   -5,75546E+94  &        -94,76       &    -435      &    29     &    9,423      \\ \cline{2-7} 
                                   & RN       &   -305,1        &          -            &    -305,1    &    19     &    10,604       \\ \hline
\multirow{3}{*}{RA-5obs-1sat-4mod} & RS       &   2,28345E-75   &        -74,641       &    -365,1    &    23     &    5,081     \\ \cline{2-7} 
                                   & RA       &   -5,75546E+94  &        -94,76        &    -435      &    29     &    5,283       \\ \cline{2-7} 
                                   & RN       &   -365,1        &           -           &    -365,1    &    23     &    5,370     \\ \hline
\multirow{3}{*}{RA-5obs-2sat-3mod} & RS       &    2,56863E-62 &        -61,59        &    -285,85   &    19    &   3,74    \\ \cline{2-7} 
                                   & RA       &   -2,97877E+78 &        -78,474        &    -360      &   24     &   5,483     \\ \cline{2-7} 
                                   & RN       &    -285,85    &        -              &  -285,85      & 19        &  3,884      \\ \hline
\multirow{3}{*}{RA-5obs-2sat-4mod} & RS       &    4,12857E-72 &    -71,384            &  -350,1     &  22    &   6,714      \\ \cline{2-7} 
                                   & RA       &    -3,18325E+91 &     -91,503         &    -420     &   28     &  8,119      \\ \cline{2-7} 
                                   & RN       &    -350,1     &        -           &   -350,1      &     22   &   7,877    \\ \hline         
\multirow{3}{*}{RA-5obs-2sat-5mod} & RS       &   6,91526E-82   &        -81,16        &    -395,95   &    25     &    6,463     \\ \cline{2-7} 
                                   & RA       &   -3,18325E+91  &        -91,502        &    -420      &    28     &    4,310      \\ \cline{2-7} 
                                   & RN       &   -391,7        &           -           &    -391,7    &    26     &    6,537     \\ \hline
\multirow{3}{*}{RA-6obs-2sat-3mod} & RS       &   7,8575E-69    &        -68,105        &    -315,85   &    21     &    3,726      \\ \cline{2-7} 
                                   & RA       &   -9,73765E+84  &        -84,989        &    -390      &    26     &    3,556     \\ \cline{2-7} 
                                   & RN       &   -315,85       &           -           &    -315,85   &    21     &    3,776     \\ \hline
\multirow{3}{*}{RA-6obs-2sat-4mod} & RS       &   1,26294E-78   &         -77,8989         &    -380,1    &    24     &    10,687      \\ \cline{2-7} 
                                   & RA       &   -1,112E+111   &         -111,046      &    -510      &    34     &    17,343      \\ \cline{2-7} 
                                   & RN       &   -380,1        &           -           &    -380,1    &    24     &    11,305     \\ \hline
\multirow{3}{*}{RA-7obs-2sat-4mod} & RS       &    4,02602E-85  &        -84,395        &   -391,7  &      26   &      5,288    \\ \cline{2-7} 
                                   & RA       &    -3,18325E+91 &        -91,503        &    -405     &    28     &    4,272     \\ \cline{2-7} 
                                   & RN       &     -391,7     &           -           &    -391,7   &     26   &     5,657    \\ \hline  
\multirow{3}{*}{RA-7obs-2sat-5mod} & RS       &    6,47104E-95  &         -94,189       &   -455,95  &     29    &    135,919      \\ \cline{2-7} 
                                   & RA       &    -3,6353E+117 &         -117,561       &   -540      &   36      &   86,504      \\ \cline{2-7} 
                                   & RN       &     -451,7     &           -           &   -451,7    &     30   &    143,845     \\ \hline  
\multirow{3}{*}{RA-7obs-2sat-6mod} & RS       &    1,0401E-104  &        -103,983        &  -520,2   &     32    &    6,497      \\ \cline{2-7} 
                                   & RA       &    -3,8849E+130 &         -130,589       &   -600      &   40      &   6,779      \\ \cline{2-7} 
                                   & RN       &     -511,7     &           -           &    -511,7   &    34    &    8,073     \\ \hline                                     
\multirow{3}{*}{RA-7obs-3sat-4mod} & RS       &    4,06669E-85  &         -84,391       &    -390,85   &    26     &    3,293       \\ \cline{2-7} 
                                   & RA       &   -1,76061E+88  &         -88,246       &    -405      &    27     &    3,209     \\ \cline{2-7} 
                                   & RN       &    -390,85      &           -           &    -390,85   &    26     &    2,862      \\ \hline
\multirow{3}{*}{RA-7obs-3sat-5mod} & RS       &   6,81162E-95   &         -94,167      &    -436,7    &    29     &    4,064      \\ \cline{2-7} 
                                   & RA       &    -1,8815E+101 &         -101,275      &    -465      &    31     &    3,8     \\ \cline{2-7} 
                                   & RN       &   -436,7        &           -           &    -436,7    &    29     &    4,195     \\ \hline
\multirow{3}{*}{RA-7obs-3sat-6mod} & RS       &   1,0948E-104   &         -103,96     &    -500,95   &    32     &    9,417     \\ \cline{2-7} 
                                   & RA       &   -2,0106E+114  &         -114,303      &    -525      &    35     &    6,329      \\ \cline{2-7} 
                                   & RN       &   -496,7        &           -           &    -496,7    &    33     &    9,868     \\ \hline
\end{longtable}}

\subsection{Discussion}
In examining the overall results of the Autonomous vehicles domain, a trend emerges: the expected utility of extremely risk-seeking agents significantly surpasses that of the agents with $RS(0.1)$ across all experiments. This is an expected result as the more risk-seeking the agent is, the more it underestimates the action costs. Contrarily, we see that extremely risk-averse agents have much lower expected utilities than those with $RA(0.1)$. This suggests that agents with $RA(0.9)$ tend to overestimate action costs, shying away from risky options, even when the probability of consuming resources is minimal. 

Across all risk attitudes in both domains, a notable observation is the decline in the expected utility of the plan with the increase in the complexity of the planning problem. This phenomenon is attributed to the greater number of risky choices that the agent has to make during planning with the increase of the complexity of the planning process. The sign of the expected utility is tight to the sign of the utility functions we use for the different risk attitudes, being positive for risk-seeking and negative for risk-averse attitudes. 

Comparing the expected utilities among the various risk attitudes in both domains does not reveal any differences in behaviours. This is because every risk attitude computes a different plan. A fair comparison requires enumerating all feasible plans for each problem instance and showing how each risk attitude chooses the one with the highest expected utility according to its risk attitude. We perform such a comparison for the first problem instances in both domains, as the relatively small search space in these cases makes it feasible to illustrate the concept. Specifically, the detailed comparison is performed on \textbf{P1} of the Autonomous Vehicles domain (see~\ref{app:avs-problem-instance}) and \textbf{RA-3obs-1sat-3mod} in the Satellite domain (see~\ref{app:satellite-problem-instance}). 

Turning to the plan length, in both domains, there are discrepancies between the risk attitudes. That is, distinct risk attitudes yield different plan lengths. However, in the Satellite domain, the risk-neutral attitude in this domain either computes identical plans to one of the other risk attitudes or generates a unique plan. Interestingly, plans computed by the agents with $RS(0.9)$ and risk-seeking agents in the Autonomous vehicles and Satellite domains, respectively, are generally shorter. Looking at the specific solutions, we observe that this is due to the fact the agents with $RS(0.9)$ tend to choose shortcut routes that are riskier than other routes. Similarly, in the Satellite domain, the risk-seeking attitude always chooses to go with the risky choices of operating instruments in parallel, which leads to computing the shortest possible plan. Conversely, the risk-averse attitude plans computed by the risk-averse attitude tend to be longer since this risk attitude avoids risky choices. In the Autonomous Vehicles domain, this entails avoiding shortcuts or other risky choices, which can lead to taking longer roads. In the Satellite domain, this entails avoiding the operation of parallel instruments, and thus, each time it switches on an instrument, any activated instrument onboard has to be switched on, adding an extra action. Additionally, this risk attitude loses the advantage of being able to switch instruments off at once, which contributed to computing shorter plans. In the Satellite domain, the risk-neutral attitude either computes plans that have a length in between the length of plans computed by the other risk attitudes or computes plans that are identical to the risk-seeking attitude and, hence, have the same length. In the former case, the resulting plans alternate between activating instruments in parallel and switching instruments on and off sequentially. 

Additionally, the plan length for all risk attitudes in both domains increases when solving more complex problems. In the Autonomous Vehicles domain, this is attributed to the involvement of more roads, locations, and decisions are involved. Similarly, in the Satellite domain, when increasing the number of modes while fixing the other two parameters increases the plan length since this requires calibrating more instruments. Similarly, increasing the number of observations increases the complexity and the plan lengths.

In both domains, we notice that the complexity of planning problems is reflected in the corresponding increase in planning time, with some exceptions in the Satellite domain. In the Autonomous Vehicles domain, planning time grows with the complexity of the problem but remains under five minutes even for the most complex. Similarly, in the Satellite domain, planning times are generally minimal, not exceeding three minutes. This illustrates the feasibility of the proposed approach.

Comparing the expected cost of plans between the risk-seeking and risk-averse attitude, we notice no occurring pattern regarding which attitude tends to incur higher expected costs in the Autonomous Vehicles domain. However, in the Satellite domain, we can see that plans computed by the risk-averse attitude have higher expected costs than the other two risk attitudes. This is because, in this domain, the risk-averse approach prioritises avoiding risky actions and instead opts for guaranteed solutions, even if they entail higher expected costs. Intuitively, in both domains, plans computed by the risk-neutral attitude have the smallest expected costs since this is the optimisation criteria for this risk attitude.

\subsection{Open Issues}

The empirical results demonstrate how the existence of uncertainty sources can affect the variability of costs, leading to different planning behaviours among the various risk attitudes. Further implications and extensions that are left for future exploration include the following.  

\subsubsection{Further experiments}

In the current evaluation of the AVs domain, we assume that the vehicle is always charged while navigating the roads. This allows us to focus on demonstrating the different risk attitudes on some planning choices. A possible extension of the problem instances, which increases their complexity, is to assume that the vehicle has a limited charging level and has to recharge whenever there is not enough power. This extension can come with further planning choices that the vehicle should make, such as selecting a charging station, determining the route to reach it, deciding the amount of charge to replenish at each station, and choosing the battery level threshold at which to initiate recharging. When confronted with such choices, risk-seeking agents might tend not to recharge the vehicle more often to reach the destination by avoiding detouring as much as possible but at the cost of depleting the battery in the middle of the road. Similarly, the risk-seeking attitude might opt not to fully recharge the vehicle, prioritising time savings despite the risk of depleting the battery and taking more time to recharge the vehicle and get it operational again. Conversely, the risk-averse attitude may lean towards caution, choosing to recharge the vehicle to its maximum capacity. It could also have a higher threshold for the charging level at which it decides to recharge compared to the risk-seeking attitude. The route choices and charging behaviours of different risk attitudes is a topic that is studied a lot in the literature about electric vehicles (e.g.,~\cite{yang2016modeling}).

In the Satellite domain, in addition to the experiments shown in Table~\ref{tab:results-sat}, we conduct further experiments on the same planning problems but with different intensities of the risk-averse and risk-seeking attitudes by changing the $\alpha$ value in the expected utility function. Particularly, we try with two values of $\alpha=0.1$ and $\alpha=0.9$. In this particular domain, changing the intensity of the risk attitude does not yield any different results, with a few exceptions in the risk-averse attitude. These exceptions appear when having a low-intensity risk-averse attitude (with $\alpha=0.1)$, where the risk-averse agent chooses to activate two instruments in parallel. However, activating three instruments in parallel is not chosen even with this low intensity of the risk attitude. Moreover, we conduct further experiments with five or more observations, one satellite, and three or more modes. These experiments did not yield any results due to memory overhead exceptions. This might be attributed to the large search space of such planning problems and the complexity of the planning problem since only one satellite is equipped with all the instruments and has to perform all the observations. When increasing the number of satellites to two or more, the planning problem becomes solvable, as shown in Table~\ref{tab:results-sat}.

 \subsubsection{Risk attitude dynamics} 
 
In our approach, we consider that the agent has a static risk attitude that does not change during planning. Another approach is to consider the dynamics of the agent's risk attitude, as discussed in Section~\ref{subsubsec:dynamic-risk-attitudes-HTN}, where the agent has a utility function that expresses the ability to switch from one risk attitude to another depending on some factor, such as the available resources. An example of such a utility function is the one-switch utility function of Equation~\ref{equ:one-switch}. In the AVs domain, the agent's risk attitude can change based on, for example, the remaining charge and/or time left to reach the destination. In this case, it can happen that when the remaining charge is under a certain threshold, the agent becomes more cautious and changes its attitude from risk-seeking to risk-averse or becomes more adventurous and changes from risk-averse to risk-seeking by taking risks, underestimating the bad outcomes, and hoping to get to the destination with the best possible outcomes. The risk attitude can also change based on the history of the agents and the outcomes resulting from past choices. This requires interleaving planning with execution. For example, let us assume that the agent is risk-seeking, and when solving the planning problems in the AVs domain, it tends to choose short but risky roads that might have high traffic jams and result in long driving times. If, during execution, the choice of these roads really led to long waiting times, after a certain amount of bad outcomes, the agent might change its risk attitude to risk-averse and make the remaining planning choices based on this risk attitude. Similarly, in the Satellite domain, the agent's risk attitude can change based on the amount of remaining power levels. When power drops below a certain threshold, the agent can become more cautious in its choices, adopting a risk-averse attitude. Alternatively, the agent can become more advantageous, adopting a risk-seeking attitude, realising that always following a risk-averse strategy in space missions might prevent achieving scientific goals if those goals are considered too risky~\cite{mimlitz2016toward}.

\subsubsection{Risk-awareness in a multi-agent system}

In the current approach, we treat other vehicles in the AVs domain as external moving obstacles regardless of their driving patterns that can be affected by their risk attitude. If the autonomous vehicle needs to adapt to the risk attitudes of other vehicles, it should be capable of adjusting its own risk attitude based on those of the surrounding vehicles. That is, the vehicle should have a dynamic risk attitude. Adapting the risk attitude of the autonomous vehicle based on other vehicles can be crucial, particularly if the surrounding vehicles share the same risk attitude as the AV. This alignment in risk attitudes may lead to similar route selections, potentially causing congestion on specific routes and resulting in extended driving durations. Similarly, in the Satellite domain, coordinating multiple satellites to execute space missions introduces additional complexity. Effective coordination requires not only managing the operational tasks of each satellite but also understanding and aligning the risk attitudes of the agents planning their respective missions. Each satellite's plan may influence or depend on the actions of others, making it essential to consider how differing risk preferences, such as risk-averse or risk-seeking approaches, impact the overall mission strategy. Misalignment in these attitudes can cause mission failure because different risk approaches can lead to conflicts over important resources like power. For example, a risk-seeking agent might use a lot of power for high-risk tasks, leaving too little for a risk-averse agent to carry out its carefully planned actions. This imbalance in resource use can disrupt the mission plan, affect goals, and reduce the chances of success.

\section{Related Work}
\label{sec:uncertainty-aiplanning}

Risk and especially uncertainty have been the concern of several works on AI planning. We overview both non-hierarchical AI planning and HTN planning approaches. We start by briefly summarising works that solve planning problems under uncertainty followed by reviewing works that incorporate utilities, risk, and/or risk attitudes. Then, we review works that study HTN or HTN-like planning under uncertainty followed by an overview of studies that incorporate some form of utilities and risks in HTN or HTN-like planning. Finally, we discuss approaches that use additional information to aid the decision-making process of HTN planning.

\subsection{Uncertainty in Non-Hierarchical Planning}

Planning under uncertainty has already been the object of scientific reviews~\cite{blythe1999overview,blythe1999decision}, where uncertainty is usually defined as incomplete or faulty information from the environment, which in turn leads to uncertain initial state and action effects~\cite{kaldeli2016domain}. 

There are two types of planning under uncertainty, namely contingent planning and conformant planning. Contingent planning deals with the task of generating a \textit{conditional plan} given uncertainty about the initial state and action effects, but with the ability to observe and sense some aspects of the current world state during execution, partially or fully~\cite{hoffmann2005contingent,albore2009translation}. Conditional plans are plans that have some branches executed conditionally based on the outcome of sensory actions~\cite{smith1998conformant}. Conformant planning is the task of generating plans given uncertainty about the initial state and action effects, but without any sensing capabilities during plan execution (no observability)~\cite{hoffmann2006conformant}.

Both types of uncertainty about action outcomes are usually expressed logically using conjunctions or numerically using probabilities~\cite{majercik2003contingent,bresina2012planning}. In the first case, planning approaches should find plans that are successful regardless of which particular initial world we start from and which action effects occur, e.g.,~\cite{smith1998conformant,etzioni1992approach,peot1992conditional}. In the second case, planning approaches aim at finding plans either with the highest probability of succeeding or with a success probability that exceeds a certain threshold, e.g.,~\cite{kushmerick1995algorithm,goldman1994epsilon,gu2004improved}. 

Most approaches to AI planning do not make a clear distinction between risk and uncertainty in action outcomes as defined in decision theory. In particular, with few exceptions that we present in the next section, approaches that model action outcomes probabilistically do not incorporate the notion of risk or deal with risk attitudes of planning agents.

\subsection{Risk in Non-Hierarchical Planning}

Our approach is closely related to studies that incorporate utilities in planning, such as~\cite{etzioni1991embedding,russell1991right,haddawy1992representations,wellman1992modular,koenig1994risk,koenig1996modeling}. However, our approach is different than these approaches in incorporating risks, risk attitudes, and utilities in hierarchical constructs. These studies extend classical planning problems with concepts from utility theory and incorporate some form of utilities in planning, usually assuming a risk-neutral attitude of the planning agent. An interesting approach is presented in~\cite{koenig1994risk}, where planning problems are characterised by probabilistic effects of actions, actions with costs (resource consumption), and rewards of goal states. Similar to our work, this approach aims at finding plans with the highest expected utility for risk-sensitive agents. The agents have utility functions (linear or exponential) that are used to quantify their preferences, in terms of utility, over the outcomes. Moreover, utility functions have been incorporated in uncertain robot navigation domains to demonstrate how utility functions can be used to model given risk attitudes and soft deadlines~\cite{koenig1996modeling}. The uncertainty in this domain results in actions with varying execution times, i.e., costs, and a totally-known probability distribution. The paper discusses the incorporation of risks by using exponential utility functions to evaluate outcomes and find plans with the highest expected utility. 

\subsection{Uncertainty in HTN Planning}

A form of uncertainty has also been considered in HTN planning. In~\cite{bouguerra2004hierarchical,bouguerra2005pc,zhao2023probabilistic}, two types of uncertainty are defined. The first type entails having partial observability of the state, which is represented as a probability distribution over belief states. The second type of uncertainty is represented as a probability distribution over action effects. This is similar to our definition of risk. In~\cite{bouguerra2004hierarchical,bouguerra2005pc}, the authors do not consider action costs and aim to find conditional plans with a certain probability of success. In~\cite{zhao2023probabilistic}, the aim is to find plans with the minimum cost, where cost is defined with respect to the literals in operator preconditions. These works have goals different from ours and do not study risk attitudes.

In~\cite{chen2021fully}, fully observable non-deterministic HTN planning problems are formalised where actions have multiple effects. While this work introduces solution criteria for solving these problems, it does not consider plan quality, action costs, or risk attitudes. A more recent work follows the same formalism and does not consider plan quality beyond the minimum-cost plans, risks, or risk attitudes~\cite{yousefi2024laying}. Another recent work solves navigation planning problems in marine environments, which have a high degree of uncertainty~\cite{lin2020bounded}. In this domain, navigation actions have uncertain costs due to the uncertainty in the flow forecast. Thus, the cost of moving from one location to another is estimated and can be different from the actual cost. This approach aims at finding the plan with the highest probability of having the total cost lower than a user-defined upper bound. Unlike our work, this approach does not consider risk when solving the problem; thus, it cannot find plans with the highest expected utility.

Computing plans that meet certain probability thresholds with respect to the consumption of critically limited resources has been the focus of the work presented in~\cite{biundo2004dealing}. The uncertainty is modelled for continuous resources as a continuous probability distribution associated with operators to indicate the resource consumption of a specific operator. This information is then propagated to the higher levels of the task hierarchy to help make informed choices. Since the probability distribution of operator costs is not totally known, the model of operator costs corresponds to uncertainty-inducing actions. This approach aims to compute heuristics to guide the planning process to compute plans that do not exceed a certain threshold of resource consumption. 

Some works adopt hybrid planning approaches in the sense that HTNs are used to guide solving non-hierarchical planning problems, or HTN planning is combined with other planning techniques to solve planning problems in uncertain domains. In~\cite{tang2011planning}, probabilistic HTNs are used for planning in Markov Decision Processes (MDP) environments such that uncertainty is represented by probabilities that encode the subjective knowledge of users on how likely a specific method is chosen to decompose a compound task. In MDP environments, utilities are assigned to states in the form of rewards and the objective is to find the plan with the highest expected utility. Another approach uses HTNs to restrict its search for plans using fully observable non-deterministic non-hierarchical planning techniques~\cite{kuter2009task}. Actions in HTNs are extended to model multiple non-deterministic effects. Similar to this approach are the works presented in~\cite{kuter2005hierarchical} and~\cite{kuter2004forward}, which consider non-deterministic effects of actions without defined probabilities but use HTNs to guide the search of the underlying non-hierarchical planning techniques. All presented approaches for hybrid planning do not consider uncertainty or risk with respect to action costs, nor do they consider agent's attitudes. In addition, the last three works do not consider the quality of a solution, but simply the identification of a solution, if available.

\subsection{Utilities in HTN Planning and HTN-Like Planning}

Actions with hard-coded utilities that are not related to action costs but are based on the type of actions and their outcome are considered in~\cite{magnenat2012integration}. The approach works in robotics domains, where the action outcome represents a binary value of success or failure, and the utility of actions with a failure outcome is 0. The approach aims to find the plan with the highest expected utility, where the expected utility is computed as a product of the probability of action success, which can be learned online, and the utility of this action. Our work not only supports the utilities and probabilities as defined in this work but also abstracts away from a specific domain and provides an HTN planning framework that incorporates risk and risk attitudes to solve problems in uncertain and risky domains. 

A utility function is used to evaluate the execution cost or effect of primitive tasks in~\cite{luo2011messy}. This approach produces only the least-cost plan and does not deal with risk and risk attitudes in HTN planning. HTN planning was extended to account for stochastic actions that have a probability distribution over multiple action effects~\cite{meneguzzi2018goco}. The approach aims at maximising the expected utility, which is computed based on the probability distribution of each action in the plan and the utility. The utility in this work is a reward function that describes the reward obtained from transitioning from one state to another. Similarly, this work does not consider risks and risk attitudes.

An emotional-based planner incorporates expected utilities in a planning formalism similar to HTN planning~\cite{macedo2004emotional}. Uncertainty is defined and modelled using conditional and probabilistic effects of actions. Since the probability distribution is given, this definition of uncertainty corresponds to our definition of risk. The planner's aim is to compute plans with the highest expected utility, where expected utilities are computed for actions and measured in terms of the intensity of the emotions, drives, and other motivations it may elicit. The expected utilities are propagated in a preprocessing step to higher hierarchical levels to help make informed choices during planning. Our work is more general and can be used to include the modelling presented in this approach. In fact, our work is not only more general but also builds upon the standard HTN planning formalism, whereas the emotional-based planner works on concepts different from HTN planning ones (cf. cases of plans or abstract plans instead of methods). Moreover, the emotional-based work does not discuss the involvement of risks and risk attitudes.

Similar to the previous work is the decision-theoretic planner, where a different way to abstract conditional probabilistic actions. The system is called DRIPS~\cite{haddawy1994abstracting}. The aim is to compute plans with the highest expected utility, where risk and risk attitudes are not taken into account.

In summary, the examined approaches for HTN and HTN-like planning do not mention or treat risk as a separate concept, but some elements in their models could be applicable to the notion of risk as we treat in this paper.

\subsection{Informed Decision-Making in HTN Planning}

Our work can be seen as related to approaches that use additional information in HTN planning to compute quality plans. Such additional information takes the form of heuristics, preferences, and advice. There is also a branch of approaches that order HTN methods using a heuristic function that defines the distance between the goal state of the given planning problem and the goal states of methods~\cite{cheng2018improving,shivashankar2017incorporating,waisbrot2008combining}. The latter approaches usually aim at improving the planning performance regardless of the resulting plan quality and agent attitudes, and are, therefore, out of scope here.

\subsubsection{Heuristics}

Heuristics are used in HTN planning to guide the computation of optimal plans, mainly in deterministic domain models. An admissible heuristic to find optimal solutions in terms of plan cost in HTN planning was recently proposed in~\cite{bercher2017admissible}. Minimising the plan cost corresponds to the simplified objective of risk-neutral agents. Similarly, another approach uses a heuristic function to sort unexplored plans based on the sum of an estimated number of steps to reach the current partial plan and the estimated number of steps needed to reach the goal~\cite{bechon2014hipop}.

The SHOP2 planner has been enhanced with a limited branch-and-bound optimisation to guide the task decomposition to the cost-optimal plan with a possibly chosen execution time limit~\cite{nau2003shop2}. Another approach also uses a branch-and-bound algorithm with an admissible heuristic to guide the search to compute cost-optimal plans and a non-admissible heuristic to search faster~\cite{menif2017applying}. 

HTN planning is translated into classical planning to compute heuristics that are then used to guide searching for plans~\cite{holler2018generic,holler2020htn}. These heuristics estimate the number of decompositions and number of actions required to accomplish a given objective. The approach can be adapted to solve cost-optimal plans. In particular, if an admissible heuristic that enables the computation of optimal-cost plans is used in the relaxed classical planning model, then it can be used to compute cost-optimal plans in the HTN model. Another study translates HTN planning into propositional logic, where optimal plans are defined in terms of their length~\cite{behnke2019finding}.

Another approach requires users to annotate abstract tasks with lower and upper bounds on the costs of the possible plans that these tasks can be used for~\cite{marthi2007angelic,marthi2008angelic}. These values are used to speed up the search towards cost-optimal plans. Unlike this approach, we do not require any additional domain-specific information to be encoded. Also, this approach follows only the risk-neutral attitude. 

User ratings and social trust are employed as indicators for choosing preferable methods when composing Web services in~\cite{kuter2009semantic}. In this specific domain, user ratings and social trust can be seen as different attitudes toward some social phenomena. Our work abstracts away such specificities by proposing a general framework that allows incorporating attitudes and accounting for risks and/or uncertainty in HTN planning for any domain.

Although the presented approaches operate in deterministic domains, since they are based on averaging costs out to compute cost-optimal plans, they all can be adapted to domains with risk-inducing actions. However, the attitude that they follow corresponds only to a risk-neutral attitude.

\subsubsection{Preferences and Advice}
Since risk attitudes represent a common type of preference structure, it comes naturally to also look at works that incorporate any other form of preferences in HTN planning. A line of works focuses on the extension of PDDL3, which is a version of Plan Domain Description Language (PDDL) that supports temporally extended preferences and hard constraints~\cite{gerevini2005plan}, with HTNs to find preferred plans~\cite{sohrabi2008htn,sohrabi2008planning,sohrabi2009htn}. The plans are computed according to encoded preferences over the occurrence, decomposition, and instantiation of HTN tasks. The quality of the resulting plans is defined in terms of the number of preferences achieved. Theoretically, it might be possible to express risk attitudes using preferences. However, this requires preferences to be provided for each compound task and to be encoded as additional domain knowledge by a planning expert. Our work does not require any additional domain knowledge, nor does it need encodings for compound tasks. Instead, utility functions can be selected automatically based on the required risk attitude or manually by a domain expert. Moreover, encoding the planning choices to comply with a specific risk attitude in the form of preferences means that the domain expert should solve the planning problem to know which choices lead to the highest expected utility plan. In addition, existing approaches that use preferences in HTN planning do not deal with action costs in uncertain and/or risk-involving domains.

HTN planning has been extended with hard constraints to model advice on how to decompose methods~\cite{myers2000planning}. These constraints specify limited combinations of expert advice. As in preference-based HTN planning, these pieces of advice have to be encoded by domain experts as well as planning experts, and moreover, are not related to attitudes or action costs. 

Finally, additional information has been incorporated in the HTN domain knowledge to help make the choice of methods in an informed manner~\cite{amigoni2004planner}. This is achieved by assigning three values to each method for estimating the performance, cost and probability of success of the respective method. Again, these values depend on the expertise of the domain author and have no association with attitudes and costs of operators in uncertain and/or risky domains as defined in our work.

\subsection{Novelty}

Reviewing related work in AI planning literature shows that uncertainty and risk are not conceptually distinguished. In particular, planning approaches that model totally known probability distribution over action effects, action costs, and states are referred to as uncertainty approaches without considering the risk involvement.

Moreover, in most planning approaches, uncertainty forms are expressed in terms of partial observability of states and uncertainty about action effects or action costs. However, approaches that deal with the latter form of uncertainty usually aim at computing plans with the highest probability of not exceeding a predefined cost limit and do not find plans that comply with a specific attitude.

Existing approaches in HTN planning that use utility functions do not consider risk and agents' risk attitudes. Specifically, these approaches use hard-coded utility functions that are not related to action costs or a specific risk attitude. On the other hand, approaches that use utilities to evaluate action costs use planning formalisms that deviate from standard HTN planning and also do not consider the risk and risk attitudes of planning agents.

HTN planning approaches that provide some form of informed guidance of HTN planning towards quality plans are also relevant to our work. Some approaches do consider action costs but are developed for deterministic domain models and aim to find cost-optimal plans. Thus, they correspond to the risk-neutral attitude if adapted to domain models with variable costs. Other approaches use preferences and could be adapted to express attitudes. However, a domain expert must encode the planning choices of risk-sensitive agents.

Thus, we conclude that while there exists a few works that incorporate risk and risk attitudes in non-hierarchical planning, there does not exist an HTN planning approach that models risks involved in action costs and also incorporates utilities in hierarchical constructs to express and guide the planning process according to a particular risk attitude.

\section{Discussion}
\label{sec:discussion}

The present work is an initial step towards the study of risk for HTN and AI planning in general. While we provide a general framework, we also identify a number of research questions that sprout from the present investigation.

\textbullet\ {\em What kind of challenges might arise when modelling risk-aware HTN planning agents with an arbitrary utility function?}

The approach presented for risk-aware plan-based HTN planning assumes that the agent has a utility function that belongs to the family of utility functions of Equation~\ref{equ:exponentialUtilityFunction}. What makes this family of utility functions suitable is that it allows for maximising the expected utility of different plan segments separately. In particular, in the risk-aware plan-based HTN planning approach, the value that the A* uses to guide the search is based on summing up the expected utilities of the operators in the plan and the estimated expected utility of the compound tasks to be decomposed. This means that the algorithm is maximising the expected utility of segments of the plan and then combines these segments into one plan.

The problem becomes more complicated if the agent can have an arbitrary utility function that does not allow segmentation, e.g., $u(c) = (-c)^3$. The reason behind this is that, in general, $u(c_1 \times c_2) \neq u(c_1) \times u(c_2)$. In the worst case, it might be necessary to enumerate all the trajectories of all possible plans to find the one with the highest expected utility. Since this approach is computationally infeasible to solve such planning problems while giving flexibility in the choice of utility functions, other planning approaches should be developed.

\textbullet\ {\em What challenges are imposed by modelling risk-aware HTN planning agents with a dynamic risk attitude?}

When the risk attitude of the agent is static, it is possible to express a whole spectrum of risk attitudes just by changing the curving coefficient in Equation~\ref{equ:exponentialUtilityFunction}. Anyhow, for dynamic risk attitudes, where the risk attitude can change based on, for example, resources or planning history, there might be a need to use utility functions that do not allow segmentation. Consider, for example, the one-switch utility function defined in Equation~\ref{equ:one-switch}. This utility function can be used to evaluate the operator outcome with respect to the possible remaining resources after executing the operator. However, computing the expected utilities of operators and propagating them to higher hierarchical levels is not correct anymore since the utility function does not allow maximising each plan's segment individually. In this case, it might be necessary to enumerate all possible plan trajectories and compute the utility function for each of them. However, as in the case of static risk attitude, doing an exhaustive search is, in general, infeasible. Moreover, the remaining amount of resources $R$ is uncertain during planning since operators have variable costs.

Another challenge lies in the necessity for the agent to not only make planning choices that maximise the expected utility but also to account for the possibly limited amount of resources. In fact, limited resources may impede the agent from reaching its maximum expected utility if plans are computed assuming unlimited resources.

\textbullet\ {\em How can the general framework be specialised?}

By modelling different degrees of uncertainty and risk along the wide spectrum of uncertainty. For example, it can be specialised to model uncertainty-inducing actions. In this case, encoding the planning agent's beliefs about the probabilities of the outcomes is an interesting yet challenging research direction that would address the complexity of real-world domains further. Learning the probabilities of outcomes could be a possible direction.

\textbullet\ {\em How can existing HTN modelling languages be adapted to model risk-aware HTN planning problems?}

In general, our formalism happens to be in line with the formalism of probabilistic planning frameworks from classical planning, e.g., PPDDL~\cite{younes2004ppddl1}. However, our formalism incorporates variable effects and variable costs in HTN planning formalism. Additionally, it focuses particularly on the variability of costs and allows the modelling of single-effect operators with variable probabilistic costs. Modelling a probability distribution of action costs can be done in different ways depending on the type of the cost function. If the cost function is constant $c^{ic}$ or external $c^{ie}$, we can model it as a new construct of the action definition in the domain model. For example, the \verb|activate_esp| action in the AVs domain has a constant cost that can be modelled directly in the domain model. If the cost function is state-dependent $c^{is}$ or hybrid $c^{ies}$, it can be modelled as a new construct of the action definition in the domain model, but the function will contain variables related to the state, and that can be bound to values that determine the exact probability distribution of costs during planning or when grounding the planning problem in a preprocessing step. Another possibility is to model it as part of the initial state by adding predicates that indicate the action name and the bound parameters with the probability distribution. For example, the \verb|accelerate| action in the AVs domain has a hybrid cost function that depends on the road the vehicle is accelerating on and the traffic. The cost function, in this case, can be modelled as part of the \verb|accelerate| action in the domain model with two variables for the start and end locations. When these variables are bound to specific values during planning, the probability distribution can be determined based on the road being traversed and the traffic on it. Another approach involves modelling the probability distribution of costs within the initial state. This can be achieved by introducing predicates named after the \verb|accelerate| action, incorporating variables bound to specific values (i.e., start and end locations), and capturing the probability distribution of costs corresponding to the specific road and its associated traffic conditions.

\textbullet\ {\em Can the proposed approach be applied to AI planning techniques other than HTN planning?}

The concepts of uncertainty, risk, risk attitudes, and utilities are general concepts applicable across various AI planning techniques. From a modelling perspective, our general framework for risk-aware HTN planning problems can be easily adapted to other planning techniques. In particular, we model risk and uncertainty as a probability distribution of action costs. Given that actions are fundamental constructs in planning problems across all planning techniques, the modelling of risk and uncertainty follows the same approach. Some existing AI planning techniques already model such constructs. For example, in probabilistic planning, an extension to classical planning, actions have a probability distribution over possible effects~\cite{kushmerick1995algorithm}.

When it comes to adapting existing AI planning approaches to generate plans with the highest expected utility, the method varies based on the specific AI planning technique being utilised. Certain existing approaches compute plans with the highest expected utility by converting planning problems involving risk into problems that can be addressed by established methods, e.g.,~\cite{koenig1994risk}. Given that our approach begins by calculating the expected utility of operators, which are fundamental elements in all planning problems, the process of adapting existing approaches can start from this step. The subsequent steps involved in maximising the expected utility rely on the specificities of each individual planning technique.

We opt for HTN planning in our approach as it is the most suitable AI planning technique for automating planning in domains involving risk and uncertainty, and this choice is motivated by several reasons. Mainly, the rich domain knowledge and its intrinsic hierarchical structure enable HTN planning to provide a natural approach to simulate the way in which one conceptualises decision-making by starting with big tasks and breaking them into smaller, more specific ones in various possible ways. This allows planning agents to express their risk attitudes on different abstraction levels and constructs. Specifically, by propagating the expected utilities calculated for operators to higher hierarchical levels, it becomes possible to express risk attitudes regarding the selection of methods for decomposing complex tasks, the ordering of these tasks, and the assignment of variables to specific values from the problem instance. Such expressiveness is absent in classical planning methodologies, primarily due to the lack of multiple abstraction levels. The expressiveness inherent in HTN planning, coupled with its intrinsic hierarchical structure, facilitates the explanation of planning decisions made by the agent, for example, a vehicle's decisions explained to the driver in a manner that aligns closely with the driver's own reasoning~\cite{alnazer2022role}. This closeness in representation, planning methodology, and behaviour becomes particularly significant when the vehicle adopts the driver's risk attitude in decision-making. In such instances, it becomes easier to communicate to the driver the rationale behind the vehicle's specific choices, owning to the similarity between the knowledge representation, planning approach, and planning behaviours of HTN planning and the natural human way of planning. In addition to that, HTN planning is known for its efficiency and capabilities to cope with the requirements of real-world planning applications~\cite{georgievski2015htn}.

\textbullet\ {\em How can the approach be adapted to consider the dynamics of risk attitudes during the execution of the plans?}

Considering the dynamics of risk attitudes during plan execution requires adapting the planning approach to compute plans that give the executing agents the possibility to follow a risk attitude different from the one that they planned with.

Our approach is applicable without modifications if the planning agent follows a static risk attitude when computing the plan and the executing agent follows the same risk attitude. If the executing agent follows a risk attitude different from the planning agent, then a possible solution is to replan according to the risk attitude of the executing agent.

If the executing agent has a dynamic risk attitude that can change due to some factors, e.g., remaining resources, a possible solution is to do contingent planning and compute conditional plans during the plan generation phase. The plans should allow the executing agent to choose specific actions that maximise its expected utility based on the changing factors and states during the execution. When considering variable-cost operators with single effects, the conditions that determine the operators to include in the plan are related to the uncertainty sources and the factors that influence the dynamics of the risk attitude. In the case of variable-cost operators with multiple possible effects, in addition to the previous factors, the conditions are also influenced by the observation of the states. In any case, an approach that combines contingent planning with HTN planning is required. Examples of such approaches include~\cite{zhao2023probabilistic,bouguerra2004hierarchical,bouguerra2005pc}. While these approaches consider contingent HTN planning, they aim at computing feasible plans regardless of their quality or plans with the lowest expected cost. Thus, these approaches should be expanded to consider the risk attitudes and their dynamics and to allow agents to execute plans that have maximum expected utility. Another possible solution for executing agents with dynamic risk attitudes is that each time the executing agent changes its risk attitude, we replan to compute a plan with the highest expected utility following the new risk attitude of the agent.

\textbullet\ {\em How can the proposed approach be empirically compared to existing approaches in AI planning?}

Upon reviewing the literature on HTN planning, we can observe that the approaches closest to ours use some form of utility function. However, existing approaches that belong to this category typically utilise hard-coded utility functions that are unrelated to action costs or a specific risk attitude (e.g.,~\cite{magnenat2012integration}). On the other hand, approaches that use utilities to evaluate action costs diverge from standard HTN planning and do not consider risk and the risk attitudes of planning agents (e.g.,~\cite{macedo2004emotional}). HTN planning approaches that provide some informed guidance for HTN planning towards quality plans are also relevant to our work. Some approaches do consider action costs but are developed for deterministic domain models and aim at finding cost-optimal plans, making them degenerate cases of our approach (e.g.,~\cite{bercher2017admissible}). If these approaches are adapted to domain models with variable costs, they correspond to the risk-neutral attitude. Other approaches use preferences and could potentially be adapted to express attitudes (e.g.,~\cite{sohrabi2008htn,sohrabi2008planning,sohrabi2009htn}), but they require a domain expert to encode the planning choices of risk-sensitive agents. Thus, while some works address risk and risk attitudes in non-hierarchical planning, no HTN planning approach currently exists that models the risks involved in action costs and integrates utilities within hierarchical constructs to guide the planning process according to a specific risk attitude. This means an empirical comparison between our approach and the existing approaches is infeasible due to the differences between our approach and existing ones in the AI planning techniques, algorithms, goals, and/or expressiveness used, including the modelling of the planning problems.

\section{Conclusions}
\label{sec:conclusion}

In decision theory, risk and uncertainty are usually defined as distinct concepts. We bring such a distinction to the field of planning by considering two types of actions: risk-inducing and uncertainty-inducing actions. We identify the possible sources of uncertainty that have effects on the action outcomes ranging from totally known to totally unknown probability distribution of effects and their corresponding costs. Using this distinction, we developed a general framework for HTN planning that can be specialised by future studies to deal with planning problems that involve risk and uncertainty related to action costs. Since our work is rooted in decision theory, a well-established field of research, we believe that it helps bring HTN planning a step closer to simulating the way in which decision makers address and solve problems while embracing risk.

We studied a specific realisation of this framework for which actions have single effects and are risk-inducing, and planning agents adopt a specific risk attitude, which can be static or dynamic, for making planning choices. We further developed an approach that can solve risk-aware plan-based HTN planning problems by incorporating expected utilities with hierarchical constructs to allow finding risk-aware plans, i.e., plans with the highest expected utility that comply with the planning agent's risk attitude. In our developed approach, we showed that by restricting the type of utility function to the family of linear and exponential utility functions, we can adapt approaches that find cost-optimal plans to solve risk-aware HTN planning problems. Based on this, we went one step further and discussed how to adapt an existing approach for cost-optimal plans to solve state-based HTN planning problems.

Most existing approaches in AI planning do not distinguish between risk and uncertainty in action outcomes, but rather they use uncertainty as an umbrella term even when having risk-inducing actions. Moreover, while some existing approaches incorporate utilities in HTN planning, risk awareness has not been considered in previous work, thus limiting the ability of the user to express its goals and generate plans that align with its risk attitude. We also showed that, in general, planning approaches that guide HTN planning by making informed choices via heuristics or preferences assume a deterministic model of the domain and adopt a risk-neutral attitude of planning agents.

\appendix
\section{Real-World Domains with Uncertainty}
\label{apx:case-studies}

We look into two case studies derived from real-world domains, namely smart homes and marine environments. These domains are selected because they exhibit elements of uncertainty and require making choices that involve risks. For each domain, we illustrate the relevant sources of uncertainty and how each uncertainty source affects action costs.

\subsection{Smart Homes}\label{apx:smart-homes}

Smart homes are home structures aware of their state and can change such state proactively for the safety, comfort, and needs of its residents~\cite{kaldeli2013coordinating}.
Consider now a family living in a home that is equipped with a system that can coordinate all devices and appliances in the home and cooperate with a domestic robot similar to the humanoid robot presented in~\cite{gravot2006cooking}. The robot has multiple sensors and can move around the home, pick up and place things, cut and clean for cooking, and clean the floors. Let us divide the discussion between the case of the resident and the robot performing the actions. 

Imagine a scenario in which the robot is helping a person cook a meal. Here, we have two types of executing agents; a human and a system (the robot). Let us assume that the robot is mixing the dressing for the meal and the person is chopping some vegetables. The speed at which each of these actions is done is uncertain due to the capabilities of the person and the robot, which are internal regular sources of uncertainty. The probability of the time needed to perform these actions could be measured based on past experience, or it could be partially known during planning. Say that while the person is chopping the vegetables, he gets a knife cut. This leads to a failure in the chopping action and is considered a random internal source of uncertainty since it is an unexpected random event that can lead to consequences hard to quantify in advance during planning.

Consider now a scenario where the robot is cleaning the floor while the inhabitants are moving around, disturbing the robot's movement. The movement of home inhabitants is an external regular source of uncertainty that can incur different costs, e.g., time and cleanness. Similarly, the robot movement can be considered an external regular source of uncertainty that affects the variability of costs of the inhabitant movement actions. Also, leaving the doors closed by inhabitants is another source of external uncertainty that can affect the action costs of the robot.

\subsection{Marine Environments}\label{apx:marine}

The exploration of the sea and the collection of samples constitutes what we refer to as ``marine environment''. For this, we consider two types of agents: scientific divers and underwater gliders, both equipped with a wide variety of sensors to collect ocean environmental data, photos, and videos. During the solo dives of scientific divers, the gliders can be used as drop-in diving buddies to increase the safety of divers by providing them support, such as giving their exact position or carrying an extra air tank~\cite{nadj2020using}.
 
Human behaviour always constitutes a big part of uncertainty in most environments due to its variability and unpredictability. This is especially true in extreme environments, such as oceans, that impose high physiological and psychological pressure on people, that is, divers. Intelligence and personality differ between individuals and can have effects on the behaviour underwater~\cite{colodro2015individual}. The variability of divers behaviour represents an internal regular source of uncertainty in marine environments, leading to variability in the costs of actions that are performed. For example, the time needed to take a picture of specific phenomena underwater might differ from one diver to another due to the differences in their behaviours. When the agents are gliders, i.e., the gliders are responsible for collecting the data, there can be multiple internal regular sources of uncertainty. For example, a malfunction of one of the navigation sensors or the loss of a wing can lead to spending more than expected time and energy to move to the location of interest, or, in the worst case, can lead to complete failure in reaching the location.

A huge part of uncertainty during planning comes from the uncertainty in the flow forecast~\cite{lin2020bounded}. This is due to the fact that these systems, like the atmosphere, are chaotic, non-linear systems. Thus, flow forecasts in marine environments are considered external regular sources of uncertainty, leading to cost variability of marine actions, such as navigation and data sampling, irrespective of whether they are performed by divers or gliders. Another external regular source of uncertainty is due to the quality of the interconnection between the diver and the surface. Since gliders need to communicate with the surface to relocate, the variability of the communication quality can lead to variability in the cost of positioning actions. On the other hand, there are also many random external sources of uncertainty. For example, an unexpected bio-fouling issue has been reported in the Gulf of Mexico, where Remora fish---which typically attach to sharks and other large marine animals---held the gliders down until they decided to detach~\cite{jones2007slocum}. Another random external source of uncertainty exists in scenarios where a glider serves as a diving buddy. In this case, if the glider accidentally hits the diver, this leads to unexpected consequences and action costs.

\section{Proof of Theorem~\ref{theo:planExpectedUtility-segmentation}}
\label{appendix:proof-eu-segmentation}
Here, we prove that the segmentation is possible according to Equation~\ref{equ:planExpectedUtility-segmentation}.
\begin{proof}
To show how the segmentation is possible according to Equation~\ref{equ:planExpectedUtility-segmentation}, we expand the left-hand side (LHS) and right-hand side (RHS) of the equation and prove that they are equal. In particular, the LHS of Equation~\ref{equ:planExpectedUtility-segmentation} can be written as follows:\\

\begin{equation}
 \label{equ:planExpectedUtility-segmentation-proof1}
    \begin{split}
 EU(\pi) & = \sum_{i=1}^{k} \sum_{j=1}^{l} \ldots \sum_{m=1}^{n} [p_{1i} p_{2j} \ldots p_{sm} \dfrac{ae^{\alpha (c_{1i}(o_1) + c_{2j}(o_2) + \ldots + c_{sm}(o_s))}}{\alpha}]\\
 & = p_{11} p_{21} \ldots p_{s1} \dfrac{ae^{a \alpha (c_{11}+ c_{21} + \ldots + c_{s1})}}{\alpha} + \ldots + p_{11} p_{22} \ldots p_{s1} \dfrac{ae^{a \alpha (c_{11}+ c_{22} + \ldots + c_{s1})}}{\alpha} + \\
 & \ \  \ldots + p_{1k} p_{2l} \ldots p_{sm} \dfrac{ae^{a \alpha (c_{1k}+ c_{2l} + \ldots + c_{sm})}}{\alpha}
\end{split}
\end{equation}

And the right-hand side (RHS) of Equation~\ref{equ:planExpectedUtility-segmentation} can be written as follows:\\

\begin{equation}
 \label{equ:planExpectedUtility-segmentation-proof2}
    \begin{split}
 EU(\pi) & = \dfrac{\alpha^{s-1}}{a^{s-1}} EU(o_1) \times EU(o_2) \times \ldots EU(o_s)\\
 & = \dfrac{ \alpha^{s-1}}{a^{s-1}} \sum_{i=1}^{k} [p_{1i} \dfrac{a e^{\alpha c_{1i}(o_1)}}{\alpha}] \times  \sum_{j=1}^{l}  [p_{2j} \dfrac{a e^{\alpha c_{2j}(o_2)}}{\alpha}] \times \ldots \times \sum_{m=1}^{n}  [p_{sm} \dfrac{a e^{\alpha c_{sm}(o_s)}}{\alpha}] \\
 & = \dfrac{ \alpha^{s-1}}{a^{s-1}} ((p_{11} \dfrac{a e^{\alpha c_{11}(o_1)}}{\alpha} + p_{12} \dfrac{a e^{\alpha c_{12}(o_1)}}{\alpha} + \ldots + p_{1k} \dfrac{a e^{\alpha c_{1k}(o_1)}}{\alpha} ) \times \\
 & \ \ \ ((p_{21} \dfrac{a e^{\alpha c_{21}(o_2)}}{\alpha} + p_{22} \dfrac{a e^{\alpha c_{22}(o_2)}}{\alpha} + \ldots + p_{2l} \dfrac{a e^{\alpha c_{2l}(o_2)}}{\alpha}) \times \ldots \times \\
 & \ \ \ (p_{s1} \dfrac{a e^{\alpha c_{s1}(o_s)}}{\alpha} + p_{s2} \dfrac{a e^{\alpha c_{s2}(o_s)}}{\alpha} + \ldots + p_{sn} \dfrac{a e^{\alpha c_{sn}(o_s)}}{\alpha}) )\\
 & = \dfrac{ \alpha^{s-1}}{a^{s-1}} (p_{11} p_{21} \ldots p_{s1} \dfrac{a^se^{a \alpha (c_{11}+ c_{21} + \ldots + c_{s1})}}{\alpha^s} + \ldots + \\
  & \ \ \  p_{11} p_{22} \ldots p_{s1} \dfrac{a^s e^{a \alpha (c_{11}+ c_{22} + \ldots + c_{s1})}}{\alpha^s} + \ldots + 
    p_{1k} p_{2l} \ldots p_{sm} \dfrac{a^se^{a \alpha (c_{1k}+ c_{2l} + \ldots + c_{sm})}}{\alpha^s}) \\
  & = p_{11} p_{21} \ldots p_{s1} \dfrac{ae^{a \alpha (c_{11}+ c_{21} + \ldots + c_{s1})}}{\alpha} + \ldots + \\
  & \ \ \  p_{11} p_{22} \ldots p_{s1} \dfrac{ae^{a \alpha (c_{11}+ c_{22} + \ldots + c_{s1})}}{\alpha} + \ldots +  p_{1k} p_{2l} \ldots p_{sm} \dfrac{ae^{a \alpha (c_{1k}+ c_{2l} + \ldots + c_{sm})}}{\alpha}
\end{split}
\end{equation}

We can see from Equations~\ref{equ:planExpectedUtility-segmentation-proof1} and~\ref{equ:planExpectedUtility-segmentation-proof2} that Equation~\ref{equ:planExpectedUtility-segmentation} holds. In this case, to compute the plan's expected utility, the expected utility of each plan's operator or set of operators can be maximised separately, and the result can be combined by multiplying the individual maximised expected utilities. 
    
\end{proof}

\section{Formal Properties of Algorithms~\ref{alg:computeUtilities} and~\ref{alg:computeHighestExpectedUtilityPlan}}
\label{app:formal-properties}

Algorithm~\ref{alg:computeHighestExpectedUtilityPlan} is correct (sound and complete) and provides optimal solutions. Next, we prove this statement formally.

\begin{theorem} [Termination of preprocessing phase]
Algorithm~\ref{alg:computeUtilities} terminates.
\end{theorem}

\begin{proof}
    The termination of Algorithm~\ref{alg:computeUtilities} follows directly from the fact that the GCV-TDG is a finite graph and the preprocessing algorithm tracks the decomposition paths as per Definition~\ref{def:decomposition_path-preprocessing}, and detects and handles cycles as per Definition~\ref{def:cycle-preprocessing}.
\end{proof}

Since our objective is to maximise the expected utility, our work is essentially a maximisation heuristic~\cite{stern2014max}. It is, therefore, crucial to prove the admissibility of the heuristic within this maximisation context to ensure the effectiveness and correctness of the approach. To that end, we have modified the standard definition of admissibility in MAX problems~\cite{stern2014max,cohen2020solving}---traditionally used to estimate the remaining cost to a goal---to now consider the estimation of the expected utility of the remaining compound tasks that must be decomposed to generate a plan.

\begin{definition}[Admissibility in risk-aware HTN planning problems]
\label{def:admissibility}
    A function $h$ is said to be admissible for risk-aware HTN planning problems if and only if for every partial plan in the search space, it holds that $h(tn)$ is an upper bound (i.e., is larger or equal) on the expected utility of the compound tasks that are part of the partial plan and need to be decomposed to reach the solution. 
\end{definition}
Thus, $h(tn)$ leads to an upper bound estimation of the expected utility of the optimal (i.e., the highest expected utility) solution. Since the heuristic considers the maximum expected utility as an estimation for each compound task and considers the maximum expected utility among all compatible groundings, according to the definition of admissibility of maximisation problems (see Definition~\ref{def:admissibility}), the heuristic is admissible and can be used to guide the search to compute the plans with the highest expected utility.

\begin{theorem} [Admissibility]
The heuristic of compound tasks ($EU_T(v)$) is admissible.
\end{theorem}

\begin{proof}
    First, we have to prove that the heuristic never leads to an underestimation of the expected utility of the optimal plan. Given an optimal plan $\pi_{optimal}$ with $s$ operators, its expected utility is defined according to Equation~\ref{equ:planExpectedUtility-segmentation} as follows:\\
    \begin{align*}
        EU(\pi_{optimal}) &=  \dfrac{\alpha^{s-1}}{a^{s-1}} EU(o_1) \times EU(o_2) \times \ldots EU(o_s))\\
 & = \dfrac{a}{\alpha} \cdot \dfrac{\alpha^s}{a^s} EU(o_1) \times EU(o_2) \times \ldots EU(o_s)\\
 & = \dfrac{a}{\alpha} \cdot \dfrac{\alpha}{a} \cdot EU(o_1) \times \dfrac{\alpha}{a} \cdot EU(o_2) \times \ldots \times \dfrac{\alpha}{a} \cdot EU(o_i) \\
 &\; \; \; \;  \underbrace{ \dfrac{\alpha}{a} \cdot EU(o_{i+1}) \times \ldots \times \dfrac{\alpha}{a} \cdot EU(o_s) }_{\leq EU(t') }
    \end{align*}
    Considering that the optimal plan results from the decomposition of a partial plan $tn'$ by a method $m_{opt} = \langle ct(m_{opt}), pre(m_{opt}), tn(m_{opt}) \rangle$. We assume that $tn'$  has $i$ operators and one compound task $t'$, resulting in $s-i+2$ operators when decomposed into the optimal plan. That is, $\exists tn'=\langle T', \varphi', \psi' \rangle$, such that 
    \begin{itemize}
        \item $\forall o_k \in \pi_{optimal},$ where $k \in [1,i]:$
        \begin{itemize}
            \item $o_k \in T' \land$
            \item $ct(m_{opt}) = t' \land$
            \item $tn' {\xrightarrow[t',m]{}}_D  \pi_{optimal}$
        \end{itemize}
        \item $\forall o_l \in \pi_{optimal},$ where $l \in [i+1,s]:$
        \begin{itemize}
            \item $o_l \in tn(m_{opt})$
        \end{itemize}
    \end{itemize}    
    $\forall m''$ such that $ct(m'') = t'$, $EU(t') = max(EU(m''))$ according to Equation~\ref{equ:plan_based_heuristic-compound_task}. We denote the method with the highest expected utility $m_{max}$. So, $\forall m_{alt} \in M_{alt}$, where $m_{alt} \neq m_{max} \land t' = ct(m_{alt}) \land M_{alt}$ is the set of all methods that can decompose $t'$ except for $m_{max}$, it holds that $EU(m_{alt}) \leq EU(m_{max})$. Since either $m_{opt} = m_{max}$ or $m_{opt} \in M_{alt}$, it holds:
    \begin{align*}
      EU(m_{opt}) &\leq  EU(m_{max})\\
    \Longrightarrow  EU(m_{opt}) &\leq  EU(t')\\
   \Longrightarrow \dfrac{\alpha}{a} \cdot EU(o_{i+1}) \times \ldots \times \dfrac{\alpha}{a} \cdot EU(o_s) &\leq \dfrac{\alpha}{a} \cdot EU(o_n) \times \ldots \times \dfrac{\alpha}{a} \cdot EU(o_m) \\
    \end{align*}
    where $o_n \ldots o_m \in tn(m_{max})$.
    Since the logarithm is monotonic,
     \begin{align*}
    log(\dfrac{\alpha}{a} \cdot EU(o_{i+1}) \times \ldots \times \dfrac{\alpha}{a} \cdot EU(o_s)) &\leq log(\dfrac{\alpha}{a} \cdot EU(o_n) \times \ldots \times \dfrac{\alpha}{a} \cdot EU(o_m))\\
     \dfrac{1}{a} log( \alpha \cdot EU(o_{i+1})) \times \ldots \times \dfrac{1}{a} log( \alpha \cdot EU(o_s)) &\leq \dfrac{1}{a} log( \alpha \cdot EU(o_n) \times \ldots \times \dfrac{1}{a} log( \alpha \cdot EU(o_m)).
    \end{align*}
    This proves that, in the last decomposition that leads to reaching the optimal plan by decomposing task $t'$, the heuristic of $t'$ overestimates the actual expected utility. By backward induction, we can see that since for each decomposition, we always take the maximum expected utility, we are always overestimating the actual expected utility of the optimal plan. That is, according to the admissibility definition in risk-aware HTN planning problems, the expected utility of the optimal plan is always less or equal to the expected utility of any partial plan that eventually decomposes into the optimal plan, where this expected utility equals the expected utility of the operators in this partial plan plus the heuristic value.
\end{proof}

\begin{theorem} [Soundness]\label{thrm:soundness}
Algorithm~\ref{alg:computeHighestExpectedUtilityPlan} is sound.
\end{theorem}

\begin{proof}
    (1) We first prove that the algorithm terminates. There are two ways to the algorithm's termination. First, when the fringe is empty, in which case the algorithm returns failure. Second, when the node popped from the fringe constitutes a solution (according to Definition~\ref{def:solution-plan_based}), the algorithm returns it. Since the algorithm is based on the A* algorithm, which is known to be sound and complete~\cite{farreny1999completeness}, the algorithm either finds a solution and returns it or runs out of nodes and returns failure.
    (2) Next, we prove that the returned node is a valid decomposition of the initial task network. This follows directly from the definition of decomposition (Definition~\ref{def:decomposition-plan_based}). In particular, the node corresponding to the initial task network is $tn_0 = \langle T_0, \varphi_0, \psi_0 \rangle$ and with each possible decomposition of a compound task $t \in T_0$ using a method $m$ (i.e., $tn {\xrightarrow[t,m]{}}_D  tn_s$), the new potential solution node added to the fringe is defined, according to Definition~\ref{def:decomposition-plan_based}, $tn_s = \langle T_s, \varphi_s, \psi_s  \rangle$ where the ordering relations are maintained. The process is repeated for each node resulting from the decomposition of the initial task network until reaching a primitive task network $tn_i$. Thus, by induction, the nodes added to the fringe, from which the solution is popped, if one exists, constitute valid decompositions of the initial task network (i.e., $tn_0 {\rightarrow}_D^*  tn_i$). Furthermore, when a node representing a primitive task network is popped from the fringe, the algorithm checks whether it is executable in the initial state before returning it as a solution. 
    (1) and (2) represent properties of the solution to any plan-based HTN planning problem. 
\end{proof}

\begin{theorem} [Completeness]
Algorithm~\ref{alg:computeHighestExpectedUtilityPlan} is complete.
\end{theorem}

\begin{proof}
    The completeness of Algorithm~\ref{alg:computeHighestExpectedUtilityPlan} follows directly from the proven completeness of the A* algorithm, which is used as a basis for our algorithm. The only exception we need to consider is when the planning problem requires solutions that run into cycles multiple times. In this case, the algorithm might not return a solution unless there is a defined bound for running into cycles that guarantees finding solutions within it.
\end{proof}

We prove that when a solution is returned, it is a solution to the risk-aware HTN planning problem, i.e., it is the optimal one. Since we use A* as a basis for our algorithm and the algorithm is guided by an admissible heuristic, the optimality of the returned solution follows directly.
 
\begin{theorem} [Optimality]\label{thrm:optimality}
Algorithm~\ref{alg:computeHighestExpectedUtilityPlan} is optimal.
\end{theorem}
\begin{proof}
   Ab absurdum, suppose that the returned solution $\pi$ is not optimal. Then, there is a solution $\pi'$ that has a higher expected utility than $\pi$. This happens if $\pi'$ is popped from the fringe before $\pi$. However, when adding a potential solution node to the fringe, the fringe is always sorted ascending with respect to the expected utility of the nodes. This means that the returned solution has an expected utility that is higher than any other node's expected utility. This is a contradiction; thus, $\pi$ is the optimal solution.
\end{proof}

\begin{theorem} [Worst-Case Time Complexity]\label{thrm:complexity}
Algorithm~\ref{alg:computeHighestExpectedUtilityPlan} has a worst-case time complexity of $O(d^{d_{max}} \cdot (f + d_{max} \cdot \log d))$, where $d$ is the maximum branching factor, $d_{max}$ is the maximum depth of decomposition, and $f$ is the average number of tasks in a decomposition.
\end{theorem}

\begin{proof}
To characterise the time complexity of the algorithm, we analyse it line by line, focusing on the main loop, where the critical computation occurs. Let $d_{max}$ be the maximum depth of decomposition, $d$ the maximum branching factor (i.e., the maximum number of decompositions (methods) applicable per task), and $f$ the average number of tasks in a decomposition.

Looping through all tasks $v_{\overline{t}} \in T_{n_0}$, doing the necessary checks, and the initialisation (Lines 1-7) takes $O(k) + O(1) \approx O(k)$ time, where $k=|T_{n_0}|$. Popping the first element from the fringe and checking whether a task is primitive (Lines 9-10) takes a constant time. Generating the decompositions (Line 12), updating the fringe (Line 20), and adding decompositions to the visited list (Line 22) takes $O(d)$ while sorting the fringe (Line 21) takes $O(n_i \log n_i) = O(d^i \cdot i \cdot log d)$ for $n_i = O(d^i)$ since, in the worst case, the algorithm explores all possible decompositions at each hierarchical level and the fringe grows exponentially in the branching factor (i.e., at depth $i$, the fringe contains $O(d^i)$ partial plans). In the main for loop (Lines 13-19), for each decomposition, we have to iterate through all its tasks, taking $O(f)$. For all decompositions, it takes $O(d \cdot f)$. Thus, the total cost per iteration of the main while loop (Lines 8-22) is $O(1) + O(d) + O(d \cdot f) + O(d^i \cdot i \cdot \log d) \approx O(d^i \cdot i \cdot \log d) + O(d \cdot f)$.

Given that the upper bound on the number of nodes (partial plans) in the search space is $O(d^{d_{max}})$, the complexity of the main loop then depends on iterating over all these nodes. Thus, the total complexity combining the upper bound of nodes becomes $O(k) + O(d^{d_{max}} \cdot (f + d_{max} \cdot \log d)) \approx O(d^{d_{max}} \cdot (f + d_{max} \cdot \log d))$.
\end{proof}

\begin{theorem} [Best-Case Time Complexity]\label{thrm:complexity-best}
Algorithm~\ref{alg:computeHighestExpectedUtilityPlan} has a best-case time complexity of $O(k) + O(d_{min} \cdot d (f + \log n))$, where $k=|T_{n_0}|$, $d_{min}$ be the minimum depth of decomposition, $d$ is the number of decompositions (methods) applicable per task, and $f$ is the average number of tasks per decomposition, assuming that the initial task network is not primitive and executable.
\end{theorem}

\begin{proof}
To determine the best-case time complexity of the algorithm, we assume that the heuristic guarantees no backtracking and that $tn_0$ is not primitive and executable. We need to evaluate the algorithm under optimal conditions where each step directly leads to the goal without exploring unnecessary nodes. The key assumptions are (1) The heuristic is perfect, guiding the search directly to the solution without backtracking, and (2) Each partial plan $tn_{current}$ leads to further decompositions that directly progress toward the goal. These two conditions are true when the methods with the highest expected utilities, computed in the preprocessing step, always lead to solutions, i.e., any node with the highest expected utility that is popped from the $fringe$ will eventually lead to a primitive and executable task network.

To characterise the time complexity of the algorithm, we analyse it line by line, focusing on the main loop, where the critical computation occurs. Let $d_{min}$ be the minimum depth of decomposition, $d$ the branching factor (i.e., the number of decompositions (methods) applicable per task), and $f$ the average number of tasks in a single decomposition.

Looping through all tasks $v_{\overline{t}} \in T_{n_0}$, doing the necessary checks, and the initialisation (Lines 1-7) takes $O(k) + O(1) \approx O(k)$ time, where $k$=$|T_{n_0}|$. In the main loop, popping the first element from the fringe and checking whether a task is primitive (Lines 9-10) takes a constant time. Generating the decompositions (Line 12), updating the fringe (Line 20), and adding decompositions to the visited list (Line 22) takes $O(d)$. Inserting $d$ decompositions to the fringe (Line 20) and sorting it (Line 21) takes $O(d \cdot \log n)$.  In the main for loop (Lines 13-19), for each decomposition, we have to iterate through all its tasks, taking $O(f)$. For all decompositions, it takes $O(d \cdot f)$. Thus, the total cost per iteration of the main while loop (Lines 8-22) is $O(1) + O(d) + O(d \cdot \log n) + O(d \cdot f) \approx O(d \cdot \log n) + O(d \cdot f) $.

Given that the heuristic is perfect, the algorithm explores exactly $d_{min}$ nodes (partial plans) in the search space. Thus, the complexity of the main loop depends on iterating over these nodes, leading to a total complexity of $O(k) + O(d_{min} \cdot d (f + \log n))$.

\end{proof}

To characterise the time complexity of the algorithm, we analyse it line by line, focusing on the main loop, where the critical computation occurs. Let $d_{min}$ be the depth of the optimal solution path in the decomposition tree, $f$ the average number of tasks in a decomposition, and $n$ the size of the fringe at any given point.

Looping through all tasks $v_{\overline{t}} \in T_{n_0}$ and doing the necessary checks takes and initialising $g(tn_0)$, $h(tn_0)$, and $f(tn_0)$ (lines 1-7) takes $O(k)$ time, where $k=|T_{n_0}|$ and a constant time, respectively. In the main loop, the computation depends on $d$ and $f$ (Lines 13-21). For each decomposition, we have to iterate through all its tasks, taking $O(f)$. For all decomposition, it takes $O(df)$. Updating and sorting the fringe (Line 22) takes $O(d)$ and $O(log\  n)$, respectively, where $n$ is the current size of the fringe. Finally, adding new decompositions to the visited set (Line 23) is $O(d)$.

Given that the upper bound on the number of nodes (partial plans) in the search space is $O(b^d)$, where $b$ is the branching factor (i.e., the average number of decompositions per task), the complexity of the main loop then depends on iterating over all these nodes. Thus, the total complexity combining the upper bound of nodes becomes $O(k) + O(b^ddf)$.

\section{Analysing Planning Problems in Risk-Involving Domains}
\label{app:planning-problems}
To better understand the functioning of the proposed risk-aware, plan-based HTN planning algorithm and to illustrate how agents with varying risk attitudes compute and select plans aligned with their preferences, we analyse two specific planning problems in the Autonomous Vehicles and Satellite domains~\cite{alnazer2025domains}. Using a brute-force approach, we compute all possible plans that solve these planning problems and demonstrate how agents choose the plans with the highest expected utility, reflecting their respective risk attitudes. 

\subsection{Autonomous Vehicles Problem Instance}
\label{app:avs-problem-instance}

\begin{figure} [H]
 	\centering
 	\includegraphics[width=0.75\columnwidth]{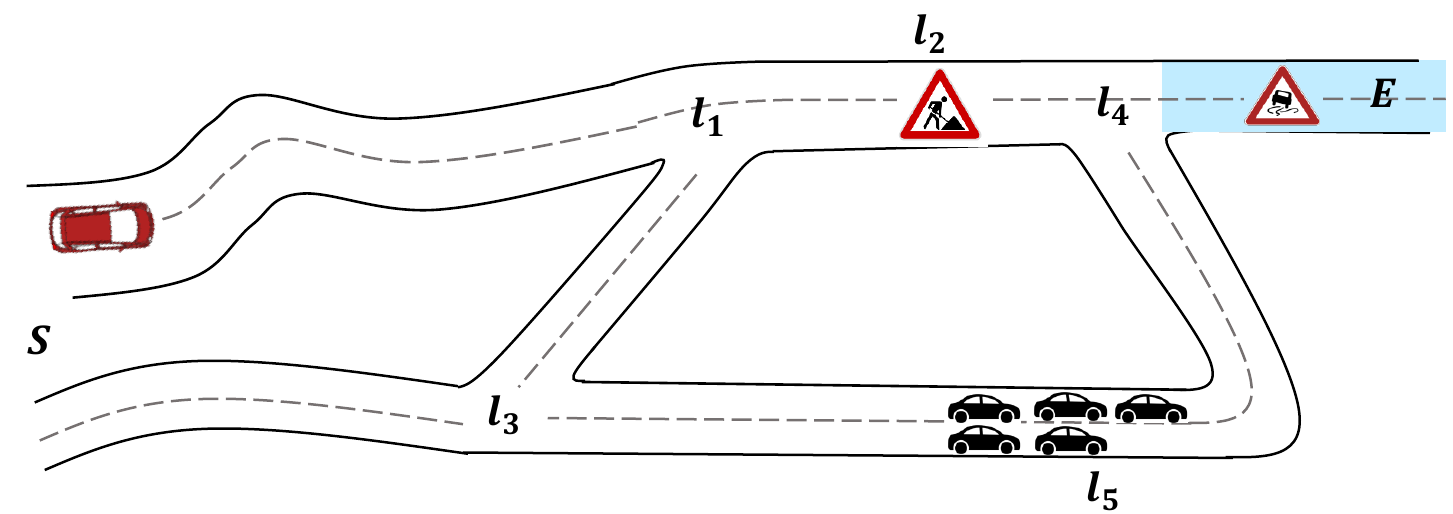}
 	\caption{A problem instance in the AVs domain with seven locations $S$, $l_1$, $l_2$, $l_3$, $l_4$, $l_5$, and $E$, and various road complexities.}
 	\label{fig:problem-instance}
\end{figure}

Consider the scenario illustrated in Figure~\ref{fig:problem-instance} that graphically shows a planning problem instance in the AVs domain. In this planning problem, there are seven different locations denoted as $S$, $l_1$, $l_2$, $l_3$, $l_4$, $l_5$, and $E$. The vehicle is initially at $S$ and should navigate to $E$ while handling the various road complexities. We assume that the vehicle has enough power to navigate all the roads, and it is nighttime, so the vehicle has to turn on the lights. The roads between $S$ and $l_1$ and between $S$ and $l_3$ are complexities-free. However, the first road is longer than the second. The road between $l_1$ and $l_4$ has constructions, at location $l_2$. The road between $l_3$ and $l_4$ has a school area that is very crowded with pedestrians and traffic jams, i.e., moving incidents, at location $l_5$. The road between $l_1$ and $l_3$ is a highway free from any complexities but has a variable amount of traffic jams. Finally, the road between $l_4$ and $E$ is icy and slippery. The probability distribution of action costs and the logic behind it is explained in~\cite{alnazer2025domains}. We denote this problem instance \textbf{P1}.

Having risk-inducing actions in this domain leads to making the planning agent, which is, in this case, the AI planning system in the autonomous vehicle, risk-aware, resulting in the computation of plans aligned with its risk attitude. In this case study, we aim to illustrate how agents with varying risk attitudes compute different plans that mirror their risk attitudes.

\begin{figure}
 	\centering
 	\includegraphics[width=0.7\columnwidth]{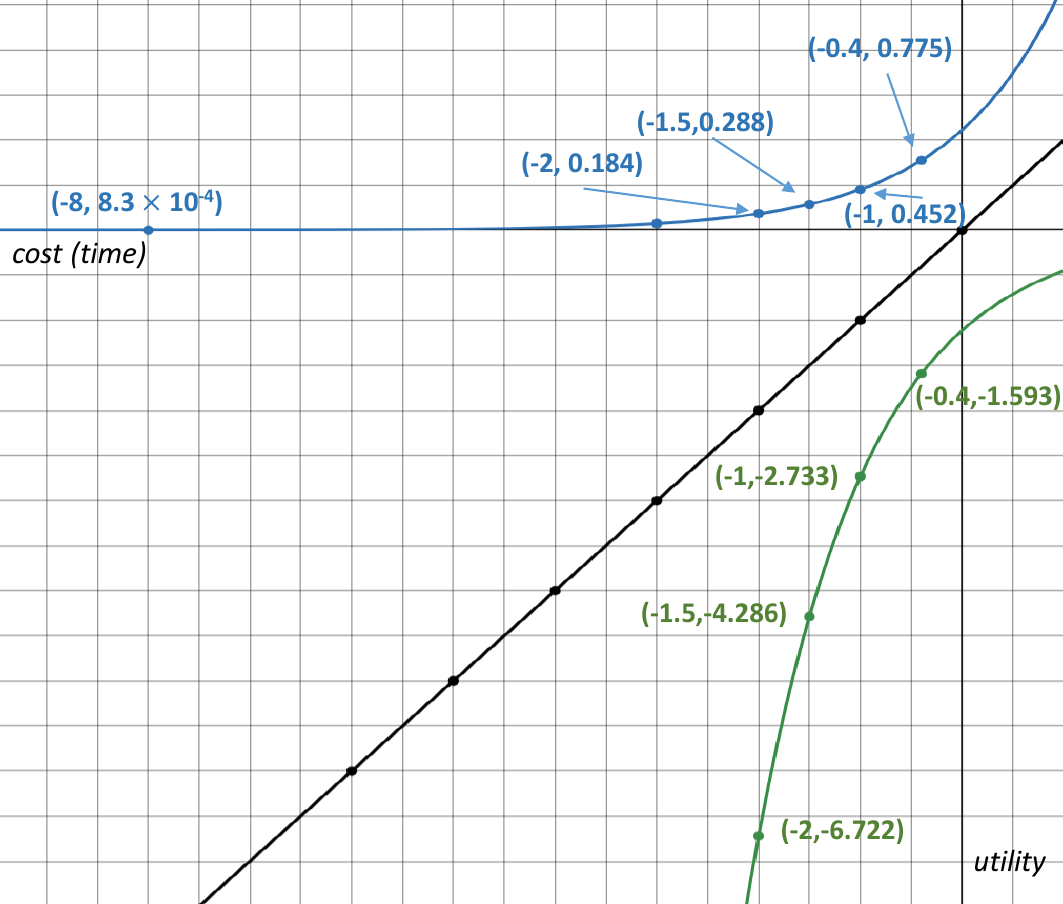}
 	\caption{Exponential utility functions (Equation~\ref{equ:exponentialUtilityFunction}) with $\alpha = 0.9$ of risk-averse (green), risk-seeking (blue), and risk-neutral (black) attitudes.}
 	\label{fig:utilityFunctions-usecase}
\end{figure}

For this demonstration, consider that the planning agent adopts a static risk attitude expressed by the exponential utility function for the risk-averse and risk-seeking attitudes given in Equation~\ref{equ:exponentialUtilityFunction}, where $\alpha=0.9$ indicating a high intensity of the risk attitude and expressed by the linear utility function for the risk-neutral attitude. Figure~\ref{fig:utilityFunctions-usecase} illustrates the exponential utility function for risk-averse and risk-seeking attitudes and the linear utility function for the risk-neutral attitude. Comparing the utilities of the risk-seeking attitude with the risk-averse attitude, we can see that the utility decline with respect to the increase of costs of the risk-seeking attitude is much smaller than the utility decline of the risk-averse attitude. This means that the risk-averse attitude is more sensitive to losing, i.e., incurring high travelling times, compared to the risk-seeking attitude. We can also see that the risk-neutral attitude utility is linear, indicating that it assesses losses as they are without exhibiting any particular attitude. 

\begin{figure}
 	\centering
 	\includegraphics[width=\columnwidth]{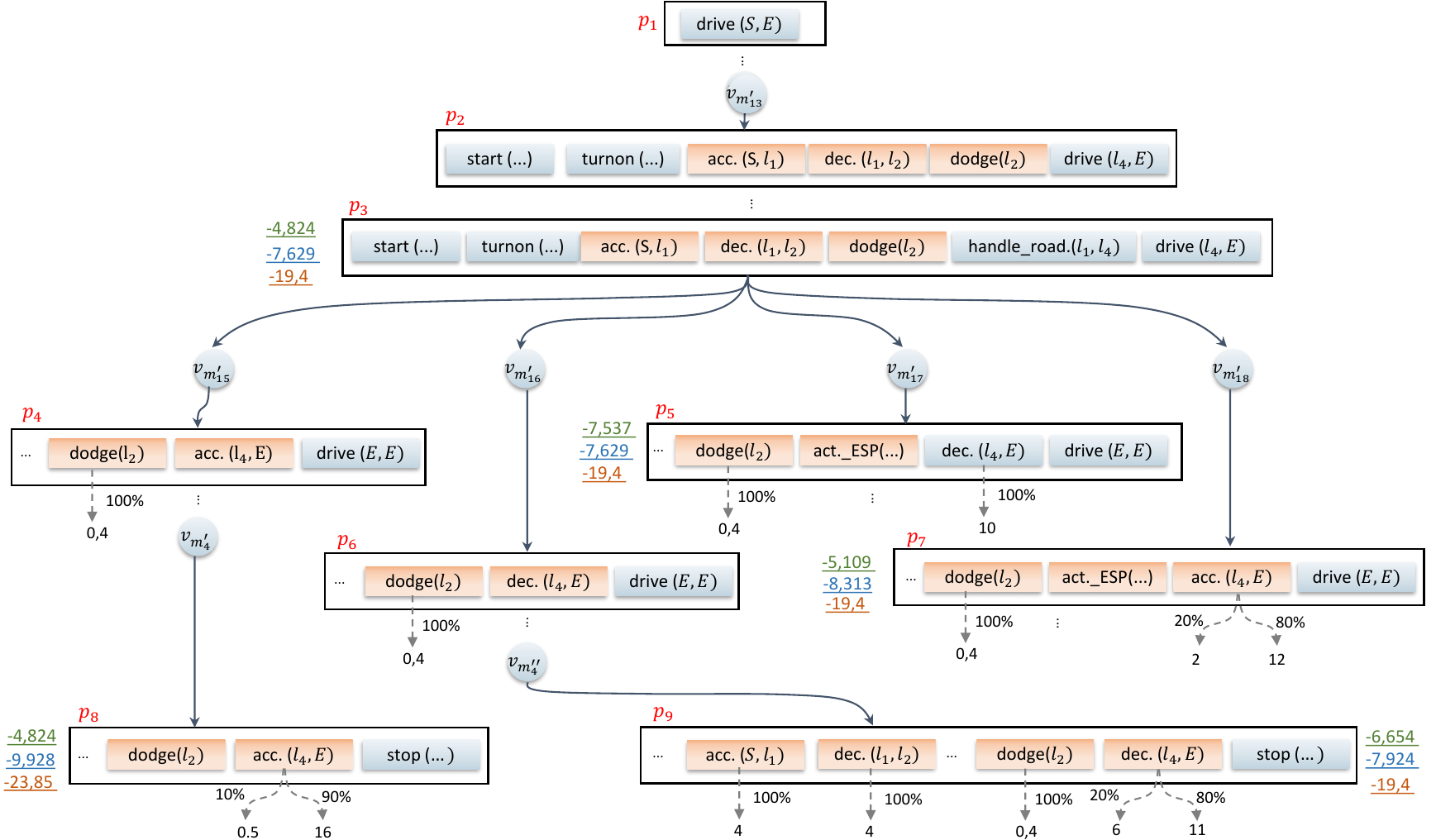}
 	\caption{Part of the search space in risk-aware plan-based HTN planning.}
 	\label{fig:AV-domain-example}
\end{figure}

Figure~\ref{fig:AV-domain-example} shows part of the search space in the risk-aware HTN planning algorithm. Black rectangles represent nodes in the search space, i.e., partial plans, notated as $p_x$. Grey and orange rectangles represent ground compound and primitive tasks with their probability distribution, respectively. Circles represent ground methods. Numbers coloured in green, blue, and orange beside partial plans are the expected utilities of the partial plans for the risk-seeking, risk-averse, and risk-neutral attitudes, respectively. Given the large search space, we focus on displaying only the nodes that can potentially lead to solutions and perform the expected utility calculations for partial plans based on this subset. In the plan generation algorithm, there may be nodes with the highest expected utility that ultimately do not lead to a solution. Although these nodes might be considered during the utility computation in the preprocessing phase, the algorithm will recognise upon reaching them that they do not form valid solutions and will backtrack to explore other possibilities. In this example, we assume that the expected utility of each node is already computed in the preprocessing step. For the risk-averse attitude, the algorithm starts with partial plan $p_1$ in the fringe. Since this is the only node initially in the fringe, the algorithm selects it and removes it from the fringe for decomposition. The result of the decomposition is the partial plan $p_2$, which also gets added and later selected since it is the only node in the fringe. The algorithm continues until selecting the partial plan $p_3$ from the fringe. The four resulting partial plans resulting from decomposing $p_3$, which are $p_4$, $p_6$, $p_5$, and $p_7$ get added to the fringe in a descending order starting from $p_5$ and ending at $p_8$. Thus, the next node that the algorithm selects for decomposition is $p_5$. For the risk-seeking attitudes, at this point, the algorithm chooses partial plan $p_4$. The algorithm continues with adding and removing nodes from the fringe until reaching a primitive task network that constitutes the solution to this planning problem for the risk-seeking attitude. The same procedure applies to risk-seeking and risk-neutral attitudes, where, in this partial search space, the risk-seeking attitude selects the partial plan $p_8$ and continues decomposing the remaining compound task \verb|stop| until reaching the solution. The risk-neutral attitude is indifferent to partial plans $p_5$, $p_6$, and $p_7$ since all of them have the same expected utility for this risk attitude.

To illustrate the choices of each risk attitude among all possible solutions, we perform the brute force method to compute all possible plans that solve the planning problem described here. There are sixteen possible solutions for this planning problem, as shown in Table~\ref{tab:useCase}. For each plan, we show the number of the plan (\textbf{\#P}), the corresponding road (\textbf{road}), the sequence of actions corresponding to the road and its complexities (\textbf{plan}), the logarithm of the expected utility of the resulting plan for the risk-seeking (\textbf{LogEU - RS}) and risk-averse (\textbf{LogEU - RA}) attitudes computed using Equation~\ref{equ:planExpectedUtility-segmentation}, the expected utility of each plan computed by the risk-neutral attitude using Equation~\ref{equ:linear-utility-eu} (\textbf{EU - RN}), and the length of the plan (\bm{$|\pi|$}). We show the logarithmic value of the plan's expected utility for the risk-averse and risk-seeking attitudes for better readability. This value is computed according to Equation~\ref{equ:plan_expected_utility-log}.

Highlighted cells in Table~\ref{tab:useCase} correspond to plans with the highest expected utility for each risk attitude. We see that the risk-neutral agent will be indifferent to choosing \textit{Plan 9}, \textit{Plan 10}, or \textit{Plan 12}. All these plans have an expected cost lower than all other plans. The difference between these plans is in the actions the vehicle can take on the icy road, where the actions of only decelerating, decelerating after activating the ESP, and accelerating after activating the ESP have the same expected value, which is ten hours. Since the action of accelerating without activating the ESP has a higher expected cost, \textit{Plan 11} is not chosen by the risk-neutral attitude.
 
The risk-averse agent chooses \textit{Plan 2}, which is the only plan with guaranteed outcomes, with 100\% probability, for all its constituent actions, i.e., it is the plan with the least risk compared to all plans with the same expected value. In this case, the risk-averse attitude chooses to drive through the long road ($S$, $l_1$) that is guaranteed to take four hours over the short road ($S$, $l_3$) that has 20\% probability of taking six hours but with 80\% probability, it takes two hours. With this choice, the risk-averse agent compromises the chance of sparing two hours of driving to avoid the risk of taking two hours more compared to the long road. The risk-averse agent applies the same logic on the choice between going through the congested area at location $l_5$, which has uncertain waiting times, in comparison to the choice of dodging the construction area at location $l_2$, which can be done in a certain amount of time. Similarly, decelerating on the icy road after activating the ESP is the option that guarantees the highest stability of the vehicle, and thus, it is the preferred choice for this risk attitude. On the contrary, the risk-seeking agent chooses \textit{Plan 11}, which has a risky non-guaranteed outcome. For this agent, choosing the short road ($S$, $l_1$) is preferable over the long road ($S$, $l_3$). The reason for this is that, although the short road is riskier than the long road, it has a high chance of resulting in a better outcome, i.e., saving two hours of driving compared to the long road. Additionally, the risk-seeking agent chooses to accelerate on the icy road without activating the ESP. This action is riskier than all other possible actions that can be executed on this road. Still, with a 10\% probability, it might result in a better outcome, i.e., spending only half an hour on the icy road. We observe that both the risk-seeking and risk-averse agents avoid decelerating without activating the ESP or accelerating after activating the ESP. This behaviour is due to their utility functions with $\alpha=0.9$, making them highly risk-seeking and highly risk-averse, respectively. Consequently, they opt for the most extreme actions: the risk-seeking agent prefers accelerating without ESP, while the risk-averse agent prefers decelerating after activating the ESP. Similarly, we notice that no agent chooses to take the highway ($l_1$, $l_3$) since driving through this road does not increase the expected utility of any agent.

When considering the expected utility of the risk-seeking and risk-averse attitudes for plans that have the same expected cost (e.g. \textit{Plan 1}, \textit{Plan 2}, and \textit{Plan 4}), we can see that the expected utility of the risk-seeking attitude becomes larger for riskier plans, which is the opposite for the risk-averse attitude. In particular, the risk-seeking attitude prefers \textit{Plan 4} over \textit{Plan 2}, which is the opposite of the risk-averse preference. Additionally, comparing the expected utility of the risk-seeking and risk-averse attitudes for the same plan, we notice that the expected utility of the risk-seeking attitude is always lower. The reason for this is that the utility risk-seeking attitude does not decrease as much for larger costs compared to the utility of the risk-averse attitude.

Comparing the length of the solution plans for each risk attitude; we see that, in this particular case study, the risk-seeking agent has a shorter, though riskier, plan compared to the risk-averse agent, which has the extra action of activating the ESP. However, in general, there is no correlation between a particular risk attitude and the plan's length since the choice of the highest expected utility plan depends solely on the plan's constituent actions with their variable costs.

{\footnotesize
\begin{longtable}[H]{|p{0.03\linewidth}|
>{\centering}p{0.14\linewidth}|
>{\centering}p{0.264\linewidth}|
>{\centering}p{0.1\linewidth}|
>{\centering}p{0.09\linewidth}|
>{\centering}p{0.12\linewidth}|
p{0.03\linewidth}|}
\caption{The results of a brute force approach on planning problem \textbf{P1} in the Autonomous Vehicles domain. In each row, we show the plan number, corresponding road, corresponding sequence of actions (i.e., plan), and the log of the plan's expected utility for risk-seeking (logEU - RS) and risk-averse (logEU - RA) attitudes (Equation~\ref{equ:plan_expected_utility-log}), the plan's expected utility of the risk-neutral (EU - RN) attitude, and the plan length $|\pi|$. Highlighted cells correspond to the highest expected utility plans for each risk attitude.} 
\label{tab:useCase} \\
\hline 
\textbf{$\#$P} & \textbf{road} & \textbf{plan} & \textbf{LogEU- RS} & \textbf{LogEU- RA} & \textbf{EU - RN} & \bm{$|\pi|$} \\ \hline
1 & $S$,$l_1$,$l_2$,$l_4$,$E$ & start - turnon - acc. - dec. - dodge - acc. - actESP - acc. - stop & -5,109 & -8,313 & -19,4 & 9 \\ \hline
2 & $S$,$l_1$,$l_2$,$l_4$,$E$ & start - turnon - acc. - dec. - dodge - acc. - actESP - dec. - stop & -7,537 & \cellcolor{red!15}-7,629 & -19,4 & 9 \\ 
\hline
3 & $S$,$l_1$,$l_2$,$l_4$,$E$ & start - turnon - acc. - dec. - dodge - acc. -acc. - stop & -4,824 & -9,928 & -23,85 & 8 \\ 
\hline
4 & $S$,$l_1$,$l_2$,$l_4$,$E$ & start - turnon - acc. - dec. - dodge - acc. - dec. - stop & -6,654 & -7,924 & -19,4 & 8 \\ 
\hline
5 & $S$,$l_3$,$l_1$,$l_2$,$l_4$,$E$ & start - turnon - acc. - acc. - dec. - dodge - acc. - actESP - acc. - stop & -5,248 & -10,159 & -20,8 & 10 \\ 
\hline
6 & $S$,$l_3$,$l_1$,$l_2$,$l_4$,$E$ & start - turnon - acc. - acc. - dec. - dodge - acc. - actESP - dec. - stop & -7,677 & -9,906 & -20,8 & 10 \\ 
\hline
7 & $S$,$l_3$,$l_1$,$l_2$,$l_4$,$E$ & start - turnon - acc. - acc. - dec. - dodge - acc. - acc. - stop & -4,963 & -12,2 & -25,25 & 9 \\ 
\hline
8 & $S$,$l_3$,$l_1$,$l_2$,$l_4$,$E$ & start - turnon - acc. - acc. - dec. - dodge - acc. - dec. - stop &  -6,793 & -10,585 & -20,8 & 9\\ 
\hline
9 & $S$,$l_3$,$l_5$,$l_4$,$E$ & start - turnon - acc. - dec. - brake - acc. - actESP - acc. - stop & -4,214 & -8,945 & \cellcolor{red!15}-18,35 & 9 \\ 
\hline
10 & $S$,$l_3$,$l_5$,$l_4$,$E$ & start - turnon - acc. - dec. - brake - acc. - actESP - dec. - stop & -6,642 & -8,26 & \cellcolor{red!15}-18,35 & 9 \\ 
\hline
11 & $S$,$l_3$,$l_5$,$l_4$,$E$ & start - turnon - acc. - dec. - brake - acc. - acc. - stop & \cellcolor{red!15}-3,928 & -10,559 & -22,8 & 8 \\ 
\hline
12 & $S$,$l_3$,$l_5$,$l_4$,$E$ & start - turnon - acc. - dec. - brake - acc. - dec. - stop & -5,758 & -8,555 & \cellcolor{red!15}-18,35 & 8 \\ 
\hline
13 & $S$,$l_1$,$l_3$,$l_5$,$l_4$,$E$ & start - turnon - acc. - acc. - dec. - brake - acc. - actESP - acc. - stop & -5,729 & 10,961 & -22,15 & 10 \\ 
\hline
14 & $S$,$l_1$,$l_3$,$l_5$,$l_4$,$E$ & start - turnon - acc. - acc. - dec. - brake - acc. - actESP - dec. - stop & -8,157 & -10,276 & -22,15 & 10 \\ 
\hline
15 & $S$,$l_1$,$l_3$,$l_5$,$l_4$,$E$ & start - turnon - acc. - acc. - dec. - brake - acc. - acc. - stop & -5,444 & -12,576 & -26,6 & 9 \\ 
\hline
16 & $S$,$l_1$,$l_3$,$l_5$,$l_4$,$E$ & start - turnon - acc. - acc. - dec. - brake - acc. - dec. - stop & -7,273 & -10,572 & -22,15 & 9 \\
\hline
\end{longtable}}

\subsection{Satellite Problem Instance}
\label{app:satellite-problem-instance}
Say there is one satellite \textit{satellite0} equipped with three instruments \textit{instrument01}, \textit{instrument02}, and \textit{instrument03}, which support \textit{thermography}, \textit{x\_ray}, and \textit{hd\_video} modes, respectively. The initial task network consists of three compound tasks to do observations of three phenomena located in three different directions \textit{Phenomenon4}, \textit{star0}, and \textit{Phenomenon6} and has to be observed in the \textit{thermography}, \textit{hd\_video}, and \textit{x\_ray} modes, respectively. We denote this problem instance \textbf{RA-3obs-1sat-
3mod}.

Listing~\ref{listing:case-study2} shows all possible solutions to this planning problem, where \textit{Plan 1} involves switching on the three instruments in parallel and switching them off at once after taking the images, \textit{Plan 2} involves switching on two instruments in parallel, but then switching them off at once before activating the third instrument, and \textit{Plan 3} involves switching on and off the instruments sequentially such that there is no more than one instrument is activated. The first plan is the riskiest one since it may result in power failure but results in skipping two switching-off actions. The second plan is less risky and results in skipping one switching-off action. The last plan is the safest, but it requires a sequence of switching on and off instruments.

\begin{lstlisting}[label=listing:case-study2,caption=All possible plans for the planning problem described in~\ref{app:satellite-problem-instance}, showstringspaces=false,
  extendedchars=true,
  basicstyle=\scriptsize\ttfamily,
  commentstyle=\slshape,
  stringstyle=\ttfamily,
  breaklines=true,
  breakatwhitespace=true,
  columns=flexible,
  basewidth=.5em,
  xleftmargin=.5cm,
   captionpos=b,
  ]
----- Plan 1 -----
0 switch_on instrument03 satellite0
1 turn_to satellite0 GroundStation0 Phenomenon6
2 calibrate satellite0 instrument03 GroundStation0
3 turn_to satellite0 Phenomenon6 GroundStation0
4 take_image satellite0 Phenomenon6 instrument03 hd_video
5 overload instrument02 satellite0
6 turn_to satellite0 GroundStation0 Phenomenon6
7 calibrate satellite0 instrument02 GroundStation0
8 turn_to satellite0 Star5 GroundStation0
9 take_image satellite0 Star5 instrument02 x_ray
10 superload instrument01 satellite0
11 turn_to satellite0 GroundStation0 Star5
12 calibrate satellite0 instrument01 GroundStation0
13 turn_to satellite0 Phenomenon4 GroundStation0
14 take_image satellite0 Phenomenon4 instrument01 thermography

----- Plan 2 -----
0 switch_on instrument03 satellite0
1 turn_to satellite0 GroundStation0 Phenomenon6
2 calibrate satellite0 instrument03 GroundStation0
3 turn_to satellite0 Phenomenon6 GroundStation0
4 take_image satellite0 Phenomenon6 instrument03 hd_video
5 overload instrument02 satellite0
6 turn_to satellite0 GroundStation0 Phenomenon6
7 calibrate satellite0 instrument02 GroundStation0
8 turn_to satellite0 Star5 GroundStation0
9 take_image satellite0 Star5 instrument02 x_ray
10 switch_off_overload instrument02 instrument02 satellite0
11 switch_on instrument01 satellite0
12 turn_to satellite0 GroundStation0 Star5
13 calibrate satellite0 instrument01 GroundStation0
14 turn_to satellite0 Phenomenon4 GroundStation0
15 take_image satellite0 Phenomenon4 instrument01 thermography

----- Plan 3 -----
0 switch_on instrument03 satellite0
1 turn_to satellite0 GroundStation0 Phenomenon6
2 calibrate satellite0 instrument03 GroundStation0
3 turn_to satellite0 Phenomenon6 GroundStation0
4 take_image satellite0 Phenomenon6 instrument03 hd_video
5 switch_off instrument03 satellite0
6 switch_on instrument02 satellite0
7 turn_to satellite0 GroundStation0 Phenomenon6
8 calibrate satellite0 instrument02 GroundStation0
9 turn_to satellite0 Star5 GroundStation0
10 take_image satellite0 Star5 instrument02 x_ray
11 switch_off instrument02 satellite0
12 switch_on instrument01 satellite0
13 turn_to satellite0 GroundStation0 Star5
14 calibrate satellite0 instrument01 GroundStation0
15 turn_to satellite0 Phenomenon4 GroundStation0
16 take_image satellite0 Phenomenon4 instrument01 thermograph
\end{lstlisting}

\begin{table}[!htp]
\centering
\caption{Detailed information on the three plans for the planning problem \textbf{RA-3obs-1sat-3mod} in the Satellite domain. For each plan, we show the plan number, the plan's expected utility for risk-seeking (RS), risk-averse (RA), and risk-neutral (RN) attitudes, and their corresponding logarithmic values. Highlighted cells correspond to the highest expected utility plans for each risk attitude.}
\label{tab:use-case-sat}
\resizebox{\textwidth}{!}{
\begin{tabular}{|c|r|r|r|r|r|}
\hline
\textbf{\# Plan} & \bm{$EU(\pi)-RS$}               & \bm{$log_{10}(EU)-RS$}         & \bm{$EU(\pi)-RA$}              & \bm{$log_{10}(EU)-RA$}         & \bm{$EU(\pi)-RN$}           \\ \hline\hline
\textbf{1}       & \cellcolor[HTML]{FFCCC9}2,60771E-49 & \cellcolor[HTML]{FFCCC9}-48,8847704 & -8,2667E+147                       & -147,91733                          & -245,1                          \\ \hline
\textbf{2}       & 1,51819E-52                         & -51,818673                          & -7,47968E+68                       & -68,873883                          & \cellcolor[HTML]{FFCCC9}-240,85 \\ \hline
\textbf{3}       & 8,4817E-56                          & -55,0715164                         & \cellcolor[HTML]{FFCCC9}-4,716E+55 & \cellcolor[HTML]{FFCCC9}-55,6735764 & -255                            \\ \hline
\end{tabular}
}
\end{table}

In Table~\ref{tab:use-case-sat}, we present detailed information for each of the three plans outlined in Listing~\ref{listing:case-study2}. For each plan,  we provide the plan number, the plan’s expected utility for risk-seeking (RS), risk-averse (RA), and risk-neutral (RN) attitudes, and their corresponding logarithmic values to enhance readability and facilitate comparison. Highlighted cells correspond to the highest expected utility plans, i.e., the solutions for each risk attitude. The expected utilities are computed using Equation~\ref{equ:planExpectedUtility-segmentation} with $\alpha = 0.5$. We can see that the risk-seeking agent chooses \textit{Plan 1}, the risk-averse agent chooses \textit{Plan 3}, and the risk-neutral agent chooses \textit{Plan 2}. We can also observe from the expected costs of the plans, i.e., the expected utility of the risk-neutral agent, that the expected cost of the plan computed by the risk-averse agent is higher than the other two plans. This is attributed to the fact that the risk-averse agent opts to avoid simultaneous operation of instruments to mitigate the risk of incurring substantial costs. Instead, it prioritises a plan that has a guaranteed cost even if, on average, this plan entails higher costs and is longer compared to the alternatives. The number of plan steps varies according to each risk attitude, with 17, 16, and 15 steps for the risk-averse, neutral, and seeking attitudes, respectively. Specific disparities become apparent in step 5 of all plans (see Listing~\ref{listing:case-study2}), where risk-neutral and seeking solutions accept the risk involved in overloading the satellite with a second instrument, whereas the risk-averse solution avoids it. Another contrast is identifiable in step 10 of \textit{Plan 1} and \textit{Plan 2} in comparison to step 11 in \textit{Plan 3}, where risk-neutral and risk-averse solutions avoid the risk that comes with switching on a third instrument, whereas the risk-seeking solution embraces it.

\clearpage

\bibliographystyle{elsarticle-num-names}
\bibliography{mybibfile}

\end{document}